\documentclass{article}
\usepackage{ijcai23}

\usepackage{times}
\usepackage{soul}
\usepackage{url}
\usepackage[hidelinks]{hyperref}
\usepackage[utf8]{inputenc}
\usepackage[small]{caption}
\usepackage{graphicx}
\usepackage{amsmath}
\usepackage{amsthm}
\usepackage{booktabs}
\usepackage{algorithm}
\usepackage[switch]{lineno}
\usepackage{subcaption}
\usepackage{comment}
\hypersetup{
    colorlinks=true,
    linkcolor=blue,
    filecolor=blue,      
    urlcolor=blue,
    citecolor=blue
}

\usepackage{todonotes}
\usepackage{multirow}
\graphicspath{{figures/}}
    \DeclareGraphicsExtensions{.pdf,.eps,.png,.jpg}
\usepackage{amsmath,amssymb}
\usepackage{bm}	% For bold italic math
\usepackage[binary-units,per-mode=symbol]{siunitx}

\usepackage{booktabs}

\usepackage{mathtools}

\DeclarePairedDelimiter\floor{\lfloor}{\rfloor}

\usepackage[noabbrev,capitalise]{cleveref} % Must be loaded after most packages
\crefname{equation}{}{}	% Use only number in brackets for equation references
\usepackage{tikz}
\usepackage{ upgreek, standalone}

\usetikzlibrary{arrows, decorations, graphs, decorations.pathmorphing, decorations.pathreplacing, decorations.shapes, backgrounds, positioning, fit, petri, bending, intersections, automata, shapes.misc, calc, arrows.meta}

\newcommand*{\True}{\ensuremath{1}}
\newcommand*{\False}{\ensuremath{0}}

\usepackage[nolist]{acronym}
\usepackage[english]{babel}

\begin{acronym}[DRAM]
    \acro{AI}{artificial intelligence}
    \acrodefindefinite{AI}{an}{an}
    \acro{ASIC}{application-specific integrated circuit}
    \acrodefindefinite{ASIC}{an}{an}
    \acro{BD}{bounded delay}
    \acro{BNN}{binarized neural network}
    \acro{CNN}{convolutional neural network}
    \acro{CTM}{convolutional Tsetlin machine}
    \acro{FSM}{finite state machine}
    \acro{LA}{learning automaton}
    \acrodefplural{LA}{learning automata}
    \acrodefindefinite{LA}{an}{a}
    \acro{ML}{machine learning}
    \acrodefindefinite{ML}{an}{a}
    \acro{TA}{Tsetlin automaton}
    \acrodefplural{TA}{Tsetlin automata}
    \acro{TAT}{Tsetlin automaton team}
    \acro{TM}{Tsetlin machine}
    \acro{RTM}{regression Tsetlin machine}
\end{acronym}

\usepackage{algpseudocode}

\algnewcommand{\LineComment}[1]{\State \(\triangleright\) #1}

\def\boxit#1{%
  \smash{\color{red}\fboxrule=1.pt\relax\fboxsep=4.1pt\relax%
  \llap{\rlap{\fbox{\vphantom{0}\makebox[#1]{}}}}}\ignorespaces
}

\title{Building Concise Logical Patterns by Constraining Tsetlin Machine Clause Size} %Rationing

\author{
K. Darshana Abeyrathna$^1$
\and
Ahmed Abdulrahem Othman Abouzeid$^2$\and
Bimal Bhattarai$^{2}$\and
Charul Giri$^2$
\and
Sondre Glimsdal$^2$
\and
Ole-Christoffer Granmo$^2$
\and
Lei Jiao$^2$
\and
Rupsa Saha$^2$
\and
Jivitesh Sharma$^2$
\and
Svein Anders Tunheim$^2$
\And
Xuan Zhang$^3$\thanks{The authors are ordered alphabetically by last name.}
\affiliations
$^2$Centre for Artificial Intelligence Research (CAIR), University of Agder, Kristiansand, Norway\\
$^3$NORCE Norwegian Research Centre, 
$^1$DNV Norway
\emails
ole.granmo@uia.no
}

\begin{document}

\maketitle

\begin{abstract}
\ac{TM} is a logic-based machine learning approach with the crucial advantages of being transparent and hardware-friendly. While TMs match or surpass deep learning accuracy for an increasing number of applications, large clause pools tend to produce clauses with many literals (long clauses). As such, they become less interpretable. Further, longer clauses increase the switching activity of the clause logic in hardware, consuming more power. This paper introduces a novel variant of TM learning -- Clause Size Constrained \acp{TM} (CSC-\acp{TM}) --  where one can set a soft constraint on the clause size. As soon as a clause includes more literals than the constraint allows, it starts expelling literals. Accordingly, oversized clauses only appear transiently. To evaluate CSC-TM, we conduct classification, clustering, and regression experiments on tabular data, natural language text, images, and board games. Our results show that CSC-TM maintains accuracy with up to 80 times fewer literals. Indeed, the accuracy increases with shorter clauses for TREC, IMDb, and BBC Sports. After the accuracy peaks, it drops gracefully as the clause size approaches a single literal. We finally analyze CSC-TM power consumption and derive new convergence properties. 
\end{abstract}

\section{Introduction}

The \ac{TM} \cite{granmo2018tsetlin} is a novel approach to machine learning where groups of \acp{TA} \cite{Tsetlin1961} produce logical (Boolean) expressions in the form of conjunctive clauses (AND-rules). As opposed to the black-box nature of deep neural networks, \acp{TM}  are inherently interpretable. Indeed, they produce models based on sparse disjunctive normal form, which is comparatively easy for humans to understand~\cite{valiant1984learnable}. Additionally, the logical representation combined with automata-based learning make \acp{TM}  natively suitable for hardware implementation, yielding low energy footprint~\cite{wheeldon2020learning}. 

\acp{TM}  now support various architectures, including classification~\cite{granmo2018tsetlin}, convolution~\cite{granmo2019convtsetlin}, regression~\cite{abeyrathna2020nonlinear}, deterministic~\cite{abeyrathna2020deterministic}, weighted~\cite{abeyrathna2020integer}, autoencoder~\cite{bhattaray2023embedding}, contextual bandit~\cite{RaihanNIPS22}, relational~\cite{Saha2022}, and multiple-input multiple-output~\cite{glimsdal2021coalesced} architectures. The independent nature of clause learning allows efficient GPU-based parallelization, providing almost constant-time scaling with reasonable clause amounts~\cite{abeyrathna2020massively}. Several schemes enhance vanilla \ac{TM} learning and inference, such as drop clause~\cite{dropclause} and focused negative sampling~\cite{Glimsdal2022}. These \ac{TM} advances have enabled many applications: keyword spotting~\cite{lei2021kws}, aspect-based sentiment analysis~\cite{rohan2021AAAI}, novelty detection~\cite{bhattarai2021word},  semantic relation analysis~\cite{saha2021semantic}, text categorization~\cite{Rohanblackbox,fakenewsLREC,yadav2022robustness}, game playing~\cite{giri2022logic}, batteryless sensing~\cite{Bakar2022,BakarRahman}, recommendation systems~\cite{Borgersen2022Recommendation}, and word embedding~\cite{bhattaray2023embedding}.

While \acp{TM}  match or surpass deep learning accuracy for an increasing number of applications, large clause pools tend to produce longer clauses containing many literals (input features and their negation). As such, they become less interpretable. Further, longer clauses require more memory and increase the switching activity in hardware, consuming more power. There is currently no direct way to control the size of the clauses learned. The challenge lies in coordinating the decentralized \acp{TA}, each independently learning whether to include a specific literal per clause.  Since the \acp{TA} seek frequent discriminative patterns, the clauses can become arbitrarily long. In short, current learning schemes seem inefficient when it comes to the size of clauses they produce. 

This paper introduces a novel variant of \ac{TM} learning -- Clause Size Constrained \acp{TM}  (CSC-\acp{TM} ) --  where one can set a soft constraint on the clause size. CSC-\ac{TM} revises the \ac{TA} feedback policy for including literals into the clauses. Specifically, the new policy discourages including additional literals once the length of a clause surpasses a predefined constraint. The \acp{TA} instead immediately start expelling literals from the offending clause by reinforcing ``exclude" actions.   Accordingly, oversized clauses only appear transiently. Otherwise, the \ac{TM} feedback scheme is left unchanged. The salient property of our approach is that the limited collections of literals ending up in the clauses maintain high discrimination power. Even with significantly constrained clause size, the performance of CSC-\ac{TM} is not compromised compared with the other \ac{TM} variants.

\paragraph{Paper Contributions:} The contributions of the paper can be summarized as follows:
\begin{itemize}
    \item We propose CSC-\ac{TM} to constrain the size of the clauses by introducing a new policy for training \acp{TM}. 
    \item We demonstrate that CSC-\ac{TM} can indeed constrain clause size within an explicit limit, without compromising accuracy in classification, regression, and clustering.
    \item Using several examples, we show that the shorter clauses become more interpretable. 
    \item  We prove analytically that CSC-\ac{TM} can converge to the intended basic operators when properly configured. 
    \item We describe how constraining the length of the clauses is beneficial for power consumption in embedded hardware solutions due to the reduced switching activity of the clause logic.
\end{itemize}

\section{Training Tsetlin Machines With Constrained Clause Length}\label{sec:brief}
In this section, we outline the difference between the vanilla TM and CSC-TM. For those who are not familiar with TM learning, a detailed description can be found in Appendix 1. 

%A \ac{TM} takes a vector $\mathbf{X}=[x_1,\ldots,x_o]$ of Boolean features as input, to be classified into one of two classes, $y=0$ or $y=1$. Together with their negated counterparts, $\bar{x}_k = \lnot x_k = 1-x_k$, the features form a literal set $L = \{x_1,\ldots,x_o,\bar{x}_1,\ldots,\bar{x}_o\}$.
A \ac{TM} processes a vector $\mathbf{X}=[x_1,\ldots,x_o]$ of propositional (Boolean) features as input, to be classified into one of two classes, $y=0$ or $y=1$. Negating these features produces a set of literals $L$ that consists of the features and their negated counterparts: $L = \{x_1,\ldots,x_o,\neg{x}_1,\ldots,\neg{x}_o\}$.

A \ac{TM} uses conjunctive clauses to represent sub-patterns. The number of clauses is given by a user set parameter~$n$. For a two-class classifier\footnote{A multi-class classifier gets $n$ clauses per class.}, half of the clauses gets positive polarity ($+$). The other half gets negative polarity ($-$). Each clause $C_j^p, j \in \{1, 2, \ldots, n/2\}, p \in \{-,+\},$ then becomes:
\begin{equation}
C_j^p(\mathbf{X})=\bigwedge_{l_k \in L_j^p} l_k.
\end{equation}
\noindent Here, $j$ is the index of the clause, $p$ its polarity, while $L_j^p$ is a subset of the literals $L$, $L_j^p \subseteq L$. For example, the clause $C_1^+(\mathbf{X}) = \neg x_1 \land x_2$ has index $1$, polarity $+$, and consists of the literals $L_1^+ = \{\neg x_1, x_2\}$. Accordingly, the clause outputs~$1$ if $x_1 =0$ and $x_2 = 1$, and $0$ otherwise.

The clause outputs are combined into a classification decision through summation and thresholding using the unit step function $u(v) = 1 ~\mathbf{if}~ v \ge 0 ~\mathbf{else}~ 0$:
\begin{equation}
\textstyle
\hat{y} = u\left(\sum_{j=1}^{n/2} C_j^+(\mathbf{X}) - \sum_{j=1}^{n/2} C_j^-(\mathbf{X})\right).
\end{equation}
Namely, classification is performed based on a majority vote, with the positive clauses voting for $y=1$ and the negative for $y=0$. The classifier $\hat{y} = u\left((x_1 \land \lnot x_2) + (\lnot x_1 \land x_2) - (x_1 \land x_2) - (\lnot x_1 \land \lnot x_2)\right)$, e.g., captures the XOR-relation.

For training, a dedicated team of \acp{TA} composes each clause $C_j^p$. Each \ac{TA} of clause $C_j^p$ decides to either \emph{Include} or \emph{Exclude} a specific literal $l_k$ in the clause. A \ac{TA} makes its decision based on the feedback it receives in the form of Reward, Inaction, and Penalty. There are two types of feedback associated with \ac{TM} learning: Type~I Feedback and Type~II Feedback. Type~I Feedback stimulates formation of frequent patterns, which suppresses false negative classifications. Type~II Feedback, on the other hand, increases the discrimination power of the patterns, counteracting false positive classifications.

The difference between vanilla TM and CSC-TM lies in Type~I Feedback. Type~II Feedback remains the same for both schemes. Table~\ref{table:type_i_feedback_New} shows how CSC-TM modifies Type~I Feedback to constrain clause size. The modification is highlighted in red. As seen, we introduce an additional condition for triggering the two leftmost feedback columns. These two columns make the clause mimic frequent patterns by reinforcing inclusion of ``1"-valued literals with probability $\frac{s-1}{s}$ and by reinforcing exclusion of ``0"-valued literals with probability $\frac{1}{s}$.
The two rightmost columns, on the other hand, exclusively reinforce exclusion of literals.

CSC-TM requires that the size $\|C^p_j(\bf{X})\|$ of the clause is within a constraint $\mathbf{b}$ to give access to the two leftmost columns. Accordingly, as soon as the number of literals in the clause surpasses the constraint $\bf{b}$, only \emph{Exclude} actions are reinforced. The reason is that all the \ac{TA} feedback then comes from the two rightmost columns, which reward \emph{Exclude} and penalizes \emph{Include} independently with probability $\frac{1}{s}$. As a result, the clause starts expelling literals when oversized, which means that oversized clauses only appear transiently.

\begin{table}[h!]
\centering
\begin{tabular}{c|ccccc}
\multicolumn{2}{r|}{$C^p_j(\bf{X})\boxit{.93in}~ \land$ $(\|C^p_j(\bf{X})\| \le b)$ }&\multicolumn{2}{c}{\True}&\multicolumn{2}{c}{\False}\\ 
\multicolumn{2}{r|}{$x_k$/$\lnot x_k$}&{\True}&{\False}&{\True}&{\False}\\
\hline
\hline
\multirow{3}{*}{\begin{tabular}[c]{@{}c@{}}TA: \bf Include \\\bf Literal\end{tabular}}&\multicolumn{1}{c|}{$P(\mathrm{Reward})$}&$\frac{s-1}{s}$&NA&$0$&$0$\\
&\multicolumn{1}{c|}{$P(\mathrm{Inaction})$}&$\frac{1}{s}$&NA&$\frac{s-1}{s}$&$\frac{s-1}{s}$\\
&\multicolumn{1}{c|}{$P(\mathrm{Penalty})$}&$0$&NA&$\frac{1}{s} $&$\frac{1}{s}$\\
\hline
\multirow{3}{*}{\begin{tabular}[c]{@{}c@{}}TA: \bf Exclude \\\bf Literal\end{tabular}}&\multicolumn{1}{c|}{$P(\mathrm{Reward})$}&$0$&$\frac{1}{s}$&$\frac{1}{s}$ &$\frac{1}{s}$\\
&\multicolumn{1}{c|}{$P(\mathrm{Inaction})$}&$\frac{1}{s}$&$\frac{s-1}{s}$&$\frac{s-1}{s}$ &$\frac{s-1}{s}$\\
&\multicolumn{1}{c|}{$P(\mathrm{Penalty})$}&$\frac{s-1}{s}$&$0$&$0$&$0$\\
\hline
\end{tabular}
\caption{Type~I Feedback for CSC-TM. The feedback is for a single \ac{TA} that decides whether to Include or Exclude a given literal $x_k/\neg x_k$ into $C^p_j$. NA means not applicable. $s$ is a hyper-parameter greater than 1. }
\label{table:type_i_feedback_New}
\end{table}

In the following sections, we analyze the impact CSC-TM has on convergence. We further investigate how constraining clause size affects accuracy in classification, regression, and clustering. Finally, we discuss effects on hardware complexity and energy consumption.

\section{Convergence Analysis}
Here we analyse the convergence property of two basic operators using the CSC-TM. By studying the XOR and the OR operators, we conjecture\footnote{Here the analysis is only constrained to the XOR and OR operator, rather than a general case. For this reason, we use the word conjecture. Nevertheless, the analysis can still offer useful insights.} that if the number of the literals in a clause is sufficient to represent a sub-pattern (or a group of sub-patterns), the \ac{TM} can learn the intended sub-pattern (or a group of sub-patterns). However, if the required number of literals for a sub-pattern is greater than the literal budget, the sub-pattern cannot be learnt. 
\subsection{XOR operator}
Here we study the convergence of the XOR operator when only one literal is given, i.e.,  $(\|C^i_j(\bf{X})\| = 1)$. We can then show that if the budget is not sufficient, the sub-pattern cannot be learnt. Clearly, the sub-patterns in XOR are mutual exclusive, and one literal cannot capture fully any sub-pattern of XOR. In what follows, we will show how the TM reacts upon training samples of XOR. 

\begin{table}
\centering
\begin{tabular}{ |c|c|c| } 
\hline
$x_1$ & $x_2$ & Output \\ 
0 & 0 & 0 \\ 
1 & 1 & 0 \\ 
0 & 1 & 1 \\
\hline
\end{tabular}
\caption{A sub-pattern in ``XOR'' case.}
\label{xorlogichalf}
\end{table} 

As already proven in~\cite{jiao2021convergence}, the vanilla TM can converge almost surely to the intended sub-pattern, i.e., $C^i_j=\neg x_1\wedge x_2$, when the training samples in Table~\ref{xorlogichalf} is given. 
However, when $(\|C^i_j(\bf{X})\| = 1)$ is given in addition, the only absorbing state of the system, i.e., $C^i_j=\neg x_1\wedge x_2$, disappears, making the system recurrent. More specifically, for vanilla TM, according to~\cite{jiao2021convergence}, when the training samples for $x_1=0$, $x_2=1$, $y=1$ is given to the system and when $\mathrm{TA}^3_1$=\emph{Exclude}, $\mathrm{TA}^3_2$=\emph{Include}, and $\mathrm{TA}^3_4$=\emph{Exclude}, the following transition\footnote{It is the 2nd transition of \textbf{Case 1} in Subsection 3.2.1 in~\cite{jiao2021convergence}. The transition diagram is derived based on the current status of the system and the input training samples. For details please refer to ~\cite{jiao2021convergence}. } holds for $\mathrm{TA}_3^3$.

\begin{minipage}{0.18\textwidth}
%$\Delta$ \textbf{Assume that:}\\
Condition: $x_{1}=0$, $x_{2}=1$, $y=1$, $\mathrm{TA}^3_{4}$=\emph{Exclude}.\\
Therefore, we have Type I Feedback for
literal $x_{2}=1$, $C_{3}= \neg x_{1} \wedge x_{2} = 1$.\\ 
\end{minipage}
\hspace{0.35cm}\resizebox{0.17\textwidth}{!}{
\begin{minipage}{0.17\textwidth}
\begin{tikzpicture}[node distance = .35cm, font=\Huge]
\tikzstyle{every node}=[scale=0.35]
% NODES
\node[state] (E) at (1,1) {};
\node[state] (F) at (2,1) {};
\node[state] (G) at (3,1) {};
\node[state] (H) at (4,1) {};
\node[state] (A) at (1,2) {};
\node[state] (B) at (2,2) {};
\node[state] (C) at (3,2) {};
\node[state] (D) at (4,2) {};
% LETTERS
\node[thick] at (0,1) {$R$};
\node[thick] at (0,2) {$P$};
\node[thick] at (1.5,2.75) {$I$};
\node[thick] at (3.5,2.75) {$E$};
\draw[dotted, thick] (2.5,0.5) -- (2.5,2.5);
% ARROWS
\draw[every loop]
(F) edge[bend left] node [scale=1.2, above=0.1 of C] {} (E)
(E) edge[loop left = 45] node [scale=1.2, below=0.1 of E] {\mbox{\Huge \boldmath$u_1\frac{s-1}{s}$}} (E);

\end{tikzpicture}
\end{minipage}
}\\
Here the superscript of $\mathrm{TA}^3_1$ is the clause index and the subscript is the TA index. $\mathrm{TA}^3_1$ has two actions, i.e., \emph{Include} or \emph{Exclude} $x_{1}$. Similarly,  $\mathrm{TA}^3_2$ corresponds to \emph{Include} or \emph{Exclude} $\neg x_{1}$.  $\mathrm{TA}^3_3$ and $\mathrm{TA}^3_4$ determine the behavior of the $x_2$ and $\neg x_2$, respectively. $P$ and $R$ here mean penalty and reward respectively while $I$ and $E$ denote \emph{Include} and \emph{Exclude} respectively. $u_1$ is a constant in $[0,1]$.  

Clearly, for vanilla TM, the new training sample $x_1=0$, $x_2=1$, $y=1$ will encourage  $\mathrm{TA}^3_3$ to be included, reinforcing $C_{3}$ being in the form $\neg x_{1} \wedge x_{2}$. 
However, when the constraint $(\|C^i_j(\bf{X})\| = 1)$ is given in addition, the transition of $\mathrm{TA}_3^3$ changes to:

\begin{minipage}{0.18\textwidth}
%$\Delta$ \textbf{Assume that:}\\
Condition: $x_{1}=0$, $x_{2}=1$, $y=1$, $\mathrm{TA}^3_{4}$=\emph{Exclude}.\\
Therefore, we have Type I Feedback for
literal $x_{2}=1$, $C_{3}= \neg x_{1} \wedge x_{2} \wedge0 = 0$.\\ 
\end{minipage}
\hspace{0.35cm}\resizebox{0.17\textwidth}{!}{
\begin{minipage}{0.17\textwidth}
\begin{tikzpicture}[node distance = .35cm, font=\Huge]
\tikzstyle{every node}=[scale=0.35]
% NODES
\node[state] (E) at (1,1) {};
\node[state] (F) at (2,1) {};
\node[state] (G) at (3,1) {};
\node[state] (H) at (4,1) {};
\node[state] (A) at (1,2) {};
\node[state] (B) at (2,2) {};
\node[state] (C) at (3,2) {};
\node[state] (D) at (4,2) {};
% LETTERS
\node[thick] at (0,1) {$R$};
\node[thick] at (0,2) {$P$};
\node[thick] at (1.5,2.8) {$I$};
\node[thick] at (3.5,2.8) {$E$};
\draw[dotted, thick] (2.5,0.5) -- (2.5,2.5);
% ARROWS
\draw[every loop]
(A) edge[bend left] node [scale=1.2, above=0.1 of C] {} (B)
(B) edge[bend left] node [scale=1.2, above=0.1 of B] {\mbox{\Huge \boldmath$~~~~~~u_1\frac{1}{s}$}} (C);
\end{tikzpicture}
\end{minipage}
}\\

The above change will make the only absorbing state, i.e., $C_3=\neg x_1\wedge x_2$, disappear. Understandably, given one literal that has already been included, the system will not encourage more literals to be included. Therefore, the TM, given the clause length being 1, cannot almost surely capture any sub-pattern in XOR. The above analysis also confirms that the newly added length constraint  operates as expected, i.e., it indeed discourages more literals to be included once the length budget is reached.    The complete proof can be found in Appendix 2.   

\subsection{OR operator}
Now we study the OR operator, aiming at showing the fact that when the budget of the literals in a clause is sufficient to represent a sub-pattern (or a group of sub-patterns), the TM can learn the intended sub-pattern (or the intended group of sub-patterns). Before we study the convergence for the OR operator, let us revisit its nature. There are three sub-patterns that can rigger a positive output, i.e., ($x_1=1$, $x_2=1$), ($x_1=0$, $x_2=1$), and ($x_1=1$, $x_2=0$). To represent each of the sub-pattern explicitly (or individually), we need two literals. However, two sub-clauses can also be represented jointly by one literal. Clearly, $C=x_1$ can cover both ($x_1=1$, $x_2=1$) and ($x_1=1$, $x_2=0$) while $C=x_2$ can cover both ($x_1=1$, $x_2=1$) and ($x_1=0$, $x_2=1$). This gives the TM possibility to learn the intended OR operator with clauses that has one literal, in collaboration with the hyper-parameter\footnote{The hyper-parameter $T$ is utilized to guide different clauses to learn distinct sub-patterns. The details can be found in~\cite{jiao2021convergence}.} $T$.

Based on the analysis in~\cite{jiao2021convergenceAND}, we understand that if $T=\floor{\frac{m}{2}}$, and when $T$ clauses learn $x_1$ and the other $T$ clauses learn $x_2$, the system is absorbed.  This absorbing state learns the intended OR operator and also coincides with the requirement for the CSC-TM, which indicate that the OR operator can possibly be learnt by the CSC-TM given literal length budget 1.  In what follows, we show that the other absorbing states for the OR operator, i.e.,  with clauses that require more than one literal, will not be encouraged due to the newly added length constraint. 

Similar to the analysis for the XOR case, once we revisit the transitions for the absorbing states with 2 literals, we realized that the absorbing states are not absorbing any more. More specifically, the Type I Feedback will encourage the included literal to be excluded. For example, in vanilla TM, for the sub-pattern below: 
\begin{align}\label{ANDlogic}
P\left ( y=1 | x_{1}=1, x_{2}=1 \right ) = 1, \\
P\left ( y=0 | x_{1}=0, x_{2}=0 \right ) = 1, \nonumber
\end{align}
the transition of $\mathrm{TA}_{3}$ when its current action is \emph{Include} and $\mathrm{TA}_{1}=$\emph{Include} and $ \mathrm{TA}_{2}=$\emph{Exclude}, namely 

%\hspace{0.5cm}
\begin{minipage}{0.18\textwidth}
Condition:
$x_{1}=1$, $x_{2}=1$, $y=1$, $\mathrm{TA}_{4}=\text{Exclude}$.\\
Thus, Type I, $x_2=1$,\\ $C=x_{1}\wedge x_{2}=1$,
\end{minipage}\hspace{0.35cm}
\resizebox{0.17\textwidth}{!}{
\begin{minipage}{0.17\textwidth}
\begin{tikzpicture}[node distance = .25cm, font=\Huge]
    \tikzstyle{every node}=[scale=0.35]
    
    %  NODES
    \node[state] (E) at (1,1) {};
    \node[state] (F) at (2,1) {};
    \node[state] (G) at (3,1) {};
    \node[state] (H) at (4,1) {};
  
    \node[state] (A) at (1,2) {};
    \node[state] (B) at (2,2) {};
    \node[state] (C) at (3,2) {};
    \node[state] (D) at (4,2) {};
    
    % LETTERS
    \node[thick] at (0,1) {$\boldmath R$};
    \node[thick] at (0,2) {$\boldmath P$};
    \node[thick] at (1.5,2.75) {$\boldmath  I$};
    \node[thick] at (3.5,2.75) {$\boldmath E$};
    
    \draw[dotted, thick] (2.5,0.5) -- (2.5,2.5);
    
    % ARROWS
    \draw[every loop]
    (F) edge[bend left] node {} (E)
    (E) edge[loop left=45] node [scale=1.2, below=0.1 of E] {\mbox{\Huge \boldmath$u_1\frac{s-1}{s}$}} (E);

\end{tikzpicture}
\end{minipage}
}\\
is replaced in CSC-TM by

\begin{minipage}{0.18\textwidth}
Condition:
$x_{1}=1$, $x_{2}=1$, $y=1$, $\mathrm{TA}_{4}=\text{Exclude}$.\\
Thus, Type I, $x_2=1$,\\ $C=x_{1} \wedge x_{2}\wedge 0=0$.
\end{minipage}\hspace{0.35cm}
\resizebox{0.17\textwidth}{!}{
\begin{minipage}{0.17\textwidth}
\begin{tikzpicture}[node distance = .35cm, font=\Huge]
    \tikzstyle{every node}=[scale=0.35]
    
    %  NODES
    \node[state] (E) at (1,1) {};
    \node[state] (F) at (2,1) {};
    \node[state] (G) at (3,1) {};
    \node[state] (H) at (4,1) {};
  
    \node[state] (A) at (1,2) {};
    \node[state] (B) at (2,2) {};
    \node[state] (C) at (3,2) {};
    \node[state] (D) at (4,2) {};
    
    % LETTERS
    \node[thick] at (0,1) {$R$};
    \node[thick] at (0,2) {$P$};
    \node[thick] at (1.5,2.75) {$I$};
    \node[thick] at (3.5,2.75) {$E$};
    
    \draw[dotted, thick] (2.5,0.5) -- (2.5,2.5);
    
     % ARROWS
    \draw[every loop]
    (A) edge[bend left] node {} (B)
    (B) edge[bend left] node [scale=1.2, above=0.1 of E] {\mbox{\Huge \boldmath$~~~~~~u_1\frac{1}{s}$}} (C);

\end{tikzpicture}
\end{minipage}
}\\
Clearly, the state, i.e., $x_1\wedge x_2$ is not absorbing any more, and the new constraint encourages the TA with included literal, in this case $\mathrm{TA}_{3}$, to move towards ``Exclude". Similar cases apply also to ($x_1=1$,$x_2=0$) and ($x_1=0$, $x_2=1$). 
Based on the above analysis, we can conclude that the CSC-TM can still learn OR operator but only with clauses that include 1 literal.  

Note that although the length of the clauses are constrained, it may still happen that the length of a clause is over the budget. First, the length of a clause may be over the budget when the system is blocked. Consider an extreme case for the OR operator when $T$ clauses have $x_1$ and $T-1$ clauses follow $x_2$. In this situation, due to the randomness, a clause may become $\neg x_1\wedge x_2$ based on a single training sample. In this situation, the system is blocked by $T$ and it will not be updated any longer. Nevertheless, although this event may happen, the probability is very low. The reason is that it requires that both TAs happen to be in the boundary states at the \emph{Exclude} side. This must happen at the same time as a training sample triggers the transitions in both of TAs toward the \emph{Include} side.   

Another reason of being over the budget is that the Type~II Feedback is not constrained and can produce clauses with more literals than the budget. In this case, the TAs of the included literals will all be in the middle states, and the literals will be quickly swapping in and out of the clause, until the training stops. In this way, Type II Feedback ensures that all the literals are explored. During literal exploration, although the length of the clause can be large, the accuracy stays low until an accurate literal configuration is found. Accordingly, the included literals at boundary due to Type II Feedback do not necessarily contribute positively to the classification. This can be observed from the numerical results for image processing when the literal budget is low (demonstrated for CIFAR-2 and MNIST with convolution in Table~\ref{table_performance}). Specifically, the convolution needs more literals so that the clause position also can be stored. With too few literals, the clauses precision suffers, triggering a large degree of Type II Feedback. The clauses will then be unable  to settle because of the tight literal budget. As a result, the TAs of the included literals will not be able to progress deeply into the \emph{Include} side of the their state space. Consequently, Type II Feedback will persistently continue to experiment with new candidate literals, without finding a sufficiently sparse high-accuracy configuration.

\section{Empirical Results}\label{sec:empirical}
%%rewrite to highlight the lesser clause length and interpretability aspects

In this section, we investigate the performance of CSC-TM, focusing on accuracy and interpretability. To this end,  we evaluate classification, regression, and clustering performance on various datasets spanning natural language, images, board games, and tabular data. The experiments use a CUDA implementation of CSC-TM and runs on Intel Xeon Platinum 8168 CPU at $2.70$ GHz and a Nvidia DGX-2 with Tesla V100 GPU. We describe the details of each task in respective sub-sections and summarize the findings in Tables~\ref{table_performance},  \ref{table_performance_clauses_mnist},  and~\ref{table_lctm}.

\setlength\tabcolsep{5pt} 
\begin{table*}
\centering
\small
\begin{tabular}{|c|c|c|c|c|c|c|c|c|} 
\hline
\multicolumn{1}{|c|}{ } & \multicolumn{8}{c|}{\textbf{Accuracy (Literals per Clause)}}\\ 
\cline{2-9}
\multicolumn{1}{|r|}{\textbf{Budget}$\rightarrow$} & \multicolumn{1}{c|}{$\le$1} & \multicolumn{1}{c|}{$\le$2} & \multicolumn{1}{c|}{$\le$4} & \multicolumn{1}{c|}{$\le$8} & \multicolumn{1}{c|}{$\le$16} & \multicolumn{1}{c|}{$\le$32} &
\multicolumn{1}{c|}{$\le$64} & \multicolumn{1}{c|}{All}\\ 
\hline
\hline
BBC Sports & \begin{tabular}{c}$98.65$\\(1.59)\end{tabular} & \begin{tabular}{c}$98.65$\\(1.65)\end{tabular} & \begin{tabular}{c}$98.65$\\(1.71)\end{tabular} &  \begin{tabular}{c}$99.1$\\(1.93)\end{tabular} & \begin{tabular}{c}$98.2$\\(2.15)\end{tabular} & \begin{tabular}{c}$98.2$\\(2.64)\end{tabular} &\begin{tabular}{c}$98.2$\\(3.35)\end{tabular} & \begin{tabular}{c}$94.14$\\(44.14)\end{tabular}\\
\hline
TREC &  \begin{tabular}{c}$91.8$\\(1.05)\end{tabular} & \begin{tabular}{c}$91.6$\\(1.07)\end{tabular} & \begin{tabular}{c}$92.4$\\(1.07)\end{tabular} & \begin{tabular}{c}$90.2$\\(1.08)\end{tabular} & \begin{tabular}{c}$90.4$\\(1.1)\end{tabular} & \begin{tabular}{c}$93.2$\\(1.12)\end{tabular} & \begin{tabular}{c}$90.6$\\(1.16)\end{tabular} & \begin{tabular}{c}$85.8$\\(96.24)\end{tabular}\\\hline
R8 & \begin{tabular}{c}$95.08$\\(1.09)\end{tabular} & \begin{tabular}{c}$94.54$\\(1.13)\end{tabular} & \begin{tabular}{c}$95.08$\\(1.14)\end{tabular} & \begin{tabular}{c}$94.9$\\(1.19)\end{tabular} & \begin{tabular}{c}$95.08$\\(1.26)\end{tabular} & \begin{tabular}{c}$95.26$\\(1.41)\end{tabular} & \begin{tabular}{c}$95.08$\\(1.7)\end{tabular} & \begin{tabular}{c}$95.8$\\(21.84)\end{tabular}\\ 
\hline
\begin{comment}
Saeed &  &  &  &  &  &  &  \\ 
\hline
Saeed &  &  &  &  &  &  &  \\ 
\hline
\end{comment}
\begin{tabular}{c}California Housing\\(5-bins)\end{tabular} & \begin{tabular}{c}$59.25$\\(1.08)\end{tabular} & \begin{tabular}{c}$62.28$\\(1.46)\end{tabular} & \begin{tabular}{c}$64.2$\\(3.09)\end{tabular} & \begin{tabular}{c}$65.02$\\(6.66)\end{tabular} & \begin{tabular}{c}$65.24$\\(12.92)\end{tabular} & \begin{tabular}{c}$65.26$\\(17.15)\end{tabular} & \begin{tabular}{c}$65.29$\\(18.48)\end{tabular} & \begin{tabular}{c}$65.79$\\(20.22)\end{tabular}\\
\hline
SEMEVAL & \begin{tabular}{c}$93.65$\\(1.23)\end{tabular} & \begin{tabular}{c}$93.00$\\(1.26)\end{tabular} & \begin{tabular}{c}$92.25$\\(1.32)\end{tabular} & \begin{tabular}{c}$93.2$\\(1.41)\end{tabular} & \begin{tabular}{c}$92.95$\\(1.62)\end{tabular} & \begin{tabular}{c}$93.00$\\(2.88)\end{tabular} & \begin{tabular}{c}$93.1$\\(5.67)\end{tabular} & \begin{tabular}{c}$93.63$\\(142.97)\end{tabular}\\  
\hline
IMDb & \begin{tabular}{c}$81.58$\\(1.08)\end{tabular} & \begin{tabular}{c}$81.51$\\(1.22)\end{tabular} & \begin{tabular}{c}$81.28$\\(1.28)\end{tabular} & \begin{tabular}{c}$82.01$\\(1.42)\end{tabular} & \begin{tabular}{c}$83.44$\\(1.75)\end{tabular} & \begin{tabular}{c}$85.67$\\(2.76)\end{tabular} & \begin{tabular}{c}$87.73$\\(4.05)\end{tabular} & \begin{tabular}{c}$84.23$\\(27.02)\end{tabular}\\ 
\hline
CIFAR-2 & \begin{tabular}{c}$69.82$\\(30.8)\end{tabular} & \begin{tabular}{c}$79.65$\\(30.1)\end{tabular} & \begin{tabular}{c}$87.01$\\(13.5)\end{tabular} & \begin{tabular}{c}$91.21$\\(6.8)\end{tabular} & \begin{tabular}{c}$93.23$\\(10.8)\end{tabular} & \begin{tabular}{c}$93.99$\\(20.1)\end{tabular} & \begin{tabular}{c}$94.24$\\(34.1)\end{tabular} & \begin{tabular}{c}$94.18$\\(60.4)\end{tabular}  \\ 
\hline
MNIST & \begin{tabular}{c}$92.09$\\(1.0)\end{tabular}& \begin{tabular}{c}$97.42$\\(1.4)\end{tabular}&\begin{tabular}{c} $98.34$\\ (3.0)\end{tabular} &\begin{tabular}{c}$98.40$\\(5.8)\end{tabular} &\begin{tabular}{c} $98.42$\\(10.8)\end{tabular}&\begin{tabular}{c}
$98.38$\\ (19.9)\end{tabular}&\begin{tabular}{c}
$98.33$\\ (33.0)\end{tabular}&\begin{tabular}{c}
$98.32$\\ (47.7)\end{tabular}\\
\hline
MNIST w/conv.& \begin{tabular}{c}$40.94$\\(18.6)\end{tabular}& \begin{tabular}{c}$60.55$\\(15.3)\end{tabular}&\begin{tabular}{c} $95.93$\\ (5.8)\end{tabular} &\begin{tabular}{c}$99.22$\\(7.1)\end{tabular} &\begin{tabular}{c} $99.33$\\(13.4)\end{tabular}&\begin{tabular}{c}
$99.29$\\ (23.5)\end{tabular}& \begin{tabular}{c}
$99.30$\\ (34.2)\end{tabular}& \begin{tabular}{c}
$99.28$\\ (40.3)\end{tabular} \\
\hline
\begin{tabular}{c}Energy Performance\\(Regression)\end{tabular}& \begin{tabular}{c}$5.65$\\(1.0)\end{tabular}& \begin{tabular}{c}$2.44$\\(1.9)\end{tabular}&\begin{tabular}{c} $1.05$\\ (3.9)\end{tabular} &\begin{tabular}{c}$0.86$\\(6.2)\end{tabular} &\begin{tabular}{c} $0.78$\\(9.3)\end{tabular}&\begin{tabular}{c}
$0.66$\\ (11.2)\end{tabular}&\begin{tabular}{c}
$0.63$\\ (11.3)\end{tabular}&\begin{tabular}{c}
$0.59$\\ (11.5)\end{tabular} \\
\hline

 \begin{tabular}{c}Hex\\(Reinforcement Learning)\end{tabular} & \begin{tabular}{c}$67.59$\\(2.7)\end{tabular}& \begin{tabular}{c}$74.69$\\(2.4)\end{tabular}&\begin{tabular}{c} $77.79$\\ (3.0)\end{tabular} &\begin{tabular}{c}$79.49$\\(4.2)\end{tabular} &\begin{tabular}{c} $81.23$\\(6.1)\end{tabular}&\begin{tabular}{c}
$81.60$\\ (9.3)\end{tabular}& \begin{tabular}{c}
$82.17$\\ (10.2)\end{tabular}& \begin{tabular}{c}
$81.43$\\ (13.8)\end{tabular} \\
\hline
\end{tabular}
\caption{Performance on multiple data sets for a literal budgets of 1, 2, 4, 8, 16, 32, 64, and all (no constraint) literals. Other than for Regression analysis, each field reports the worst maximum accuracy across 5 independent runs, followed by, in brackets, the average number of literals used by the TM. For Energy Performance, we report MAE instead of accuracy.}
\label{table_performance}
\end{table*}

\begin{table*}
\centering
\small
\begin{tabular}{|c|c|c|c|c|c|c|c|c|} 
\hline
\multicolumn{1}{|c|}{ } & \multicolumn{8}{c|}{\textbf{Accuracy (Literals per Clause)}}\\ 
\cline{2-9}
\multicolumn{1}{|r|}{\textbf{Budget}$\rightarrow$} & \multicolumn{1}{c|}{$\le$1} & 
\multicolumn{1}{c|}{$\le$2} & \multicolumn{1}{c|}{$\le$4} & \multicolumn{1}{c|}{$\le$8} & \multicolumn{1}{c|}{$\le$16} & \multicolumn{1}{c|}{$\le$32} &
\multicolumn{1}{c|}{$\le$64} & \multicolumn{1}{c|}{All}\\ 
\hline
\hline
MNIST w/250 clauses & \begin{tabular}{c}$88.20$\\(1.1)\end{tabular} & \begin{tabular}{c}$93.02$\\(1.4)\end{tabular} & \begin{tabular}{c}$95.96$\\(2.1)\end{tabular} &  \begin{tabular}{c}$96.65$\\(2.8)\end{tabular} & \begin{tabular}{c}$96.92$\\(3.2)\end{tabular} & \begin{tabular}{c}$97.03$\\(3.6)\end{tabular}  &\begin{tabular}{c}$97.06$\\(3.8)\end{tabular} & \begin{tabular}{c}$97.04$\\(4.0)\end{tabular}\\
\hline
MNIST w/500 clauses &  \begin{tabular}{c}$89.05$\\(1.0)\end{tabular} & \begin{tabular}{c}$94.28$\\(1.3)\end{tabular} & \begin{tabular}{c}$96.82$\\(2.1)\end{tabular} & \begin{tabular}{c}$97.50$\\(3.2)\end{tabular} & \begin{tabular}{c}$97.67$\\(4.5)\end{tabular} & \begin{tabular}{c}$97.74$\\(6.2)\end{tabular} & \begin{tabular}{c}$97.78$\\(7.8)\end{tabular} & \begin{tabular}{c}$97.77$\\(9.0)\end{tabular}\\\hline
MNIST w/1000 clauses & \begin{tabular}{c}$90.22$\\(1.0)\end{tabular} & \begin{tabular}{c}$95.26$\\(1.2)\end{tabular} & \begin{tabular}{c}$97.67$\\(2.6)\end{tabular} & \begin{tabular}{c}$98.03$\\(4.7)\end{tabular} & \begin{tabular}{c}$98.10$\\(8.4)\end{tabular} & \begin{tabular}{c}$98.14$\\(14.6)\end{tabular} & \begin{tabular}{c}$98.09$\\(22.8)\end{tabular} & \begin{tabular}{c}$98.05$\\(31.2)\end{tabular}\\ 
\hline
MNIST w/2000 clauses & \begin{tabular}{c}$90.88$\\(1.0)\end{tabular} & \begin{tabular}{c}$96.23$\\(1.2)\end{tabular} & \begin{tabular}{c}$98.01$\\(2.7)\end{tabular} & \begin{tabular}{c}$98.26$\\(5.3)\end{tabular} & \begin{tabular}{c}$98.26$\\(9.7)\end{tabular} & \begin{tabular}{c}$98.33$\\(17.5)\end{tabular} & \begin{tabular}{c}$98.22$\\(28.5)\end{tabular} & \begin{tabular}{c}$98.2$\\(40.1)\end{tabular} \\  
\hline
MNIST w/4000 clauses & \begin{tabular}{c}$91.68$\\(1.0)\end{tabular} & \begin{tabular}{c}$97.08$\\(1.3)\end{tabular} & \begin{tabular}{c}$98.19$\\(2.9)\end{tabular} & \begin{tabular}{c}$98.37$\\(5.6)\end{tabular} & \begin{tabular}{c}$98.4$\\(10.5)\end{tabular} & \begin{tabular}{c}$98.35$\\(19.1)\end{tabular} & \begin{tabular}{c}$98.32$\\(31.4)\end{tabular} & \begin{tabular}{c}$98.29$\\(45.2)\end{tabular} \\ 
\hline
MNIST w/8000 clauses& \begin{tabular}{c}$92.09$\\(1.0)\end{tabular}& \begin{tabular}{c}$97.42$\\(1.4)\end{tabular}&\begin{tabular}{c} $98.34$\\ (3.0)\end{tabular} &\begin{tabular}{c}$98.40$\\(5.8)\end{tabular} &\begin{tabular}{c} $98.42$\\(10.8)\end{tabular}&\begin{tabular}{c}
$98.38$\\ (19.9)\end{tabular}&\begin{tabular}{c}
$98.33$\\ (33.0)\end{tabular}&\begin{tabular}{c}
$98.32$\\ (47.7)\end{tabular} \\
\hline
\end{tabular}
\caption{Performance on MNIST with different number of total clauses, where each clause has a literal budget of 1, 2, 4, 8, 16, 32, 64, and all (no constraint) literals. Each field reports worst maximum accuracy across 5 independent runs, followed by, in brackets, the average number of literals used by the TM.}
\label{table_performance_clauses_mnist}
\end{table*}

\subsection{Natural Language Processing}
We first evaluate CSC-TM on five NLP datasets: BBC sports~\cite{greene06icml}, R8~\cite{debole2005analysis}, TREC-6~\cite{chang2002system}, SemEval 2010 Semantic Relations~\cite{hendrickx2009semeval}, and ACL Internet Movie Database (IMDb)~\cite{maas2011learning}. Starting from a maximal constraint of $1$, we progressively increase the literal constraint to $64$, recording the resulting accuracy and the average number of literals included per clause. For BBC Sports, we notice that a literal constraint of $8$ yields the maximum accuracy of $99.1 \%$. Similarly, the maximum accuracy for TREC and R8 are achieved with literal counts $32$ and $all$, respectively. We further observe that incorporating all the literals reduces the accuracy in BBC and TREC. For all the datasets, the average literal count drops significantly with literal budgeting. For instance, the literal constraint of $32$ for BBC Sports gives $2.64$ literals per clause on average, whereas the average is $44.14$ without constraints. Accordingly, the clause length is considerable shortened, and the clauses can be quickly glanced by humans for interpretation. Similar trends can also be seen for the other data sets. \footnote{For TM hyperparameters in BBC Sports, TREC, and R8, we use $8000$ clauses, a voting margin $T$ of $100$, and specificity $s$ of $10.0$}\par

\begin{figure}[ht]
\centering
\includegraphics[width=0.8\linewidth]{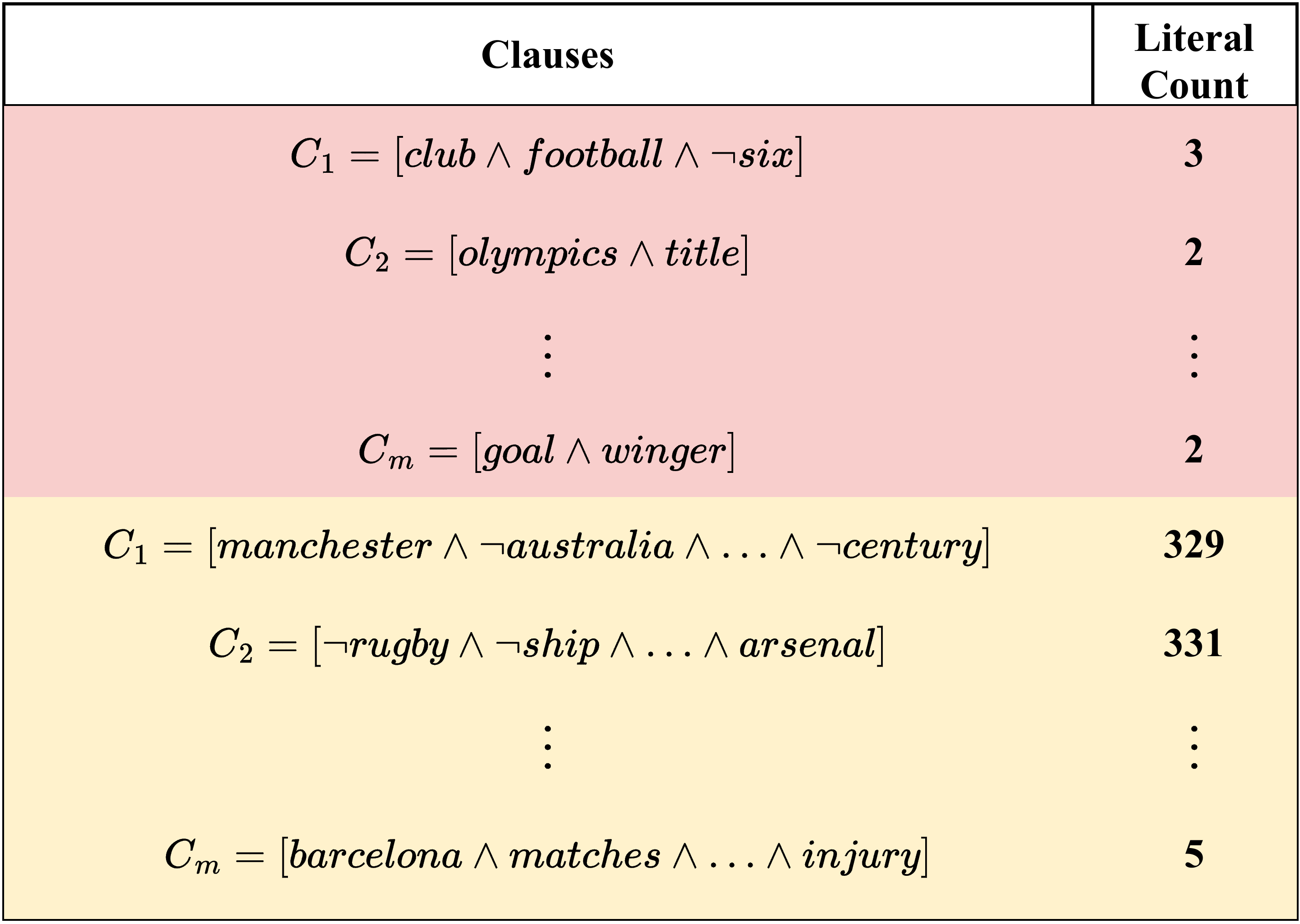}
\caption{Interpretability of clauses with constrained clause size for class ``football'' from BBC Sports. The rows highlighted in red are clauses from CLC-TM, while the yellow rows are from vanilla TM. }\label{figure:clause_interpretability}
\end{figure}

Consider as an example the results from R8 in Table \ref{table_performance}. Notice how a constraint as strict as $4$ still almost maximizes accuracy. Indeed, the NLP results overall show that CSC-TM allows us build concise and accurate propositional rules for better understandability. The improved interpretability is showcased in Figure~\ref{figure:clause_interpretability}, where we retrieve a few typical clauses from the ``Football'' class of BBC Sports. The literal-constrained clauses (in red) only contain 2-3 literals, and they clearly relate to the ``Football'' class. The vanilla clauses (in yellow), however, contain a much larger number of literals and rely extensively on feature negation.
\subsection{Image Processing}
We evaluate our approach on two image datasets: MNIST and CIFAR-2 (a two-class variant of CIFAR-10
that groups vehicle images and animal images into two separate classes).\footnote{As TM hyperparameters, we use $8000$ clauses per class, a voting margin $T$ of $10000$, and specificity $5.0$ in the MNIST experiments. For CIFAR-2, the number of clauses is $8000$, $T$ is $6000$, and $s$ is $10.0$.} For MNIST, we perform experiments utilizing both vanilla and convolutional TM with constrained clause length. Observe from Table~\ref{table_performance} how a constraint as small as $8$ still yields competitive accuracy. Specifically, the maximum accuracy is obtained in CIFAR-2, MNIST, and MNIST w/conv. with literal constrains $64$, $16$, and $16$, respectively. Also notice that CSC-TM on average keeps the number of literals per clause well below the set constraints. For example, for the latter constraints, the corresponding average number of literals per clause are respectively $34.1$, $10.8$, and $13.4$. Without clause length constraints, however, the corresponding average number of included literals are $60.4$, $47.7$, and $40.3$. Finally, notice how setting a too tight literal budget (below $4$) for MNIST w/conv. and CIFAR-2 increases the average number of literals used. This can be explained by CSC-TM not finding sufficiently accurate patterns. AS a result, it stays in literal exploration mode throughout the epochs. In conclusion, we observe that the maximum accuracy can be achieved using significantly fewer literals per clauses using CLC-TM. This allows us to significantly reduce computational complexity and increase the readability of the clauses.

To investigate how the number of clauses interact with the literal constraint, we now measure the effect of jointly varying the number of clauses and literal budget. From Table \ref{table_performance_clauses_mnist}, we observe a graceful degradation of accuracy as the number of clauses drops to $250$ and the literal budget falls to $1$. Also notice that fewer clauses produces fewer literals per clause on average. We believe this is the case because when fewer clauses are available, they must become less specialized to solve the task. However, when looking at attaining maximum accuracy, we observe that $1000$ clauses require more literals than  $4000$ ($32$ vs. $16$ literals on average per clause). The reason may be that fewer clauses needs to be more specific to compensate and maintain accuracy. As we increase the number of clauses, each clause includes fewer literals, solving the task as an ensemble.  In conclusion, CSC-TM allows a fine-grained trade-off between the length of clauses and the number of clauses.

\subsection{Self-Supervised Learning}
For the self-supervised learning task, we evaluate how the clause literal budget influences both training time and interpretability. Here, we evaluate the previously proposed Label-Critic TM (LCTM)~\cite{abouzeid2022label}, which is a novel architecture to self-learn data samples' labels without knowing the ground truths. The LCTM architecture runs on top of the standard CUDA TM implementation and starts by randomly initializing the data labels. Thereafter, it performs hierarchically clustering while learning the sub-patterns and their associated labels. Eventually, the learned sub-patterns represent interpretable clusters, each associated with a single supporting and a single discriminating clause. As a result, the method is interpretable, however, can still benefit from smaller clauses.

\begin{figure}[ht]
\centering
\includegraphics[width=.6\linewidth]{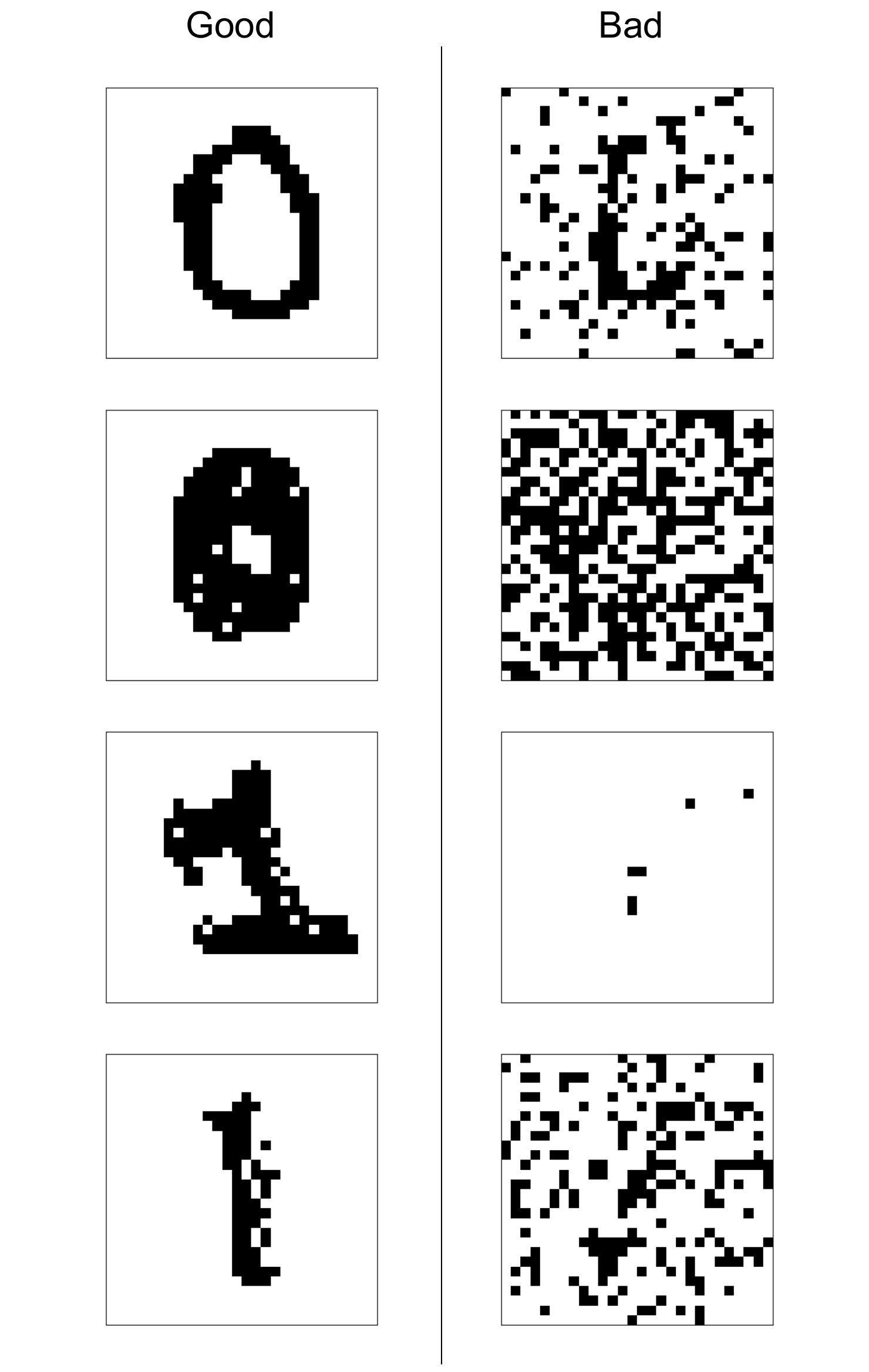}
\caption{Example of four clusters deemed human-interpretable (Good) and four clusters not being interpretable (Bad).}\label{lctm_sup_interp}

\end{figure}

Table \ref{table_lctm} shows the empirical results from different clause literal budgets on a subset of the MNIST dataset. Here, LCTM is to learn the labels and sub-patterns of the MNIST samples, associated with the digits ``One" and ``Zero". The results captures how the literal constraint influences both the training time and the interpretability. The interpretability metrics are as follows. \emph{Supporting interpretability} is the percentage of the positive polarity clauses that a human verifies as recognizable. See Figure \ref{lctm_sup_interp} for examples of clauses that are deemed interpretable and not interpretable.  Similarly, \emph{discriminating interpretability} is the percentage of negative polarity clauses that are recognized by humans. 
%Figures \ref{lctm_sup_interp} and \ref{lctm_disc_interp} explain how good and bad interpretations would occur by visualizing the LCTM learned clauses (supporting and discriminating) for the different learned sub-patterns. The visual clusters were generated by reshaping the clause literals to a ($28 \times 28$) indexed by the literal indices and the pixel position on the  board. Then, we converted the positive and negative clause literals to black and white pixels, respectively. Eventually, the non-existed indices (literals) were represented as white pixels.

\begin{table*}
\centering
\small
\begin{tabular}{|l|c|c|c|c|} 
\hline
\multicolumn{1}{|c|}{ \textbf{Literals \#} }  &\textbf{Supporting Interpretability (\%)} &\textbf{Discriminating Interpretability (\%)}  &\textbf{Clusters \#}    &\textbf{Speed up} \\ 
\hline
\hline
$400$ &84.92 $\scriptstyle\pm$ 5.21 & 32.17 $\scriptstyle\pm$ 6.34& 36 $\scriptstyle\pm$ 7.04  & $1.1$$\times$\\
\hline
$800$ & 85.48 $\scriptstyle\pm$ 7.80 &  33.02 $\scriptstyle\pm$ 7.21  &  35.8$ \scriptstyle\pm$ 6.76 & $1.3$ $\times$ \\
\hline
$1,200$ & 88.84 $\scriptstyle\pm$ 2.71 & 39.05 $\scriptstyle\pm$ 8.03  &  34.6 $\scriptstyle\pm$ 9.54 & $1.3$ $\times$\\
\hline
$1,568$ (all) & 86.51 $\scriptstyle\pm$ 1.33 & 33.63 $\scriptstyle\pm$ 7.59 &  33.8 $\scriptstyle\pm$ 7.34 & $1$ $\times$\\
\hline

\end{tabular}
\caption{LCTM performance on MNIST with different literal budgets. Mean and standard deviation are calculated over 5 independent runs.}
\label{table_lctm}
\end{table*}

%\begin{figure}[ht]
%\centering
%\includegraphics[width=.8\linewidth]{figures/lctm_discrimi_interp.pdf}
%\caption{Discriminating clauses interpretability quality example on four well discriminated cluster clauses, and four bad discriminated cluster clauses on the MNIST dataset.}\label{lctm_disc_interp}

%\end{figure}

As shown in Table \ref{table_lctm}, when the literal budget was reduced to $1,200$, the LCTM was both faster and produced more interpretable clauses. In conclusion, constraining the size of clauses yielded both increased interpretability and LCTM learning speed across all our evaluations.

%its low discriminating interpretability regardless the literal budget was utilized or not, (2) the improvement in training time and interpretability quality was not significant when applying literal budget. However, there was still a convenient performance when we reduced the literal budget. Therefore, we believe using a reduced literal budget would not sacrifice the performance of at least some of TM applications, while still achieving memory allocation reduction caused by an optimized literal budget at the same time.

\subsection{Regression}
We use the Energy Performance dataset to evaluate regression performance based on \cite{abeyrathna2020nonlinear}. The results of the experiment are reported in terms of Mean Average Error (MAE). In brief, Table~\ref{table_performance} shows that the MAE decreases, i.e. performance is better, as the literal budget is increased from 1 to 64. Again, the degradation of performance is graceful, and one can trade off clause size against MAE.

\subsection{Board Game Winner Prediction}

We here the Hex game as an example of reinforcement learning with CSC-TM, where the task is to predict the winner (value) of any given board configuration.  %Specifically, the Hex game  \cite{giri2022logic} is employed to evaluate how clause size constraint affects predicting the result of the game. To guide a Monte-Carlo tree search toward a winning move, one needs to assess intermediate board configurations.
To investigate how the 
prediction accuracy varies for CSC-TM, we compare the vanilla TM \cite{giri2022logic} with the CSC-TM for distinct literal budgets. The details of the experiment setup can be found in \cite{giri2022logic}. %, starting from 1 literal to all the literals. % present to predict the winner of the Hex game, given any board configuration.
Bottom row of Table~\ref{table_performance} summarises how the accuracy varies with the literal budget. We observe that a literal budget of $64$ reaches the maximum accuracy of $82.17\%$ against an accuracy of $81.43\%$ for all the literals. However, it is to be noted that the smaller literal budgets provide relatively poor accuracy. We believe this is the case because describing a Hex board configuration accurately generally requires information on a sufficient number of piece positions due to the nature of the game.
%\input{empirical_results_darshana}

\begin{comment}
\input{empirical_results_saeed}
\end{comment}

%\input{empirical_results_rupsa}

%\input{empirical_results_rohan}

\section{Hardware Complexity and Energy Consumption Analysis}

TM hardware accelerators will typically be implemented by either Field Programmable Gate Arrays~(FPGAs) or Application Specific Integrated Circuits~(ASICs). In most cases the dominating part of the energy consumption is related to the switching of digital circuits. It should be noted, however, that the static power consumption due to transistor leakage current for high performance processes can reach up to 30\% of total power~\cite{Dally}. With implementations in low power processes the static power consumption will be much less.

The dynamic power, $P$, consumed by a digital circuit with load capacitance $C$, operating frequency $f$, supply voltage $V_{S}$, and an activity factor $\alpha$ (transitions per clock cycle) is given by Eq.~(\ref{equation: swithcing power consumption})~\cite{Dally},
\begin{equation} \label{equation: swithcing power consumption}
\textstyle
P=0.5 \times C \times V_{S}^2 \times f \times \alpha.
\end{equation}\noindent
Limiting the number of literals will reduce the $\alpha$ value in several gates that implement the clause logic. Only those gates that process the included literals will switch and consume energy. The reduction in energy consumption of the clause logic can therefore roughly be estimated to $\frac{\bf{b}}{l_{ave}}$, where $\bf{b}$ is the clause size constraint and $l_{ave}$ is the average number of literals in the model without the length constraint. The exact savings will depend on the dataset.

% \begin{figure}[ht]
% \centering
% \includegraphics[width=0.7\linewidth]{figures/Clause_logic.pdf}
% \caption{Hardware implementation of clause logic based on a) the classical approach and b) multiplexers.}\label{figure:Clause_logic}
% \end{figure}

% Figure \ref{figure:Clause_logic_classical} shows two alternative solutions for hardware implementation of a clause. Both can be used for literal budgeting. In this example there are 272 literals, and the number of literals per clause is limited to four.

\begin{figure}[ht]
\centering
\includegraphics[width=0.5\linewidth]{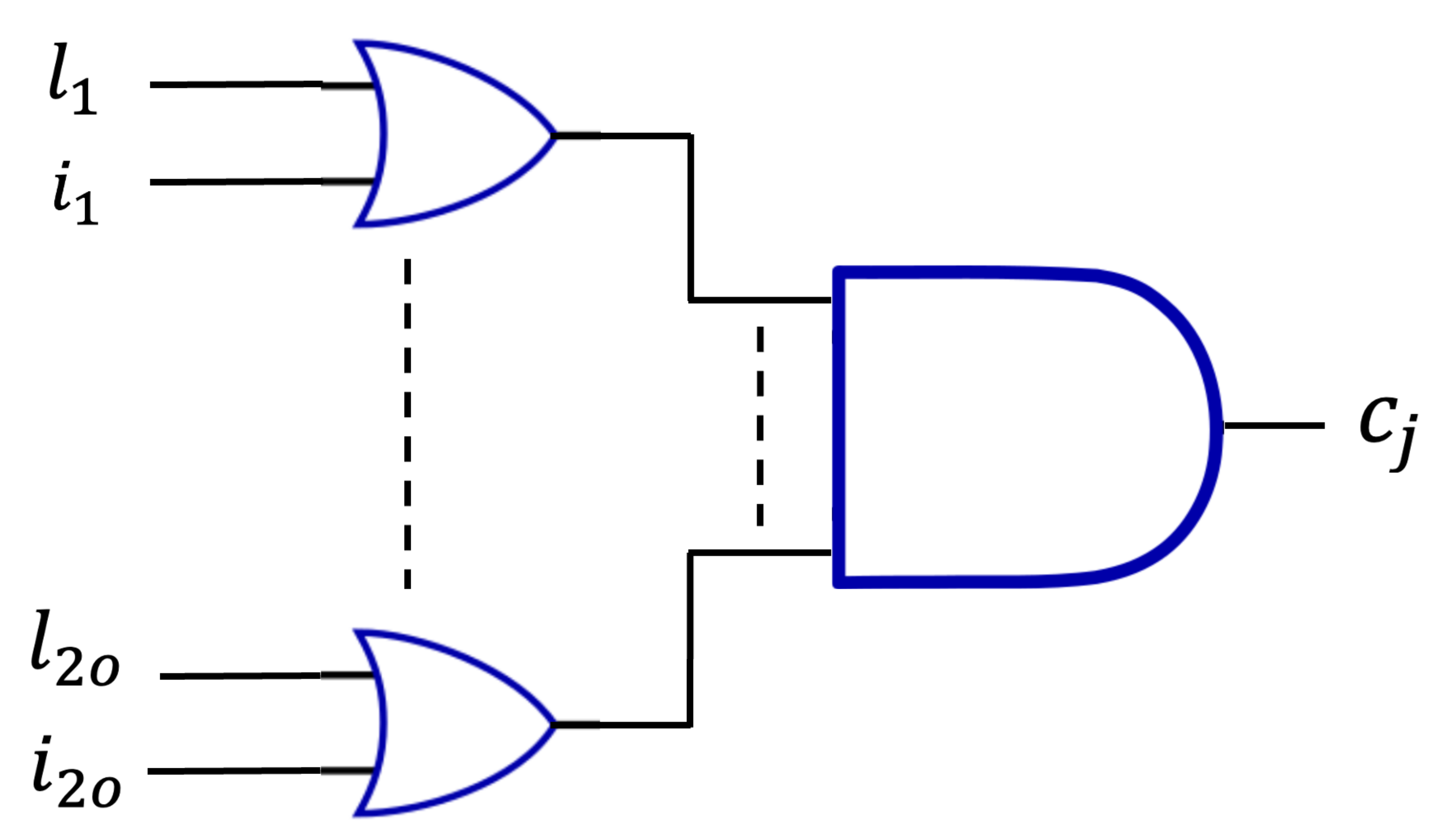}
\caption{Hardware implementation of clause logic.}\label{figure:Clause_logic_classical}
\end{figure}

The classical approach~\cite{wheeldon2020learning} for implementing clause logic is shown in Figure~\ref{figure:Clause_logic_classical}. 
Here each literal ${l_1, \ldots,l_{2o}}$ is either included or excluded by the associated include signal ${i_1, \ldots,i_{2o}}$ (active low) by using OR-gates. The include signals will typically all be simultaneously available from a register. The outputs from the OR-gates are then fed to a wide-input AND-gate, which will normally be implemented by several smaller AND-gates connected in a tree-structure to reduce path delay. Clearly, for a certain fixed application, given a smaller number of included literals, we can reduce the number of OR-gates to $\bf{b}$, and use an AND gate with only $\bf{b}$ inputs. In this way, the hardware complexity and power consumption can be reduced. 
For a general case, where the TM needs to be programmable in distinct applications, we need to have sufficient amount of available literals. Nevertheless, CSC-TM still has benefits in reduced switching activity, thus saving power.

It should be noted that it is only the energy consumption related to the clause logic that is affected by constraining clause size. The ensuing hardware processing, e.g., with clause weighting and summation is not affected. However, for systems with a huge number of clauses, the clause logic will occupy a significant part of the digital circuitry, and reducing its switching activity can enable significant energy savings.

An important system level benefit of literal budgeting is the time required for a model to be loaded from external or on-chip memory into registers in the ASIC or FPGA.  During this time the system's processor and the data transfer itself will consume energy. With less included literals, the model size and thus the load time can be reduced with a suitable encoding scheme, such as Run Length Encoding (RLE)~\cite{BakarRahman}. Reduction of the model size is also beneficial for other embedded TM solutions based on low-power microcontrollers.

% Finally, for submicron integracted circuit (IC) technologies, the static power consumption caused by leakage current can be significant. To reduce this one can use transistors with higher threshold voltages, at the expense of lower speed or higher supply voltage. To reduce the static energy consumption of the clause logic, one should use the solution that results in the smallest number of logical gates.  % Could be included in other sections if this fits better.

\section{Conclusions and Future Work}\label{sec:conclusion}
In this paper, we proposed CSC-\ac{TM} --- a novel \ac{TM} mechanism that constrains the size of clauses. We argued how limiting the number of literals significantly reduce switching activity in hardware, and thereby power consumption. We further analyzed and confirmed the convergence of CSC-TM.  Experimental results showed that CSC-TM can achieve the same or even better accuracy with shorter clauses, providing better interpretability. Future work includes introducing other kinds of constraints, with the intent of supporting constrained machine learning in general.

\bibliographystyle{named}
\bibliography{References}

\begin{thebibliography}{}

\bibitem[\protect\citeauthoryear{Abeyrathna \bgroup \em et al.\egroup
  }{2020a}]{abeyrathna2020deterministic}
K.~Darshana Abeyrathna, Ole-Christoffer Granmo, Rishad Shafik, Alex Yakovlev,
  Adrian Wheeldon, Jie Lei, and Morten Goodwin.
\newblock {A Novel Multi-Step Finite-State Automaton for Arbitrarily
  Deterministic Tsetlin Machine Learning}.
\newblock In {\em Lecture Notes in Computer Science: Proceedings of the 40th
  International Conference on Innovative Techniques and Applications of
  Artificial Intelligence (SGAI-2020)}. Springer International Publishing,
  2020.

\bibitem[\protect\citeauthoryear{{Abeyrathna} \bgroup \em et al.\egroup
  }{2020b}]{abeyrathna2020nonlinear}
K.~Darshana {Abeyrathna}, Ole-Christoffer {Granmo}, Xuan {Zhang}, Lei {Jiao},
  and Morten {Goodwin}.
\newblock {The Regression {T}setlin Machine - A Novel Approach to Interpretable
  Non-Linear Regression}.
\newblock {\em Philosophical Transactions of the Royal Society A}, 378, 2020.

\bibitem[\protect\citeauthoryear{Abeyrathna \bgroup \em et al.\egroup
  }{2020c}]{abeyrathna2020integer}
Kuruge~Darshana Abeyrathna, Ole-Christoffer Granmo, and Morten Goodwin.
\newblock {Extending the Tsetlin Machine With Integer-Weighted Clauses for
  Increased Interpretability}.
\newblock {\em arXiv preprint arXiv:2005.05131}, 2020.

\bibitem[\protect\citeauthoryear{Abeyrathna \bgroup \em et al.\egroup
  }{2021}]{abeyrathna2020massively}
K.~Darshana Abeyrathna, Bimal Bhattarai, Morten Goodwin, Saeed Gorji,
  Ole-Christoffer Granmo, Lei Jiao, Rupsa Saha, and Rohan~K. Yadav.
\newblock {Massively Parallel and Asynchronous Tsetlin Machine Architecture
  Supporting Almost Constant-Time Scaling}.
\newblock In {\em ICML}, 2021.

\bibitem[\protect\citeauthoryear{Abouzeid \bgroup \em et al.\egroup
  }{2022}]{abouzeid2022label}
Ahmed Abouzeid, Ole-Christoffer Granmo, Morten Goodwin, and Christian Webersik.
\newblock {Label-Critic Tsetlin Machine: A Novel Self-supervised Learning
  Scheme for Interpretable Clustering}.
\newblock In {\em International Symposium on the Tsetlin Machine (ISTM)}, pages
  41--48. IEEE, 2022.

\bibitem[\protect\citeauthoryear{Bakar \bgroup \em et al.\egroup
  }{2022a}]{BakarRahman}
A.~Bakar, T.~Rahman, A.~Montanari, J.~Lei, R.~Shafik, and F.~Kawsar.
\newblock {Logic-based Intelligence for Batteryless Sensors}.
\newblock In {\em the Annual International Workshop on Mobile Computing Systems
  and Applications (HotMobile)}, pages 22--28. Association for Computing
  Machinery, 2022.

\bibitem[\protect\citeauthoryear{Bakar \bgroup \em et al.\egroup
  }{2022b}]{Bakar2022}
Abu Bakar, Tousif Rahman, Rishad Shafik, Fahim Kawsar, and Alessandro
  Montanari.
\newblock {Adaptive Intelligence for Batteryless Sensors Using
  Software-Accelerated Tsetlin Machines}.
\newblock In {\em Proceedings of SenSys 2022}. ACM, 2022.

\bibitem[\protect\citeauthoryear{Bhattarai \bgroup \em et al.\egroup
  }{2022a}]{fakenewsLREC}
Bimal Bhattarai, Ole-Christoffer Granmo, and Lei Jiao.
\newblock {Explainable Tsetlin Machine Framework for Fake News Detection with
  Credibility Score Assessment}.
\newblock In {\em LREC}, 2022.

\bibitem[\protect\citeauthoryear{Bhattarai \bgroup \em et al.\egroup
  }{2022b}]{bhattarai2021word}
Bimal Bhattarai, Ole-Christoffer Granmo, and Lei Jiao.
\newblock {Word-level Human Interpretable Scoring Mechanism for Novel Text
  Detection Using Tsetlin Machines}.
\newblock {\em Applied Intelligence}, 2022.

\bibitem[\protect\citeauthoryear{Bhattarai \bgroup \em et al.\egroup
  }{2023}]{bhattaray2023embedding}
Bimal Bhattarai, Ole-Christoffer Granmo, Lei Jiao, Rohan Yadav, and Jivitesh
  Sharma.
\newblock {Tsetlin Machine Embedding: Representing Words Using Logical
  Expressions}.
\newblock {\em arXiv preprint arXiv:2301.00709}, 2023.

\bibitem[\protect\citeauthoryear{Borgersen \bgroup \em et al.\egroup
  }{2022}]{Borgersen2022Recommendation}
Karl Audun~Kagnes Borgersen, Morten Goodwin, and Jivitesh Sharma.
\newblock {A Comparison Between Tsetlin Machines and Deep Neural Networks in
  the Context of Recommendation Systems}.
\newblock {\em arXiv preprint arXiv:2212.10136}, 2022.

\bibitem[\protect\citeauthoryear{Chang \bgroup \em et al.\egroup
  }{2002}]{chang2002system}
Eric Chang, Frank Seide, Helen~M Meng, Zhuoran Chen, Yu~Shi, and Yuk-Chi Li.
\newblock {A System for Spoken Query Information Retrieval on Mobile Devices}.
\newblock {\em IEEE Transactions on Speech and Audio processing},
  10(8):531--541, 2002.

\bibitem[\protect\citeauthoryear{Dally \bgroup \em et al.\egroup
  }{2016}]{Dally}
William~J. Dally, Harting R.~Curtis, and Tor~M. Aamodt.
\newblock {\em {Digital Design Using VHDL: a Systems Approach}}.
\newblock Cambridge University Press, 2016.

\bibitem[\protect\citeauthoryear{Debole and
  Sebastiani}{2005}]{debole2005analysis}
Franca Debole and Fabrizio Sebastiani.
\newblock An analysis of the relative hardness of reuters-21578 subsets.
\newblock {\em Journal of the American Society for Information Science and
  technology}, 56(6):584--596, 2005.

\bibitem[\protect\citeauthoryear{Giri \bgroup \em et al.\egroup
  }{2022}]{giri2022logic}
Charul Giri, Ole-Christoffer Granmo, Herke Van~Hoof, and Christian~D. Blakely.
\newblock {Logic-based AI for Interpretable Board Game Winner Prediction with
  Tsetlin Machine}.
\newblock In {\em IJCNN}, pages 1--9, 2022.

\bibitem[\protect\citeauthoryear{Glimsdal and
  Granmo}{2021}]{glimsdal2021coalesced}
Sondre Glimsdal and Ole-Christoffer Granmo.
\newblock {Coalesced Multi-Output Tsetlin Machines with Clause Sharing}.
\newblock {\em arXiv preprint, arXiv:2108.07594}, 2021.

\bibitem[\protect\citeauthoryear{Glimsdal \bgroup \em et al.\egroup
  }{2022}]{Glimsdal2022}
Sondre Glimsdal, Rupsa Saha, Bimal Bhattarai, Charul Giri, Jivitesh Sharma,
  Svein~Anders Tunheim, and Rohan~Kumar Yadav.
\newblock {Focused Negative Sampling for Increased Discriminative Power in
  Tsetlin Machines}.
\newblock In {\em 2022 International Symposium on the Tsetlin Machine (ISTM)},
  pages 73--80, 2022.

\bibitem[\protect\citeauthoryear{{Granmo} \bgroup \em et al.\egroup
  }{2019}]{granmo2019convtsetlin}
Ole-Christoffer {Granmo}, Sondre {Glimsdal}, Lei {Jiao}, Morten {Goodwin},
  Christian~W. {Omlin}, and Geir~Thore {Berge}.
\newblock {The Convolutional {T}setlin Machine}.
\newblock {\em arXiv preprint arXiv:1905.09688}, 2019.

\bibitem[\protect\citeauthoryear{{Granmo}}{2018}]{granmo2018tsetlin}
Ole-Christoffer {Granmo}.
\newblock {The Tsetlin Machine - A Game Theoretic Bandit Driven Approach to
  Optimal Pattern Recognition with Propositional Logic}.
\newblock {\em arXiv preprint arXiv:1804.01508}, 2018.

\bibitem[\protect\citeauthoryear{Greene and Cunningham}{2006}]{greene06icml}
Derek Greene and P\'{a}draig Cunningham.
\newblock {Practical Solutions to the Problem of Diagonal Dominance in Kernel
  Document Clustering}.
\newblock In {\em ICML}, pages 377--384. ACM Press, 2006.

\bibitem[\protect\citeauthoryear{Hendrickx \bgroup \em et al.\egroup
  }{2009}]{hendrickx2009semeval}
Iris Hendrickx, Su~Nam Kim, Zornitsa Kozareva, Preslav Nakov, Diarmuid
  {\'O}~S{\'e}aghdha, Sebastian Pad{\'o}, Marco Pennacchiotti, Lorenza Romano,
  and Stan Szpakowicz.
\newblock Semeval-2010 task 8: Multi-way classification of semantic relations
  between pairs of nominals.
\newblock In {\em Proceedings of the Workshop on Semantic Evaluations: Recent
  Achievements and Future Directions}, pages 94--99. Association for
  Computational Linguistics, 2009.

\bibitem[\protect\citeauthoryear{Jiao \bgroup \em et al.\egroup
  }{2021}]{jiao2021convergenceAND}
Lei Jiao, Xuan Zhang, and Ole-Christoffer Granmo.
\newblock {On the Convergence of Tsetlin Machines for the AND and the OR
  Operators}.
\newblock {\em arXiv preprint https://arxiv.org/abs/2109.09488}, 2021.

\bibitem[\protect\citeauthoryear{Jiao \bgroup \em et al.\egroup }{Aug
  2022}]{jiao2021convergence}
Lei Jiao, Xuan Zhang, Ole-Christoffer Granmo, and K~Darshana Abeyrathna.
\newblock {On the Convergence of Tsetlin Machines for the XOR Operator}.
\newblock {\em IEEE Trans. Pattern Anal. Mach. Intell.}, Aug. 2022.

\bibitem[\protect\citeauthoryear{Lei \bgroup \em et al.\egroup
  }{2021}]{lei2021kws}
Jie Lei, Tousif Rahman, Rishad Shafik, Adrian Wheeldon, Alex Yakovlev,
  Ole-Christoffer Granmo, Fahim Kawsar, and Akhil Mathur.
\newblock {Low-Power Audio Keyword Spotting Using Tsetlin Machines}.
\newblock {\em Journal of Low Power Electronics and Applications}, 11, 2021.

\bibitem[\protect\citeauthoryear{Maas \bgroup \em et al.\egroup
  }{2011}]{maas2011learning}
Andrew Maas, Raymond~E Daly, Peter~T Pham, Dan Huang, Andrew~Y Ng, and
  Christopher Potts.
\newblock {Learning Word Vectors for Sentiment Analysis}.
\newblock In {\em ACL}, pages 142--150, 2011.

\bibitem[\protect\citeauthoryear{Saha \bgroup \em et al.\egroup
  }{2021}]{saha2021semantic}
Rupsa Saha, Ole-Christoffer Granmo, and Morten Goodwin.
\newblock {Using Tsetlin Machine to Discover Interpretable Rules in Natural
  Language Processing Applications}.
\newblock {\em Expert Systems}, page e12873, 2021.

\bibitem[\protect\citeauthoryear{Saha \bgroup \em et al.\egroup
  }{2022}]{Saha2022}
R.~Saha, O.-C. Granmo, V.I. Zadorozhny, and M.~Goodwin.
\newblock {A relational Tsetlin machine with applications to natural language
  understanding}.
\newblock {\em Journal of Intelligent Information Systems}, 2022.

\bibitem[\protect\citeauthoryear{{Seraj} \bgroup \em et al.\egroup
  }{2022}]{RaihanNIPS22}
Raihan {Seraj}, Jivitesh {Sharma}, and Ole~Christoffer Granmo.
\newblock {Tsetlin Machine for Solving Contextual Bandit Problems}.
\newblock In {\em NeurIPS}, 2022.

\bibitem[\protect\citeauthoryear{Sharma \bgroup \em et al.\egroup
  }{2023}]{dropclause}
Jivitesh Sharma, Rohan~Kumar Yadav, Ole-Christoffer~Granmo Granmo, and Lei
  Jiao.
\newblock Drop clause: Enhancing performance, robustness and pattern
  recognition capabilities of the tsetlin machine.
\newblock In {\em AAAI}, 2023.

\bibitem[\protect\citeauthoryear{Tsetlin}{1961}]{Tsetlin1961}
Michael~Lvovitch Tsetlin.
\newblock {On behaviour of finite automata in random medium}.
\newblock {\em Avtomat. i Telemekh}, 22(10):1345--1354, 1961.

\bibitem[\protect\citeauthoryear{Valiant}{1984}]{valiant1984learnable}
Leslie~G Valiant.
\newblock {A} {T}heory of the {L}earnable.
\newblock {\em Communications of the ACM}, 27(11):1134--1142, 1984.

\bibitem[\protect\citeauthoryear{{Wheeldon} \bgroup \em et al.\egroup
  }{2020}]{wheeldon2020learning}
Adrian {Wheeldon}, Rishad {Shafik}, Tousif {Rahman}, Jie {Lei}, Alex
  {Yakovlev}, and Ole-Christoffer {Granmo}.
\newblock {Learning Automata based Energy-efficient AI Hardware Design for
  IoT}.
\newblock {\em Philosophical Transactions of the Royal Society A}, 2020.

\bibitem[\protect\citeauthoryear{Yadav \bgroup \em et al.\egroup
  }{2021a}]{Rohanblackbox}
Rohan Yadav, Lei Jiao, Ole-Christoffer Granmo, and Morten Goodwin.
\newblock Enhancing interpretable clauses semantically using pretrained word
  representation.
\newblock In {\em the 4th BlackboxNLP Workshop on Analyzing and Interpreting
  Neural Networks for NLP}, 2021.

\bibitem[\protect\citeauthoryear{Yadav \bgroup \em et al.\egroup
  }{2021b}]{rohan2021AAAI}
Rohan Yadav, Lei Jiao, Ole-Christoffer Granmo, and Morten Goodwin.
\newblock {Human-Level Interpretable Learning for Aspect-Based Sentiment
  Analysis}.
\newblock In {\em AAAI}, 2021.

\bibitem[\protect\citeauthoryear{{Yadav} \bgroup \em et al.\egroup
  }{2022}]{yadav2022robustness}
Rohan~Kumar {Yadav}, Lei {Jiao}, Ole~Christoffer Granmo, and Morten {Goodwin}.
\newblock {Robust Interpretable Text Classification against Spurious
  Correlations Using AND-rules with Negation}.
\newblock In {\em IJCAI}, 2022.

\end{thebibliography}
\appendix

\section{Tsetlin Machine Basics}\label{app:basics}

\subsection{Classification}

%A \ac{TM} takes a vector $X=[x_1,\ldots,x_o]$ of Boolean features as input, to be classified into one of two classes, $y=0$ or $y=1$. Together with their negated counterparts, $\bar{x}_k = \lnot x_k = 1-x_k$, the features form a literal set $L = \{x_1,\ldots,x_o,\bar{x}_1,\ldots,\bar{x}_o\}$.
A \ac{TM} takes a vector $X=[x_1,\ldots,x_o]$ of Boolean features as input, to be classified into one of two classes, $y=0$ or $y=1$. The features are converted into the set of literals that consists of itself as well as its negated counterparts as $L = \{x_1,\ldots,x_o,\neg{x}_1,\ldots,\neg{x}_o\}$.

%A \ac{TM} pattern is formulated as a conjunctive clause $C_j$, formed by ANDing a subset $L_j \subseteq L$ of the literal set:
If there are $q$ number of classes and $n$ sub-patterns, TM pattern is formulated using \(q \times n\) conjunctive clauses, \(1 \leq j \leq n\) and is given by:
\begin{equation}
\textstyle
%C_j (X)=\bigwedge_{l_k \in L_j} l_k = \prod_{l_k \in L_j} l_k.
C_j (X)=\left(\bigwedge_{l_k \in L_j} l_k \right)\bigwedge \left(\bigwedge_{l_k \in {\bar{L}_j}} l_k \right),
\end{equation}
\noindent where \(L_j\) is the subset of the set of literals \(L\) which consists the original form of literals whereas \(\bar{L}_j\) is the subset of \(L\) that consists of the negated form of the literals. For example, the clause $C_j(X) = x_1 \land x_2 = x_1 x_2$ consists of the literals $L_j = \{x_1, x_2\}$, $\bar{L}_j = \{\neg x_1, \neg x_2\}$ and outputs $1$ if $x_1 = x_2 = 1$.

The number of clauses employed is a user set parameter $n$. Half of the clauses are assigned positive polarity which are odd indexed. The other half is assigned negative polarity which are odd indexed. The clause outputs are combined into a classification decision through summation and thresholding using the unit step function $u(v) = 1 ~\mathbf{if}~ v \ge 0 ~\mathbf{else}~ 0$:
\begin{equation}
\textstyle
\hat{y} = u\left(\sum_{j=1}^{n/2} C_j^+(X) - \sum_{j=1}^{n/2} C_j^-(X)\right).
\end{equation}
Namely, classification is performed based on a majority vote, with the positive clauses voting for $y=1$ and the negative for $y=0$. 
%The classifier $\hat{y} = u\left(x_1 \bar{x}_2 + \bar{x}_1 x_2 - x_1 x_2 - \bar{x}_1 \bar{x}_2\right)$, e.g., captures the XOR-relation.

\subsection{Learning}

A clause $C_j(X)$ is composed by a team of \acp{TA} \cite{Tsetlin1961}, each \ac{TA} deciding to \emph{Include} or \emph{Exclude} a specific literal $l_k$ in the clause. \ac{TA} makes decision based on the feedback it receives in the form of Reward, Inaction, and Penalty. There are two types of feedback associated with the learning of \ac{TM}: Type I Feedback and Type II Feedback, which is shown in Table \ref{table:type_i_feedback} and Table \ref{table:type_ii_feedback}. \par
\textbf{Type I feedback} is given stochastically to clauses with positive polarity when $y=1$ and to clauses with negative polarity when $y=0$. Each clause, in turn, reinforces its \acp{TA} based on: (1) its output $C_j(X)$; (2) the action of the \ac{TA} -- \emph{Include} or \emph{Exclude}; and (3) the value of the literal $l_k$ assigned to the \ac{TA}. Two rules govern Type I feedback:
\begin{itemize}
\item \emph{Include} is rewarded and \emph{Exclude} is penalized with probability $\frac{s-1}{s}~\mathbf{if}~C_j(X)=1~\mathbf{and}~l_k=1$. This reinforcement is strong (triggered with high probability) and makes the clause remember and refine the pattern it recognizes in $X$.\footnote{Note that the probability $\frac{s-1}{s}$ is replaced by $1$ when boosting true positives.} 
\item \emph{Include} is penalized and \emph{Exclude} is rewarded with probability $\frac{1}{s}~\mathbf{if}~C_j(X)=0~\mathbf{or}~l_k=0$. This reinforcement is weak (triggered with low probability) and coarsens infrequent patterns, making them frequent.
\end{itemize}
Above, parameter $s$ controls pattern frequency.

\textbf{Type II feedback} is given stochastically to clauses with positive polarity when $y\!=\!0$ and to clauses with negative polarity when $y\!=\!1$. It penalizes \emph{Exclude} with probability $1$ $\mathbf{if}~C_j(X)=1~\mathbf{and}~l_k=0$. Thus, this feedback produces literals for discriminating between $y=0$ and $y=1$.

\begin{table}[h!]
\centering
\begin{tabular}{c|ccccc}
\multicolumn{2}{r|}{\it Value of the clause $C^i_j(\bf{X})$ }&\multicolumn{2}{c}{\True}&\multicolumn{2}{c}{\False}\\ 
\multicolumn{2}{r|}{\it Value of the literal $x_k$/$\lnot x_k$}&{\True}&{\False}&{\True}&{\False}\\
\hline
\hline
\multirow{3}{*}{\begin{tabular}[c]{@{}c@{}}TA: \bf Include \\\bf Literal\end{tabular}}&\multicolumn{1}{c|}{$P(\mathrm{Reward})$}&$\frac{s-1}{s}$&NA&$0$&$0$\\
&\multicolumn{1}{c|}{$P(\mathrm{Inaction})$}&$\frac{1}{s}$&NA&$\frac{s-1}{s}$&$\frac{s-1}{s}$\\
&\multicolumn{1}{c|}{$P(\mathrm{Penalty})$}&$0$&NA&$\frac{1}{s} $&$\frac{1}{s}$\\
\hline
\multirow{3}{*}{\begin{tabular}[c]{@{}c@{}}TA: \bf Exclude \\\bf Literal\end{tabular}}&\multicolumn{1}{c|}{$P(\mathrm{Reward})$}&$0$&$\frac{1}{s}$&$\frac{1}{s}$ &$\frac{1}{s}$\\
&\multicolumn{1}{c|}{$P(\mathrm{Inaction})$}&$\frac{1}{s}$&$\frac{s-1}{s}$&$\frac{s-1}{s}$ &$\frac{s-1}{s}$\\
&\multicolumn{1}{c|}{$P(\mathrm{Penalty})$}&$\frac{s-1}{s}$&$0$&$0$&$0$\\
\hline
\end{tabular}
\caption{Type I Feedback for vanilla TM --- Feedback upon receiving a sample with label $y=1$, for a single TA to decide whether to Include or Exclude a given literal $x_k/\neg x_k$ into $C^i_j$. NA means not applicable.}
\label{table:type_i_feedback}
\end{table}

\begin{table}[h!]
\centering
\begin{tabular}{c|ccccc}
\multicolumn{2}{r|}{\it Value of the clause $C^i_j(\bf{X})$}&\multicolumn{2}{c}{\True}&\multicolumn{2}{c}{\False}\\ 
\multicolumn{2}{r|}{\it Value of the literal $x_k/\neg x_k$}&{\True}&{\False}&{\True}&{\False}\\
\hline
\hline
\multirow{3}{*}{\begin{tabular}[c]{@{}c@{}}TA: \bf Include \\\bf Literal\end{tabular}}&\multicolumn{1}{c|}{$P(\mathrm{Reward})$}&$0$&$\mathrm{NA}$&$0$&$0$\\
&\multicolumn{1}{c|}{$P(\mathrm{Inaction})$}&$1.0$&$\mathrm{NA}$&$1.0$&$1.0$\\
&\multicolumn{1}{c|}{$P(\mathrm{Penalty})$}&$0$&$\mathrm{NA}$&$0$&$0$\\
\hline
\multirow{3}{*}{\begin{tabular}[c]{@{}c@{}}TA: \bf Exclude \\\bf Literal\end{tabular}}&\multicolumn{1}{c|}{$P(\mathrm{Reward})$}&$0$&$0$&$0$&$0$\\
&\multicolumn{1}{c|}{$P(\mathrm{Inaction})$}&$1.0$&$0$&$1.0$ &$1.0$\\
&\multicolumn{1}{c|}{$P(\mathrm{Penalty})$}&$0$&$1.0$&$0$&$0$\\
\hline
\end{tabular}
\caption{Type II Feedback --- Feedback upon receiving a sample with label $y=0$, for a single TA to decide whether to Include or Exclude a given literal $x_k/\neg x_k$ into $C^i_j$. NA means not applicable.}
\label{table:type_ii_feedback}
\end{table}

\begin{figure}[ht]
\centering
\includegraphics[width=1\linewidth]{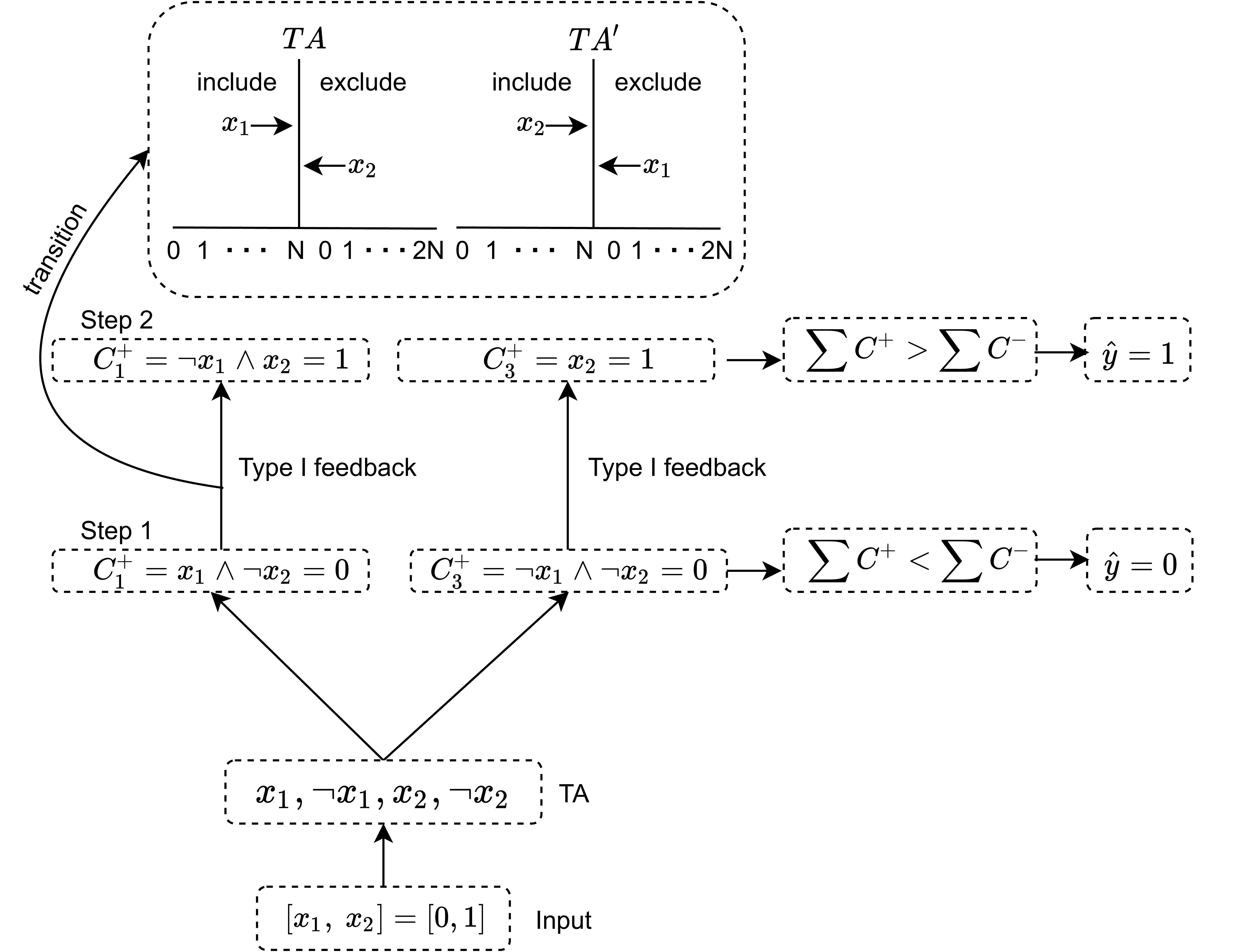}
\caption{The learning of Tsetlin Machine for a sample of XOR gate.}\label{figure:tm_architecture_basic}
\end{figure}

\par Let us consider a sample of XOR gate \((x_1 = 0, x_2 = 1) = 1\) to visualize the learning process as shown in Fig. \ref{figure:tm_architecture_basic}. There are $n$ clauses required to learn the XOR pattern and here let us consider $n=4$ per class. Among $4$ clauses, the clauses $C_1$ and $C_3$ votes for the presence $y=1$ and $C_0$ and $C_2$ votes against it. For simplification, let us only consider how $C_1$ and $C_3$ learns the pattern for the given sample of XOR gate. At step 1, the clauses has not learnt the pattern for given sample, which leads to wrong prediction of class thereby triggering Type I feedback for corresponding literals. From Table \ref{table:type_i_feedback} for literal $x_1$, if the clause score is $0$ and literal is $0$, it receives Inaction or Penalty for being included with the probability of $\frac{s-1}{s}$ and $\frac{1}{s}$ respectively. After several penalty, it changes its state to exclude action and gets removed from the clause $C_1$. On the other hand, the literal $\neg{x_1}$ gets penalty for being excluded and eventually jumps to include section as shown in $C_1$ at step 2. Similarly, when literal $\neg{x_2}=0$ and $C_1=0$, it receives Inaction or Penalty for being included with the probability of $\frac{s-1}{s}$ and $\frac{1}{s}$ respectively. After several penalties, $\neg{x_2}$ gets excluded and $x_2$ becomes included as shown in step 2.  This indeed reaches intended pattern thereby making the clauses $C_1=1$ and $C_3=1$, and finally results in $\hat{y}=1$. 

%A clause $C_j(X)$ is composed by a team of \acp{TA} \cite{Tsetlin1961}, each \ac{TA} deciding to \emph{Include} or \emph{Exclude} a specific literal $l_k$ in the clause. Learning which literals to include is based on reinforcement: Type I feedback produces frequent patterns, while Type II feedback increases the discrimination power of the patterns. \acp{TM} learn on-line, processing one training example $(X, y)$ at a time.

\textbf{Resource allocation} ensures that clauses distribute themselves across the frequent patterns, rather than missing some and over-concentrating on others. That is, for any input $X$, the probability of reinforcing a clause gradually drops to zero as the clause output sum
\begin{equation}\label{eqn:aggregated_votes}
\textstyle
    v = \sum_{j=1}^{n/2} C_j^+(X) - \sum_{j=1}^{n/2} C_j^-(X)
\end{equation}
approaches a user-configured target $T$ for $y=1$ (and $-T$ for $y=0$). 
If a clause is not reinforced, it does not give feedback to its \acp{TA}, and these are thus left unchanged.  In the extreme, when the voting sum $v$ equals or exceeds the target $T$ (the TM has successfully recognized the input $X$), no clauses are reinforced. They are then free to learn new patterns, naturally balancing the pattern representation resources  \cite{granmo2018tsetlin}.
\section{Detailed transition of a XOR sub-pattern given the new constraint}\label{app:proof}

This ection deals with the convergence of the XOR operator when only one literal is given, i.e.,  $(\|C^i_j(\bf{X})\| = 1)$.

\begin{table}
\centering
\begin{tabular}{ |c|c|c| } 
\hline
$x_1$ & $x_2$ & Output \\ 
0 & 0 & 0 \\ 
1 & 1 & 0 \\ 
0 & 1 & 1 \\
\hline
\end{tabular}
\caption{A sub-pattern in ``XOR'' case.}
\label{xorlogichalf2}
\end{table} 

Here we study the detailed transitions of TAs in a TM with length constraint, given the sub-patterns shown in Table \ref{xorlogichalf2} in XOR as input. Compared with the analysis in~\cite{jiao2021convergence}, the changes due to the new constraint are highlighted in red. 

Without loss of generality, we study clause $C_3$, which has
$\mathrm{TA}^3_{1}$ with actions ``Include" $x_{1}$ or ``Exclude" it, $\mathrm{TA}^3_{2}$ with actions ``Include" $\neg x_{1}$ or ``Exclude" it, $\mathrm{TA}^3_{3} $ with actions ``Include" $x_{2}$ or ``Exclude" it, and $\mathrm{TA}^3_{4}$ with actions ``Include" $\neg x_{2}$ or ``Exclude" it. To analyze the convergence of those four TAs, we perform a quasi-stationary analysis, where we freeze the behavior of three of them, and then study the transitions of the remaining one. More specifically, the analysis is organized as follows:
\begin{enumerate}
\item We freeze $\mathrm{TA}^3_{1}$ and $\mathrm{TA}^3_{2}$ respectively at ``Exclude" and ``Include". In this case, the first bit becomes $\neg x_{1}$. There are four sub-cases for $\mathrm{TA}^3_{3}$ and $\mathrm{TA}^3_{4}$: 
\begin{enumerate}
\item We study the transition of $\mathrm{TA}^3_{3}$ when it has the action ``Include" as its current action, given different training samples shown in Table \ref{xorlogichalf2} and different actions of $\mathrm{TA}^3_{4}$ (i.e., when the action of $\mathrm{TA}^3_{4}$ is frozen at ``Include" or ``Exclude"). \label {subcase1}
\item We study the transition of $\mathrm{TA}^3_{3}$ when it has ``Exclude" as its current action, given different training samples shown in Table \ref{xorlogichalf2} and different actions of $\mathrm{TA}^3_{4}$ (i.e., when the action of $\mathrm{TA}^3_{4}$ is frozen at ``Include" or ``Exclude"). \label {subcase2}
\item We study the transition of $\mathrm{TA}^3_{4}$ when it has ``Include" as its current action, given different training samples shown in Table \ref{xorlogichalf2} and different actions of $\mathrm{TA}^3_{3}$ (i.e., when the action of $\mathrm{TA}^3_{3}$ is frozen at ``Include" or ``Exclude"). \label {subcase3}
\item We study the transition of $\mathrm{TA}^3_{4}$ when it has ``Exclude" as its current action, given different training samples shown in Table \ref{xorlogichalf2} and different actions of $\mathrm{TA}^3_{3}$ (i.e., when the action of $\mathrm{TA}^3_{3}$ is frozen as ``Include" or ``Exclude"). \label {subcase4}
\end{enumerate}
%%%%%%%%%%%%%%%%%%%%%%%
\item We freeze $\mathrm{TA}^3_{1}$ and $\mathrm{TA}^3_{2}$ respectively at ``Include" and ``Exclude". In this case, the first bit becomes  $x_{1}$. The sub-cases for $\mathrm{TA}^3_{3}$ and $\mathrm{TA}^3_{4}$ are identical to the sub-cases in the previous case. 

%%%%%%%%%%%%%%%%%%%%%%%
\item We freeze $\mathrm{TA}^3_{1}$ and $\mathrm{TA}^3_{2}$ at ``Exclude" and ``Exclude". In this case, the first bit is excluded and will not influence the final output. 
The sub-cases for $\mathrm{TA}^3_{3}$ and $\mathrm{TA}^3_{4}$ are identical to the sub-cases in the previous case. 
%%%%%%%%%%%%%%%%%%%%%%%%%%%%%%%%
\item We freeze $\mathrm{TA}^3_{1}$ and $\mathrm{TA}^3_{2}$ at ``Include" and ``Include". In this case, we always have $C_3=0$ because the clause contains the contradiction $x_1 \land \lnot x_1$. The sub-cases for $\mathrm{TA}^3_{3}$ and $\mathrm{TA}^3_{4}$ are identical to the sub-cases in the previous case. 
%%%%%%%%%%%%%%%%%%%%%%%%%%%%%%%%%%%%%
\end{enumerate} 
In the analysis below, we will study each of the four cases, one by one.
%Note that the last bullet point is not interesting as the state that both $\mathrm{TA}^3_{1}$ and $\mathrm{TA}^3_{2}$ are Include is transient, which will always give $C_3= 0$ as output. Therefore, the first three cases are more important in terms of convergence analysis. 

%{\bf{Sub-Case \ref{subcase1}}}\\
%\subsubsection
\noindent {\bf Case 1} \\
We now analyze the first sub-case, i.e., Sub-case 1 (a). In this case, the first bit is in the form of $\neg x_1$ always. We here study the transition of $\mathrm{TA}^3_{3}$ when its current action is ``Include''. Depending on different training samples and actions of $\mathrm{TA}^3_{4}$, we have the following possible transitions. Below, ``I'' and ``E'' mean ``Include'' and ``Exclude'', respectively. For sake of conciseness, we remove the instances where no transition happens.

{\color{red}
\begin{minipage}{0.18\textwidth}
%$\Delta$ \textbf{Assume that:}\\
Condition: $x_{1}=0$, $x_{2}=1$, $y=1$, $\mathrm{TA}^3_{4}$=E.\\
Therefore, we have Type I feedback for
literal $x_{2}=1$, $C_{3}= \neg x_{1} \wedge x_{2} \wedge0 = 0$.\\ 
\end{minipage}
\hspace{0.35cm}\resizebox{0.17\textwidth}{!}{
\begin{minipage}{0.17\textwidth}
\begin{tikzpicture}[node distance = .3cm, font=\Huge]
\tikzstyle{every node}=[scale=0.3]
% NODES
\node[state] (E) at (0.5,1) {};
\node[state] (F) at (1.5,1) {};
\node[state] (G) at (2.5,1) {};
\node[state] (H) at (3.5,1) {};
\node[state] (A) at (0.5,2) {};
\node[state] (B) at (1.5,2) {};
\node[state] (C) at (2.5,2) {};
\node[state] (D) at (3.5,2) {};
% LETTERS
\node[thick] at (0,1) {$R$};
\node[thick] at (0,2) {$P$};
\node[thick] at (1.5,2.75) {$I$};
\node[thick] at (3.5,2.75) {$E$};
\draw[dotted, thick] (2,0.5) -- (2,2.5);
% ARROWS
\draw[every loop]
(A) edge[bend left] node [scale=1.2, above=0.1 of C] {} (B)
(B) edge[bend left] node [scale=1.2, above=0.1 of B] {\mbox{\Huge \boldmath$ ~~~~~~u_1\frac{1}{s}$}} (C);

\end{tikzpicture}
\end{minipage}
}\\
}

\begin{minipage}{0.18\textwidth}
Condition: $x_{1}=0$, $x_{2}=1$, $y=1$, $\mathrm{TA}^3_{4}$=I.\\
Therefore, we have Type I feedback for 
literal $x_{2}=1$, $C_{3}=0$.
\end{minipage}
\hspace{0.35cm}\resizebox{0.18\textwidth}{!}{
\begin{minipage}{0.18\textwidth}
\begin{tikzpicture}[node distance = .3cm, font=\Huge]
\tikzstyle{every node}=[scale=0.3]
% NODES
\node[state] (E) at (0.5,1) {};
\node[state] (F) at (1.5,1) {};
\node[state] (G) at (2.5,1) {};
\node[state] (H) at (3.5,1) {};
\node[state] (A) at (0.5,2) {};
\node[state] (B) at (1.5,2) {};
\node[state] (C) at (2.5,2) {};
\node[state] (D) at (3.5,2) {};
% LETTERS
\node[thick] at (0,1) {$R$};
\node[thick] at (0,2) {$P$};
\node[thick] at (1.5,2.75) {$I$};
\node[thick] at (3.5,2.75) {$E$};
\draw[dotted, thick] (2,0.5) -- (2,2.5);
% ARROWS
\draw[every loop]
(A) edge[bend left] node [scale=1.2, above=0.1 of C] {} (B)
(B) edge[bend left] node [scale=1.2, above=0.1 of B] {\mbox{\Huge \boldmath$ ~~~~~~u_1\frac{1}{s}$}} (C);

\end{tikzpicture}
\end{minipage}
}
\vspace{0.3cm}

We now consider Sub-case 1 (b). The literal $\neg x_1$ is still included, and we study the transition of $\mathrm{TA}^3_{3}$ when its current action is ``Exclude''. The possible transitions are listed below. 

%Study $\mathrm{TA}_{3,3}$ with action \textsl{exclude}

\begin{minipage}{0.18\textwidth}
Condition: $x_{1}=0$, $x_{2}=1$, $y=1$, $\mathrm{TA}^3_{4}$=E.\\
Therefore, Type I, $x_{2}=1$, $C_{3}=\neg x_{1}=1$. 
\end{minipage}
\hspace{0.35cm}\resizebox{0.18\textwidth}{!}{
\begin{minipage}{0.18\textwidth}
\begin{tikzpicture}[node distance = .3cm, font=\Huge]
\tikzstyle{every node}=[scale=0.3]
% NODES
\node[state] (E) at (0.5,1) {};
\node[state] (F) at (1.5,1) {};
\node[state] (G) at (2.5,1) {};
\node[state] (H) at (3.5,1) {};
\node[state] (A) at (0.5,2) {};
\node[state] (B) at (1.5,2) {};
\node[state] (C) at (2.5,2) {};
\node[state] (D) at (3.5,2) {};
% LETTERS
\node[thick] at (0,1) {$R$};
\node[thick] at (0,2) {$P$};
\node[thick] at (1.5,2.75) {$I$};
\node[thick] at (3.5,2.75) {$E$};
\draw[dotted, thick] (2,0.5) -- (2,2.5);
% ARROWS
\draw[every loop]
(D) edge[bend right] node [scale=1.2, above=0.1 of C] {} (C)
(C) edge[bend right] node [scale=1.2, above=0.1 of B] {\mbox{\Huge \boldmath$ ~~~~~~u_1\frac{1}{s}$}} (B);

\end{tikzpicture}
\end{minipage}
}

\begin{minipage}{0.18\textwidth}
Condition: $x_{1}=0$, $x_{2}=0$, $y=0$, $\mathrm{TA}^3_{4}$=E.\\
Therefore, Type II, $x_{2}=0$, $C_{3}=\neg x_{1}=1$. 
\end{minipage}
\hspace{0.35cm}\resizebox{0.18\textwidth}{!}{
\begin{minipage}{0.18\textwidth}
\begin{tikzpicture}[node distance = .3cm, font=\Huge]
\tikzstyle{every node}=[scale=0.3]
% NODES
\node[state] (E) at (0.5,1) {};
\node[state] (F) at (1.5,1) {};
\node[state] (G) at (2.5,1) {};
\node[state] (H) at (3.5,1) {};
\node[state] (A) at (0.5,2) {};
\node[state] (B) at (1.5,2) {};
\node[state] (C) at (2.5,2) {};
\node[state] (D) at (3.5,2) {};
% LETTERS
\node[thick] at (0,1) {$R$};
\node[thick] at (0,2) {$P$};
\node[thick] at (1.5,2.75) {$I$};
\node[thick] at (3.5,2.75) {$E$};
\draw[dotted, thick] (2,0.5) -- (2,2.5);
% ARROWS
\draw[every loop]
(D) edge[bend right] node [scale=1.2, above=0.1 of C] {\mbox{\Huge \boldmath$u_2\times 1$}} (C)
(C) edge[bend right] node [scale=1.2, above=0.1 of B] {} (B);

\end{tikzpicture}
\end{minipage}
}

\begin{minipage}{0.18\textwidth}
Condition: $x_{1}=0$, $x_{2}=1$, $y=1$,  $\mathrm{TA}^3_{4}$=I.\\
Therefore, Type I, $x_{2}=1$, 
$C_{3}=\neg x_{1} \wedge \neg x_{2}=0$.
\end{minipage}
\hspace{0.35cm}\resizebox{0.18\textwidth}{!}{
\begin{minipage}{0.18\textwidth}
\begin{tikzpicture}[node distance = .3cm, font=\Huge]
\tikzstyle{every node}=[scale=0.3]
% NODES
\node[state] (E) at (0.5,1) {};
\node[state] (F) at (1.5,1) {};
\node[state] (G) at (2.5,1) {};
\node[state] (H) at (3.5,1) {};
\node[state] (A) at (0.5,2) {};
\node[state] (B) at (1.5,2) {};
\node[state] (C) at (2.5,2) {};
\node[state] (D) at (3.5,2) {};
% LETTERS
\node[thick] at (0,1) {$R$};
\node[thick] at (0,2) {$P$};
\node[thick] at (1.5,2.75) {$I$};
\node[thick] at (3.5,2.75) {$E$};
\draw[dotted, thick] (2,0.5) -- (2,2.5);
% ARROWS
\draw[every loop]
(G) edge[bend left] node [scale=1.2, above=0.1 of C] {} (H)
(H) edge[loop right] node [scale=1.2, below=0.1 of H] {\mbox{\Huge \boldmath$ u_1\frac{1}{s}$}} (H);

\end{tikzpicture}
\end{minipage}
}

\begin{minipage}{0.18\textwidth}
Condition: $x_{1}=0$, $x_{2}=0$, $y=0$, $\mathrm{TA}^3_4$=I.\\
Therefore, Type II, $x_{2}=0$, 
$C_{3}=\neg x_{1} \wedge \neg x_{2}=1$.
\end{minipage}
\hspace{0.35cm}\resizebox{0.18\textwidth}{!}{
\begin{minipage}{0.18\textwidth}
\begin{tikzpicture}[node distance = .3cm, font=\Huge]
\tikzstyle{every node}=[scale=0.3]
% NODES
\node[state] (E) at (0.5,1) {};
\node[state] (F) at (1.5,1) {};
\node[state] (G) at (2.5,1) {};
\node[state] (H) at (3.5,1) {};
\node[state] (A) at (0.5,2) {};
\node[state] (B) at (1.5,2) {};
\node[state] (C) at (2.5,2) {};
\node[state] (D) at (3.5,2) {};
% LETTERS
\node[thick] at (0,1) {$R$};
\node[thick] at (0,2) {$P$};
\node[thick] at (1.5,2.75) {$I$};
\node[thick] at (3.5,2.75) {$E$};
\draw[dotted, thick] (2,0.5) -- (2,2.5);
% ARROWS
\draw[every loop]
(D) edge[bend right] node [scale=1.2, above=0.1 of C] {\mbox{\Huge \boldmath$ u_2\times 1$}} (C)
(C) edge[bend right] node [scale=1.2, above=0.1 of B] {\mbox{\Huge \boldmath$  $}} (B);

\end{tikzpicture}
\end{minipage}
}

Now let us move onto the third sub-case in Case 1, i.e., Sub-case 1 (c). The literal $\neg x_1$ is still included, and we study the transition of $\mathrm{TA}^3_{4}$ when its current action is ``Include''. Note that we are now studying $\mathrm{TA}^3_4$ that corresponds to $\neg x_{2}$ rather than $x_{2}$. Therefore, the literal in Tables \ref{table:type_i_feedback} and \ref{table:type_ii_feedback} becomes $\neg x_{2}$. 

%Study $\mathrm{TA}_{3,4}$ with \textsl{include}

\begin{minipage}{0.18\textwidth}
Condition: $x_{1}=0$, $x_{2}=1$, $y=1$, $\mathrm{TA}^3_{3}$=E. \\
Therefore, Type I, $\neg x_{2}=0$,
 $C_{3}=\neg x_{1} \wedge \neg x_{2}=0$.
\end{minipage}
\hspace{0.35cm}\resizebox{0.18\textwidth}{!}{
\begin{minipage}{0.18\textwidth}
\begin{tikzpicture}[node distance = .3cm, font=\Huge]
\tikzstyle{every node}=[scale=0.3]
% NODES
\node[state] (E) at (0.5,1) {};
\node[state] (F) at (1.5,1) {};
\node[state] (G) at (2.5,1) {};
\node[state] (H) at (3.5,1) {};
\node[state] (A) at (0.5,2) {};
\node[state] (B) at (1.5,2) {};
\node[state] (C) at (2.5,2) {};
\node[state] (D) at (3.5,2) {};
% LETTERS
\node[thick] at (0,1) {$R$};
\node[thick] at (0,2) {$P$};
\node[thick] at (1.5,2.75) {$I$};
\node[thick] at (3.5,2.75) {$E$};
\draw[dotted, thick] (2,0.5) -- (2,2.5);
% ARROWS
\draw[every loop]
(A) edge[bend left] node [scale=1.2, above=0.1 of C] {\mbox{\Huge \boldmath$ u_1\frac{1}{s}$}} (B)
(B) edge[bend left] node [scale=1.2, above=0.1 of B] {} (C);

\end{tikzpicture}
\end{minipage}
}

\begin{minipage}{0.18\textwidth}
Condition: $x_{1}=0$, $x_{2}=1$, $y=1$, $\mathrm{TA}^3_{3}$=I.\\
Therefore, Type I, $\neg x_{2}=0$,
$C_{3}=0$.
\end{minipage}
\hspace{0.35cm}\resizebox{0.18\textwidth}{!}{
\begin{minipage}{0.18\textwidth}
\begin{tikzpicture}[node distance = .3cm, font=\Huge]
\tikzstyle{every node}=[scale=0.3]
% NODES
\node[state] (E) at (0.5,1) {};
\node[state] (F) at (1.5,1) {};
\node[state] (G) at (2.5,1) {};
\node[state] (H) at (3.5,1) {};
\node[state] (A) at (0.5,2) {};
\node[state] (B) at (1.5,2) {};
\node[state] (C) at (2.5,2) {};
\node[state] (D) at (3.5,2) {};
% LETTERS
\node[thick] at (0,1) {$R$};
\node[thick] at (0,2) {$P$};
\node[thick] at (1.5,2.75) {$I$};
\node[thick] at (3.5,2.75) {$E$};
\draw[dotted, thick] (2,0.5) -- (2,2.5);
% ARROWS
\draw[every loop]
(A) edge[bend left] node [scale=1.2, above=0.1 of C] {} (B)
(B) edge[bend left] node [scale=1.2, above=0.1 of B] {\mbox{\Huge \boldmath$ ~~~~~~~u_1\frac{1}{s}$}} (C);

\end{tikzpicture}
\end{minipage}
}

For the Sub-case 1 (d), we study the transition of $\mathrm{TA}^3_{4}$ when it has the current action ``Exclude''. %To save space, we remove the scenarios where no transition happens. %Note that we are now studying $\mathrm{TA}^3_4$, the literal in Tables \ref{table:type_i_feedback} and \ref{table:type_ii_feedback} become $\neg x_{2}$. 
%Study $\mathrm{TA}_{3,4}$ with \textsl{exclude}

\begin{minipage}{0.18\textwidth}
Condition: $x_{1}=0$, $x_{2}=1$, $y=1$, $\mathrm{TA}^3_{3}$=E.\\
Therefore, Type I, $\neg x_{2}=0$,
$C_{3}=\neg x_{1}=1$.
\end{minipage}
\hspace{0.35cm}\resizebox{0.18\textwidth}{!}{
\begin{minipage}{0.18\textwidth}
\begin{tikzpicture}[node distance = .3cm, font=\Huge]
\tikzstyle{every node}=[scale=0.3]
% NODES
\node[state] (E) at (0.5,1) {};
\node[state] (F) at (1.5,1) {};
\node[state] (G) at (2.5,1) {};
\node[state] (H) at (3.5,1) {};
\node[state] (A) at (0.5,2) {};
\node[state] (B) at (1.5,2) {};
\node[state] (C) at (2.5,2) {};
\node[state] (D) at (3.5,2) {};
% LETTERS
\node[thick] at (0,1) {$R$};
\node[thick] at (0,2) {$P$};
\node[thick] at (1.5,2.75) {$I$};
\node[thick] at (3.5,2.75) {$E$};
\draw[dotted, thick] (2,0.5) -- (2,2.5);
% ARROWS
\draw[every loop]
(G) edge[bend left] node [scale=1.2, above=0.1 of C] {} (H)
(H) edge[loop right] node [scale=1.2, below=0.1 of H] {\mbox{\Huge \boldmath$ u_1\frac{1}{s}$}} (H);

\end{tikzpicture}
\end{minipage}
}

\begin{minipage}{0.18\textwidth}
Condition: $x_{1}=0$, $x_{2}=1$, $y=1$, $\mathrm{TA}^3_{3}$=I. \\
Therefore, Type I, $\neg x_{2}=0$,
{\color{red}$C_{3}=\neg x_{1} \wedge x_{2} \wedge 0=1$.}
\end{minipage}
\hspace{0.35cm}\resizebox{0.18\textwidth}{!}{
\begin{minipage}{0.18\textwidth}
\begin{tikzpicture}[node distance = .3cm, font=\Huge]
\tikzstyle{every node}=[scale=0.3]
% NODES
\node[state] (E) at (0.5,1) {};
\node[state] (F) at (1.5,1) {};
\node[state] (G) at (2.5,1) {};
\node[state] (H) at (3.5,1) {};
\node[state] (A) at (0.5,2) {};
\node[state] (B) at (1.5,2) {};
\node[state] (C) at (2.5,2) {};
\node[state] (D) at (3.5,2) {};
% LETTERS
\node[thick] at (0,1) {$R$};
\node[thick] at (0,2) {$P$};
\node[thick] at (1.5,2.75) {$I$};
\node[thick] at (3.5,2.75) {$E$};
\draw[dotted, thick] (2,0.5) -- (2,2.5);
% ARROWS
\draw[every loop]
(G) edge[bend left] node [scale=1.2, above=0.1 of C] {} (H)
(H) edge[loop right] node [scale=1.2, below=0.1 of H] {\mbox{\Huge \boldmath$ u_1\frac{1}{s}$}} (H);

\end{tikzpicture}
\end{minipage}
}

So far, we have gone through all sub-cases in Case 1. %We are now ready to sum up Case 1 by looking at the transitions of $\mathrm{TA}^3_{3}$ and $\mathrm{TA}^3_4$ in different scenarios. In this case, $\mathrm{TA}^3_{4}$ will become ``\textsl{Exclude}" in the long run because it has only one direction of transition, i.e., towards action ``Exclude''. Given $\mathrm{TA}^3_{4}$ is ``Exclude'', action ``Include'' of $\mathrm{TA}^3_{3}$ is an absorbing state. Therefore, if $\mathrm{TA}^3_{1}$ and $\mathrm{TA}^3_{2}$ are ``Exclude" and ``Include", respectively, $\mathrm{TA}^3_{3}$ will become ``Include", and $\mathrm{TA}^3_{4}$ will eventually be ``Exclude". In other words, $C_{3}$ will converge to $\neg x_{1} \wedge x_{2}$ in Case 1. \\

%%%%%%%%%%%%%%%%%%
%%%%%%%%%%%%%%%%%%%%%
%%%%%%%%%%%%%%%%%%%%%%
%%%%%%%%%%%%%%%%%%%%%%%
%%%%%%%%%%%%%%%%%%%%%Case 2
\noindent {\bf Case 2} \\
Case 2 studies the behavior of $\mathrm{TA}^3_3$ and $\mathrm{TA}^3_4$ when $\mathrm{TA}^3_1$ and $\mathrm{TA}^3_2$ select ``Include" and ``Exclude", respectively. In this case, the first bit is in the form of $x_1$ always.  There are here also four sub-cases and we will detail them presently. 

We first study $\mathrm{TA}^3_3$ with action ``Include", providing the below transitions. %Again, we remove all the ``No transition'' case. 

%Study $\mathrm{TA}_{3,3}$ with action \textsl{include}

\begin{minipage}{0.18\textwidth}
Conditions: $x_{1}=0$, $x_{2}=1$, ${y}=1$, $\mathrm{TA}^3_{4}$=E. \\
Therefore, Type I, $ x_{2} = 1$, $C_{3} = 0$.
\end{minipage}
\hspace{0.32cm}\resizebox{0.18\textwidth}{!}{
\begin{minipage}{0.18\textwidth}
\begin{tikzpicture}[node distance = .3cm, font=\Huge]
\tikzstyle{every node}=[scale=0.3]
% NODES
\node[state] (E) at (0.5,1) {};
\node[state] (F) at (1.5,1) {};
\node[state] (G) at (2.5,1) {};
\node[state] (H) at (3.5,1) {};
\node[state] (A) at (0.5,2) {};
\node[state] (B) at (1.5,2) {};
\node[state] (C) at (2.5,2) {};
\node[state] (D) at (3.5,2) {};
% LETTERS
\node[thick] at (0,1) {$R$};
\node[thick] at (0,2) {$P$};
\node[thick] at (1.5,2.75) {$I$};
\node[thick] at (3.5,2.75) {$E$};
\draw[dotted, thick] (2,0.5) -- (2,2.5);
% ARROWS
\draw[every loop]
(A) edge[bend left] node [scale=1.2, above=0.1 of C] {} (B)
(B) edge[bend left] node [scale=1.2, above=0.1 of C] {~~~~~\mbox{\Huge \boldmath$ u_1\frac{1}{s}$}} (C);

\end{tikzpicture}
\end{minipage}
}

\begin{minipage}{0.18\textwidth}
Conditions: $x_{1}=0$, $x_{2}=1$, $y=1$, $\mathrm{TA}^3_{4}$=I. \\
Therefore, Type I, $x_{2}=1$, $C_{3} = 0$. 
\end{minipage}
\hspace{0.32cm}\resizebox{0.18\textwidth}{!}{
\begin{minipage}{0.18\textwidth}
\begin{tikzpicture}[node distance = .3cm, font=\Huge]
\tikzstyle{every node}=[scale=0.3]
% NODES
\node[state] (E) at (0.5,1) {};
\node[state] (F) at (1.5,1) {};
\node[state] (G) at (2.5,1) {};
\node[state] (H) at (3.5,1) {};
\node[state] (A) at (0.5,2) {};
\node[state] (B) at (1.5,2) {};
\node[state] (C) at (2.5,2) {};
\node[state] (D) at (3.5,2) {};
% LETTERS
\node[thick] at (0,1) {$R$};
\node[thick] at (0,2) {$P$};
\node[thick] at (1.5,2.75) {$I$};
\node[thick] at (3.5,2.75) {$E$};
\draw[dotted, thick] (2,0.5) -- (2,2.5);
% ARROWS
\draw[every loop]
(A) edge[bend left] node [scale=1.2, above=0.1 of C] {} (B)
(B) edge[bend left] node [scale=1.2, above=0.1 of C] {~~~~~\mbox{\Huge \boldmath$ u_1\frac{1}{s}$}} (C);

\end{tikzpicture}
\end{minipage}
}

We then study $\mathrm{TA}^3_3$ with action ``Exclude", and transitions are shown below. %We remove all the ``No transition'' case. 

\begin{minipage}{0.18\textwidth}
Conditions: $x_{1}=0$, $x_{2}=1$, $y=1$, $\mathrm{TA}^3_{4}$=E. \\
Therefore, Type I, $x_{2}=1$, $C_3 = 0$.
\end{minipage}
\hspace{0.32cm}\resizebox{0.18\textwidth}{!}{
\begin{minipage}{0.18\textwidth}
\begin{tikzpicture}[node distance = .3cm, font=\Huge]
\tikzstyle{every node}=[scale=0.3]
% NODES
\node[state] (E) at (0.5,1) {};
\node[state] (F) at (1.5,1) {};
\node[state] (G) at (2.5,1) {};
\node[state] (H) at (3.5,1) {};
\node[state] (A) at (0.5,2) {};
\node[state] (B) at (1.5,2) {};
\node[state] (C) at (2.5,2) {};
\node[state] (D) at (3.5,2) {};
% LETTERS
\node[thick] at (0,1) {$R$};
\node[thick] at (0,2) {$P$};
\node[thick] at (1.5,2.75) {$I$};
\node[thick] at (3.5,2.75) {$E$};
\draw[dotted, thick] (2,0.5) -- (2,2.5);
% ARROWS
\draw[every loop]
(G) edge[bend left] node [scale=1.2, above=0.1 of C] {} (H)
(H) edge[loop right] node [scale=1.2, below=0.1 of H] {\mbox{\Huge \boldmath$ u_1\frac{1}{s}$}} (H);

\end{tikzpicture}
\end{minipage}}

\begin{minipage}{0.18\textwidth}
Conditions: $x_{1}=0$, $x_{2}=1$, $y=1$, $\mathrm{TA}^3_{4}$=I. \\
Therefore, Type I, $x_2=1$, $C_3 = 0$.
\end{minipage}
\hspace{0.32cm}\resizebox{0.18\textwidth}{!}{
\begin{minipage}{0.18\textwidth}
\begin{tikzpicture}[node distance = .3cm, font=\Huge]
\tikzstyle{every node}=[scale=0.3]
% NODES
\node[state] (E) at (0.5,1) {};
\node[state] (F) at (1.5,1) {};
\node[state] (G) at (2.5,1) {};
\node[state] (H) at (3.5,1) {};
\node[state] (A) at (0.5,2) {};
\node[state] (B) at (1.5,2) {};
\node[state] (C) at (2.5,2) {};
\node[state] (D) at (3.5,2) {};
% LETTERS
\node[thick] at (0,1) {$R$};
\node[thick] at (0,2) {$P$};
\node[thick] at (1.5,2.75) {$I$};
\node[thick] at (3.5,2.75) {$E$};
\draw[dotted, thick] (2,0.5) -- (2,2.5);
% ARROWS
\draw[every loop]
(G) edge[bend left] node [scale=1.2, above=0.1 of C] {} (H)
(H) edge[loop right] node [scale=1.2, below=0.1 of H] {\mbox{\Huge \boldmath$ u_1\frac{1}{s}$}} (H);

\end{tikzpicture}
\end{minipage}
}

We now study $\mathrm{TA}^3_4$ with action ``Include" and the transitions are presented below. %shown below. We remove all the ``No transition'' case.

\begin{minipage}{0.18\textwidth}
Condition: $x_{1}=0$, $x_{2}=1$, $y=1$, $\mathrm{TA}^3_{3}$=E. \\
Therefore, Type I, $\neg x_{2} = 0$, $C_3 =x_{1} \wedge \neg x_{2}= 0$.
\end{minipage}
\hspace{0.32cm}\resizebox{0.18\textwidth}{!}{
\begin{minipage}{0.18\textwidth}
\begin{tikzpicture}[node distance = .3cm, font=\Huge]
\tikzstyle{every node}=[scale=0.3]
% NODES
\node[state] (E) at (0.5,1) {};
\node[state] (F) at (1.5,1) {};
\node[state] (G) at (2.5,1) {};
\node[state] (H) at (3.5,1) {};
\node[state] (A) at (0.5,2) {};
\node[state] (B) at (1.5,2) {};
\node[state] (C) at (2.5,2) {};
\node[state] (D) at (3.5,2) {};
% LETTERS
\node[thick] at (0,1) {$R$};
\node[thick] at (0,2) {$P$};
\node[thick] at (1.5,2.75) {$I$};
\node[thick] at (3.5,2.75) {$E$};
\draw[dotted, thick] (2,0.5) -- (2,2.5);
% ARROWS
\draw[every loop]
(A) edge[bend left] node [scale=1.2, above=0.1 of C] {} (B)
(B) edge[bend left] node [scale=1.2, above=0.1 of C] {~~~~~\mbox{\Huge \boldmath$ u_1\frac{1}{s}$}} (C);

\end{tikzpicture}
\end{minipage}
}

\begin{minipage}{0.18\textwidth}
Conditions: $x_{1}=0$, $x_{2}=1$, $y=1$, $\mathrm{TA}^3_{3}$=I. \\
Therefore, Type I, $\neg x_{2} = 0$ , $ C_3= 0$.
\end{minipage}
\hspace{0.32cm}\resizebox{0.18\textwidth}{!}{
\begin{minipage}{0.18\textwidth}
\begin{tikzpicture}[node distance = .3cm, font=\Huge]
\tikzstyle{every node}=[scale=0.3]
% NODES
\node[state] (E) at (0.5,1) {};
\node[state] (F) at (1.5,1) {};
\node[state] (G) at (2.5,1) {};
\node[state] (H) at (3.5,1) {};
\node[state] (A) at (0.5,2) {};
\node[state] (B) at (1.5,2) {};
\node[state] (C) at (2.5,2) {};
\node[state] (D) at (3.5,2) {};
% LETTERS
\node[thick] at (0,1) {$R$};
\node[thick] at (0,2) {$P$};
\node[thick] at (1.5,2.75) {$I$};
\node[thick] at (3.5,2.75) {$E$};
\draw[dotted, thick] (2,0.5) -- (2,2.5);
% ARROWS
\draw[every loop]
(A) edge[bend left] node [scale=1.2, above=0.1 of C] {} (B)
(B) edge[bend left] node [scale=1.2, above=0.1 of C] {~~~~~\mbox{\Huge \boldmath$ u_1\frac{1}{s}$}} (C);

\end{tikzpicture}
\end{minipage}
}

We study lastly $\mathrm{TA}^3_4$ with action ``Exclude", leading to the following transitions. %We remove all the ``No transition'' case.

\begin{minipage}{0.18\textwidth}
Conditions: $x_{1}=1$, $x_{2}=1$, $y=0$, $\mathrm{TA}^3_{3}$=E. \\
Therefore, Type II, $\neg x_{2} = 0$, $C_3 =x_{1}= 1$.
\end{minipage}
\hspace{0.32cm}\resizebox{0.18\textwidth}{!}{
\begin{minipage}{0.18\textwidth}
\begin{tikzpicture}[node distance = .3cm, font=\Huge]
\tikzstyle{every node}=[scale=0.3]
% NODES
\node[state] (E) at (0.5,1) {};
\node[state] (F) at (1.5,1) {};
\node[state] (G) at (2.5,1) {};
\node[state] (H) at (3.5,1) {};
\node[state] (A) at (0.5,2) {};
\node[state] (B) at (1.5,2) {};
\node[state] (C) at (2.5,2) {};
\node[state] (D) at (3.5,2) {};
% LETTERS
\node[thick] at (0,1) {$R$};
\node[thick] at (0,2) {$P$};
\node[thick] at (1.5,2.75) {$I$};
\node[thick] at (3.5,2.75) {$E$};
\draw[dotted, thick] (2,0.5) -- (2,2.5);
% ARROWS
\draw[every loop]
(D) edge[bend right] node [scale=1.2, above=0.1 of C] {\mbox{\Huge \boldmath$ u_2\times1$}} (C)
(C) edge[bend right] node [scale=1.2, above=0.1 of B] {} (B);

\end{tikzpicture}
\end{minipage}
}

\begin{minipage}{0.18\textwidth}
Conditions: $x_{1}=0$, $x_{2}=1$, $y=1$, $\mathrm{TA}^3_{3}$=E. \\
Therefore, Type I, $\neg x_{2}=0$, $C_3 =0$.
\end{minipage}
\hspace{0.32cm}\resizebox{0.18\textwidth}{!}{
\begin{minipage}{0.18\textwidth}
\begin{tikzpicture}[node distance = .3cm, font=\Huge]
\tikzstyle{every node}=[scale=0.3]
% NODES
\node[state] (E) at (0.5,1) {};
\node[state] (F) at (1.5,1) {};
\node[state] (G) at (2.5,1) {};
\node[state] (H) at (3.5,1) {};
\node[state] (A) at (0.5,2) {};
\node[state] (B) at (1.5,2) {};
\node[state] (C) at (2.5,2) {};
\node[state] (D) at (3.5,2) {};
% LETTERS
\node[thick] at (0,1) {$R$};
\node[thick] at (0,2) {$P$};
\node[thick] at (1.5,2.75) {$I$};
\node[thick] at (3.5,2.75) {$E$};
\draw[dotted, thick] (2,0.5) -- (2,2.5);
% ARROWS
\draw[every loop]
(G) edge[bend left] node [scale=1.2, above=0.1 of C] {} (H)
(H) edge[loop right] node [scale=1.2, below=0.1 of H] {\mbox{\Huge \boldmath$ u_1\frac{1}{s}$}} (H);

\end{tikzpicture}
\end{minipage}
}

\begin{minipage}{0.18\textwidth}
Conditions: $x_{1}=1$, $x_{2}=1$, $y=0$, $\mathrm{TA}^3_{3}$=I. \\
Therefore, Type II, $\neg x_{2} = 0$,
\\$C_3 =x_{1} \wedge x_{2}= 1$. 
\end{minipage}
\hspace{0.32cm}\resizebox{0.18\textwidth}{!}{
\begin{minipage}{0.18\textwidth}
\begin{tikzpicture}[node distance = .3cm, font=\Huge]
\tikzstyle{every node}=[scale=0.3]
% NODES
\node[state] (E) at (0.5,1) {};
\node[state] (F) at (1.5,1) {};
\node[state] (G) at (2.5,1) {};
\node[state] (H) at (3.5,1) {};
\node[state] (A) at (0.5,2) {};
\node[state] (B) at (1.5,2) {};
\node[state] (C) at (2.5,2) {};
\node[state] (D) at (3.5,2) {};
% LETTERS
\node[thick] at (0,1) {$R$};
\node[thick] at (0,2) {$P$};
\node[thick] at (1.5,2.75) {$I$};
\node[thick] at (3.5,2.75) {$E$};
\draw[dotted, thick] (2,0.5) -- (2,2.5);
% ARROWS
%\draw[every loop]
%(H) edge[bend right] node [scale=1.2, above=0.1 of C] {\mbox{\Huge \boldmath$ u_2\times1$}} (G)
%(G) edge[bend right] node [scale=1.2, below=0.1 of B] { } (F);
\draw[every loop]
(D) edge[bend right] node [scale=1.2, above=0.1 of C] {\mbox{\Huge \boldmath$ u_2\times1$}} (C)
(C) edge[bend right] node [scale=1.2, above=0.1 of B] {} (B);

\end{tikzpicture}
\end{minipage}
}

\begin{minipage}{0.18\textwidth}
Conditions: $x_{1}=0$, $x_{2}=1$, $y=1$, $\mathrm{TA}^3_{3}$=I. \\
Therefore, Type I, $\neg x_{2} = 0$, $C_3 =x_{1} \wedge x_{2}= 0$.
\end{minipage}
\hspace{0.32cm}\resizebox{0.18\textwidth}{!}{
\begin{minipage}{0.18\textwidth}
\begin{tikzpicture}[node distance = .3cm, font=\Huge]
\tikzstyle{every node}=[scale=0.3]
% NODES
\node[state] (E) at (0.5,1) {};
\node[state] (F) at (1.5,1) {};
\node[state] (G) at (2.5,1) {};
\node[state] (H) at (3.5,1) {};
\node[state] (A) at (0.5,2) {};
\node[state] (B) at (1.5,2) {};
\node[state] (C) at (2.5,2) {};
\node[state] (D) at (3.5,2) {};
% LETTERS
\node[thick] at (0,1) {$R$};
\node[thick] at (0,2) {$P$};
\node[thick] at (1.5,2.75) {$I$};
\node[thick] at (3.5,2.75) {$E$};
\draw[dotted, thick] (2,0.5) -- (2,2.5);
% ARROWS
\draw[every loop]
(G) edge[bend left]node [scale=1.2, above=0.1 of C] {} (H)
(H) edge[loop right] node [scale=1.2, below=0.1 of H] {\mbox{\Huge \boldmath$ u_1\frac{1}{s}$}} (H);

\end{tikzpicture}
\end{minipage}
}

%To sum up Case 2, we understand that $\mathrm{TA}^3_{3}$ will select ``Exclude", and $\mathrm{TA}^3_{4}$ will switch between ``Include" or ``Exclude", depending on the training samples and system status.\\

%%%%%%%%%%%%%
%%%%%%%%%%%%%%
%%%%%%%%%%%%%
%%%%%%%%%%%%Case 3

\noindent {\bf Case 3} \\
Now we move onto Case 3, where $\mathrm{TA}^3_{1}$ and $\mathrm{TA}^3_{2}$ both select ``Exclude". We study the behavior of $\mathrm{TA}^3_{3}$ and $\mathrm{TA}^3_{4}$ for different sub-cases. In this case, the first bit $x_1$ does not play any role for the output. 

We first examine $\mathrm{TA}^3_3$ with action ``Include'', providing the transitions below. %We remove all the ``No transition'' case.

\begin{minipage}{0.18\textwidth}
{Conditions}: $x_{1}=0$, $x_{2}=1$, $y=1$, $\mathrm{TA}^3_4$=E. \\
Therefore, Type I, $x_{2} = 1$, $C_{3}=x_{2} = 1$.
\end{minipage}
\hspace{0.32cm}\resizebox{0.18\textwidth}{!}{
\begin{minipage}{0.18\textwidth}
\begin{tikzpicture}[node distance = .3cm, font=\Huge]
\tikzstyle{every node}=[scale=0.3]
% NODES
\node[state] (E) at (0.5,1) {};
\node[state] (F) at (1.5,1) {};
\node[state] (G) at (2.5,1) {};
\node[state] (H) at (3.5,1) {};
\node[state] (A) at (0.5,2) {};
\node[state] (B) at (1.5,2) {};
\node[state] (C) at (2.5,2) {};
\node[state] (D) at (3.5,2) {};
% LETTERS
\node[thick] at (0,1) {$R$};
\node[thick] at (0,2) {$P$};
\node[thick] at (1.5,2.75) {$I$};
\node[thick] at (3.5,2.75) {$E$};
\draw[dotted, thick] (2,0.5) -- (2,2.5);
% ARROWS
% ARROWS
\draw[every loop]
(F) edge[bend left] node [scale=1.2, above=0.1 of C] {} (E)
(E) edge[loop left = 45] node [scale=1.2, below=0.1 of E] {\mbox{\Huge \boldmath$ u_1\frac{s-1}{s}$}} (E);

\end{tikzpicture}
\end{minipage}
}

\begin{minipage}{0.18\textwidth}
Conditions: $x_{1}=0$, $x_{2}=1$, $y=1$, $\mathrm{TA}^3_{4}$=I. \\
Therefore, Type I, $x_{2} = 1$, $C_3= 0$. 
\end{minipage}
\hspace{0.32cm}\resizebox{0.18\textwidth}{!}{
\begin{minipage}{0.18\textwidth}
\begin{tikzpicture}[node distance = .3cm, font=\Huge]
\tikzstyle{every node}=[scale=0.3]
% NODES
\node[state] (E) at (0.5,1) {};
\node[state] (F) at (1.5,1) {};
\node[state] (G) at (2.5,1) {};
\node[state] (H) at (3.5,1) {};
\node[state] (A) at (0.5,2) {};
\node[state] (B) at (1.5,2) {};
\node[state] (C) at (2.5,2) {};
\node[state] (D) at (3.5,2) {};
% LETTERS
\node[thick] at (0,1) {$R$};
\node[thick] at (0,2) {$P$};
\node[thick] at (1.5,2.75) {$I$};
\node[thick] at (3.5,2.75) {$E$};
\draw[dotted, thick] (2,0.5) -- (2,2.5);
% ARROWS
\draw[every loop]
(A) edge[bend left] node [scale=1.2, above=0.1 of C] {} (B)
(B) edge[bend left] node [scale=1.2, above=0.1 of C] {~~~~~\mbox{\Huge \boldmath$ u_1\frac{1}{s}$}} (C);

\end{tikzpicture}
\end{minipage}
}

We then study $\mathrm{TA}^3_3$ with action ``Exclude'', transitions shown below. %$We remove all the ``No transition'' case. 
In this situation, if $\mathrm{TA}^3_{4}$ is also excluded, $C_3$ is ``empty'' since all its associated TA select action ``Exclude''. To make the training proceed, according to the training rule of TM, we assign $C_3=1$ in this situation. %Therefore, we only focus for $T_{3,4}$ is included. 

\begin{minipage}{0.18\textwidth}
{Condition:} $x_{1}=0$, $x_{2}=1$, $y=1$, $\mathrm{TA}^3_{4}$=E. \\
Therefore, Type I, $x_{2} = 1$, $C_{3}=1$.
\end{minipage}
\hspace{0.32cm}\resizebox{0.18\textwidth}{!}{
\begin{minipage}{0.18\textwidth}
\begin{tikzpicture}[node distance = .3cm, font=\Huge]
\tikzstyle{every node}=[scale=0.3]
% NODES
\node[state] (E) at (0.5,1) {};
\node[state] (F) at (1.5,1) {};
\node[state] (G) at (2.5,1) {};
\node[state] (H) at (3.5,1) {};
\node[state] (A) at (0.5,2) {};
\node[state] (B) at (1.5,2) {};
\node[state] (C) at (2.5,2) {};
\node[state] (D) at (3.5,2) {};
% LETTERS
\node[thick] at (0,1) {$R$};
\node[thick] at (0,2) {$P$};
\node[thick] at (1.5,2.85) {$I$};
\node[thick] at (3.5,2.85) {$E$};
\draw[dotted, thick] (2,0.5) -- (2,2.5);
% ARROWS
\draw[every loop]
(D) edge[bend right] node [scale=1.2, above=0.1 of C] {\mbox{\Huge \boldmath$ u_1\frac{s-1}{s}$}} (C)
(C) edge[bend right] node [scale=1.2, above=0.1 of C] {} (B);

\end{tikzpicture}
\end{minipage}
}

\begin{minipage}{0.18\textwidth}
{Condition:} $x_{1}=0$, $x_{2}=0$, $y=0$, $\mathrm{TA}^3_{4}$=E. \\
Therefore, Type II, $x_{2} = 0$, $C_{3}=1$.
\end{minipage}
\hspace{0.32cm}\resizebox{0.18\textwidth}{!}{
\begin{minipage}{0.18\textwidth}
\begin{tikzpicture}[node distance = .3cm, font=\Huge]
\tikzstyle{every node}=[scale=0.3]
% NODES
\node[state] (E) at (0.5,1) {};
\node[state] (F) at (1.5,1) {};
\node[state] (G) at (2.5,1) {};
\node[state] (H) at (3.5,1) {};
\node[state] (A) at (0.5,2) {};
\node[state] (B) at (1.5,2) {};
\node[state] (C) at (2.5,2) {};
\node[state] (D) at (3.5,2) {};
% LETTERS
\node[thick] at (0,1) {$R$};
\node[thick] at (0,2) {$P$};
\node[thick] at (1.5,2.75) {$I$};
\node[thick] at (3.5,2.75) {$E$};
\draw[dotted, thick] (2,0.5) -- (2,2.5);
% ARROWS
\draw[every loop]
(D) edge[bend right] node [scale=1.2, above=0.1 of C] {\mbox{\Huge \boldmath$ u_2\times 1$}} (C)
(C) edge[bend right] node [scale=1.2, above=0.1 of C] {} (B);

\end{tikzpicture}
\end{minipage}
}

\begin{minipage}{0.18\textwidth}
Condition: $x_{1}=0$, $x_{2}=1$, $y=1$, $\mathrm{TA}^3_{4}$=I. \\
Therefore, Type I, $x_{2} = 1$,  $C_{3}=\neg x_{2} = 0$.
\end{minipage}
\hspace{0.32cm}\resizebox{0.18\textwidth}{!}{
\begin{minipage}{0.18\textwidth}
\begin{tikzpicture}[node distance = .3cm, font=\Huge]
\tikzstyle{every node}=[scale=0.3]
% NODES
\node[state] (E) at (0.5,1) {};
\node[state] (F) at (1.5,1) {};
\node[state] (G) at (2.5,1) {};
\node[state] (H) at (3.5,1) {};
\node[state] (A) at (0.5,2) {};
\node[state] (B) at (1.5,2) {};
\node[state] (C) at (2.5,2) {};
\node[state] (D) at (3.5,2) {};
% LETTERS
\node[thick] at (0,1) {$R$};
\node[thick] at (0,2) {$P$};
\node[thick] at (1.5,2.75) {$I$};
\node[thick] at (3.5,2.75) {$E$};
\draw[dotted, thick] (2,0.5) -- (2,2.5);
% ARROWS
\draw[every loop]
(G) edge[bend left] node [scale=1.2, above=0.1 of C] {} (H)
(H) edge[loop right] node [scale=1.2, below=0.1 of H] {\mbox{\Huge \boldmath$ u_1\frac{1}{s}$}} (H);

\end{tikzpicture}
\end{minipage}
}

\begin{minipage}{0.18\textwidth}
Condition: $x_{1}=0$, $x_{2}=0$, $y=0$, $\mathrm{TA}^3_{4}$=I. \\
Therefore, Type II, $x_{2} = 0$, $C_{3}=\neg x_{2} = 1$.
\end{minipage}
\hspace{0.32cm}\resizebox{0.18\textwidth}{!}{
\begin{minipage}{0.18\textwidth}
\begin{tikzpicture}[node distance = .3cm, font=\Huge]
\tikzstyle{every node}=[scale=0.3]
% NODES
\node[state] (E) at (0.5,1) {};
\node[state] (F) at (1.5,1) {};
\node[state] (G) at (2.5,1) {};
\node[state] (H) at (3.5,1) {};
\node[state] (A) at (0.5,2) {};
\node[state] (B) at (1.5,2) {};
\node[state] (C) at (2.5,2) {};
\node[state] (D) at (3.5,2) {};
% LETTERS
\node[thick] at (0,1) {$R$};
\node[thick] at (0,2) {$P$};
\node[thick] at (1.5,2.75) {$I$};
\node[thick] at (3.5,2.75) {$E$};
\draw[dotted, thick] (2,0.5) -- (2,2.5);
% ARROWS
\draw[every loop]
(D) edge[bend right] node [scale=1.2, above=0.1 of C] {\mbox{\Huge \boldmath$ u_2\times 1$}} (C)
(C) edge[bend right] node [scale=1.2, above=0.1 of C] {} (B);

\end{tikzpicture}
\end{minipage}
}

We thirdly study $\mathrm{TA}^3_4$ with action ``Include'', covering the transitions shown below. %We remove all the ``No transition'' case. 

\begin{minipage}{0.18\textwidth}
Condition: $x_{1}=0$, $x_{2}=1$, $y=1$ $\mathrm{TA}^3_{3}$=E. \\
Therefore, Type I, $\neg x_{2} = 0$, $C_3=0$.
\end{minipage}
\hspace{0.32cm}\resizebox{0.18\textwidth}{!}{
\begin{minipage}{0.18\textwidth}
\begin{tikzpicture}[node distance = .3cm, font=\Huge]
\tikzstyle{every node}=[scale=0.3]
% NODES
\node[state] (E) at (0.5,1) {};
\node[state] (F) at (1.5,1) {};
\node[state] (G) at (2.5,1) {};
\node[state] (H) at (3.5,1) {};
\node[state] (A) at (0.5,2) {};
\node[state] (B) at (1.5,2) {};
\node[state] (C) at (2.5,2) {};
\node[state] (D) at (3.5,2) {};
% LETTERS
\node[thick] at (0,1) {$R$};
\node[thick] at (0,2) {$P$};
\node[thick] at (1.5,2.75) {$I$};
\node[thick] at (3.5,2.75) {$E$};
\draw[dotted, thick] (2,0.5) -- (2,2.5);
% ARROWS
\draw[every loop]
(A) edge[bend left] node [scale=1.2, above=0.1 of C] {} (B)
(B) edge[bend left] node [scale=1.2, above=0.1 of C] {~~~~~\mbox{\Huge \boldmath$ u_1\frac{1}{s}$}} (C);

\end{tikzpicture}
\end{minipage}
}

\begin{minipage}{0.18\textwidth}
Condition: $x_{1}=0$, $x_{2}=1$, $y=1$, $\mathrm{TA}^3_{3}$=I.\\
Therefore, Type I, $\neg x_{2} = 0$, $C_3=0$. 
\end{minipage}
\hspace{0.32cm}\resizebox{0.18\textwidth}{!}{
\begin{minipage}{0.18\textwidth}
\begin{tikzpicture}[node distance = .3cm, font=\Huge]
\tikzstyle{every node}=[scale=0.3]
% NODES
\node[state] (E) at (0.5,1) {};
\node[state] (F) at (1.5,1) {};
\node[state] (G) at (2.5,1) {};
\node[state] (H) at (3.5,1) {};
\node[state] (A) at (0.5,2) {};
\node[state] (B) at (1.5,2) {};
\node[state] (C) at (2.5,2) {};
\node[state] (D) at (3.5,2) {};
% LETTERS
\node[thick] at (0,1) {$R$};
\node[thick] at (0,2) {$P$};
\node[thick] at (1.5,2.75) {$I$};
\node[thick] at (3.5,2.75) {$E$};
\draw[dotted, thick] (2,0.5) -- (2,2.5);
% ARROWS
\draw[every loop]
(A) edge[bend left] node [scale=1.2, above=0.1 of C] {} (B)
(B) edge[bend left] node [scale=1.2, above=0.1 of C] {~~~~~\mbox{\Huge \boldmath$ u_1\frac{1}{s}$}} (C);

\end{tikzpicture}
\end{minipage}
}

Lastly, we study $\mathrm{TA}^3_4$ with action ``Exclude'', transitions shown below. %We remove all the ``No transition'' case. 
Similarly, in this situation, when $\mathrm{TA}^3_{3}$ is also excluded, $C_3$ becomes ``empty'' again, as all its associated TAs select action ``Exclude''. Following the training rule of TM, we assign $C_3=1$.

\begin{minipage}{0.18\textwidth}
Conditions: $x_{1}=1$, $x_{2}=1$, $y=0$, $\mathrm{TA}^3_{3}$=E. \\
Therefore, Type II, $\neg x_{2} = 0$, $C_3=1$.
\end{minipage}
\hspace{0.32cm}\resizebox{0.18\textwidth}{!}{
\begin{minipage}{0.18\textwidth}
\begin{tikzpicture}[node distance = .3cm, font=\Huge]
\tikzstyle{every node}=[scale=0.3]
% NODES
\node[state] (E) at (0.5,1) {};
\node[state] (F) at (1.5,1) {};
\node[state] (G) at (2.5,1) {};
\node[state] (H) at (3.5,1) {};
\node[state] (A) at (0.5,2) {};
\node[state] (B) at (1.5,2) {};
\node[state] (C) at (2.5,2) {};
\node[state] (D) at (3.5,2) {};
% LETTERS
\node[thick] at (0,1) {$R$};
\node[thick] at (0,2) {$P$};
\node[thick] at (1.5,2.75) {$I$};
\node[thick] at (3.5,2.75) {$E$};
\draw[dotted, thick] (2,0.5) -- (2,2.5);
% ARROWS
\draw[every loop]
(D) edge[bend right] node [scale=1.2, above=0.1 of C] {} (C)
(C) edge[bend right] node [scale=1.2, above=0.1 of B] {\mbox{\Huge \boldmath$ u_2\times1$}} (B);

\end{tikzpicture}
\end{minipage}
}

\begin{minipage}{0.18\textwidth}
Conditions: $x_{1}=0$, $x_{2}=1$, $y=1$, $\mathrm{TA}^3_{3}$=E. \\
Therefore, Type I, $\neg x_{2} = 0$, $C_3=1$.
\end{minipage}
\hspace{0.32cm}\resizebox{0.18\textwidth}{!}{
\begin{minipage}{0.18\textwidth}
\begin{tikzpicture}[node distance = .3cm, font=\Huge]
\tikzstyle{every node}=[scale=0.3]
% NODES
\node[state] (E) at (0.5,1) {};
\node[state] (F) at (1.5,1) {};
\node[state] (G) at (2.5,1) {};
\node[state] (H) at (3.5,1) {};
\node[state] (A) at (0.5,2) {};
\node[state] (B) at (1.5,2) {};
\node[state] (C) at (2.5,2) {};
\node[state] (D) at (3.5,2) {};
% LETTERS
\node[thick] at (0,1) {$R$};
\node[thick] at (0,2) {$P$};
\node[thick] at (1.5,2.75) {$I$};
\node[thick] at (3.5,2.75) {$E$};
\draw[dotted, thick] (2,0.5) -- (2,2.5);
% ARROWS
\draw[every loop]
(G) edge[bend left] node [scale=1.2, above=0.1 of C] {} (H)
(H) edge[loop right] node [scale=1.2, below=0.1 of H] {\mbox{\Huge \boldmath$ u_1\frac{1}{s}$}} (H);

\end{tikzpicture}
\end{minipage}
}

\begin{minipage}{0.18\textwidth}
Conditions: $x_{1}=1$, $x_{2}=1$, $y=0$, $\mathrm{TA}^3_{3}$=I. \\
Therefore, Type II, $\neg x_{2} = 0$, $C_3=1$.
\end{minipage}
\hspace{0.32cm}\resizebox{0.18\textwidth}{!}{
\begin{minipage}{0.18\textwidth}
\begin{tikzpicture}[node distance = .3cm, font=\Huge]
\tikzstyle{every node}=[scale=0.3]
% NODES
\node[state] (E) at (0.5,1) {};
\node[state] (F) at (1.5,1) {};
\node[state] (G) at (2.5,1) {};
\node[state] (H) at (3.5,1) {};
\node[state] (A) at (0.5,2) {};
\node[state] (B) at (1.5,2) {};
\node[state] (C) at (2.5,2) {};
\node[state] (D) at (3.5,2) {};
% LETTERS
\node[thick] at (0,1) {$R$};
\node[thick] at (0,2) {$P$};
\node[thick] at (1.5,2.75) {$I$};
\node[thick] at (3.5,2.75) {$E$};
\draw[dotted, thick] (2,0.5) -- (2,2.5);
% ARROWS
\draw[every loop]
(D) edge[bend right] node [scale=1.2, above=0.1 of C] {} (C)
(C) edge[bend right] node [scale=1.2, above=0.1 of B] {\mbox{\Huge \boldmath$ u_2\times1$}} (B);

\end{tikzpicture}
\end{minipage}
}

\begin{minipage}{0.18\textwidth}
Conditions: $x_{1}=0$, $x_{2}=1$, $y=1$, $\mathrm{TA}^3_{3}$=I. \\
Therefore, Type I, $\neg x_{2} = 0$, $C_3=1$.
\end{minipage}
\hspace{0.32cm}\resizebox{0.18\textwidth}{!}{
\begin{minipage}{0.18\textwidth}
\begin{tikzpicture}[node distance = .3cm, font=\Huge]
\tikzstyle{every node}=[scale=0.3]
% NODES
\node[state] (E) at (0.5,1) {};
\node[state] (F) at (1.5,1) {};
\node[state] (G) at (2.5,1) {};
\node[state] (H) at (3.5,1) {};
\node[state] (A) at (0.5,2) {};
\node[state] (B) at (1.5,2) {};
\node[state] (C) at (2.5,2) {};
\node[state] (D) at (3.5,2) {};
% LETTERS
\node[thick] at (0,1) {$R$};
\node[thick] at (0,2) {$P$};
\node[thick] at (1.5,2.75) {$I$};
\node[thick] at (3.5,2.75) {$E$};
\draw[dotted, thick] (2,0.5) -- (2,2.5);
% ARROWS
\draw[every loop]
(G) edge[bend left] node [scale=1.2, above=0.1 of C] {} (H)
(H) edge[loop right] node [scale=1.2, below=0.1 of H] {\mbox{\Huge \boldmath$ u_1\frac{1}{s}$}} (H);

\end{tikzpicture}
\end{minipage}
}

Obviously, in Case 3, there is no absorbing state. \\

%%%%%%%%%%%%%%%%%%%%%%%%
%%%%%%%%%%%%%%%%%%%%%%%%%%%
%%%%%%%%%%%%%%%%%%%%%%%Case 4

\noindent {\bf Case 4} \\
Now, we study Case 4, where $\neg x_{1}$ and $x_{1}$ both select ``Include''. For this reason, in this case, we always have $C_{3}=0$. We study firstly $\mathrm{TA}^3_3$ with action ``Include'' and the transitions are shown below. %We remove all the ``No transition'' case. 

\begin{minipage}{0.18\textwidth}
Condition: $x_{1}=0$, $x_{2}=1$, $y=1$, $\mathrm{TA}^3_{4}$=E. \\
Therefore, Type I,  $x_{2} = 1$, $C_{3}= 0$.
\end{minipage}
\hspace{0.32cm}\resizebox{0.18\textwidth}{!}{
\begin{minipage}{0.18\textwidth}
\begin{tikzpicture}[node distance = .3cm, font=\Huge]
\tikzstyle{every node}=[scale=0.3]
% NODES
\node[state] (E) at (0.5,1) {};
\node[state] (F) at (1.5,1) {};
\node[state] (G) at (2.5,1) {};
\node[state] (H) at (3.5,1) {};
\node[state] (A) at (0.5,2) {};
\node[state] (B) at (1.5,2) {};
\node[state] (C) at (2.5,2) {};
\node[state] (D) at (3.5,2) {};
% LETTERS
\node[thick] at (0,1) {$R$};
\node[thick] at (0,2) {$P$};
\node[thick] at (1.5,2.75) {$I$};
\node[thick] at (3.5,2.75) {$E$};
\draw[dotted, thick] (2,0.5) -- (2,2.5);
% ARROWS
\draw[every loop]
(A) edge[bend left] node [scale=1.2, above=0.1 of C] {} (B)
(B) edge[bend left] node [scale=1.2, above=0.1 of C] {\mbox{\Huge \boldmath$ ~~~~~~u_1\frac{1}{s}$}} (C);

\end{tikzpicture}
\end{minipage}
}

\begin{minipage}{0.18\textwidth}
Condition: $x_{1}=0$, $x_{2}=1$, $y=1$, $\mathrm{TA}^3_{4}$=I. \\
Therefore, Type I, $x_{2} = 1$, $C_{3}= 0$.
\end{minipage}
\hspace{0.32cm}\resizebox{0.18\textwidth}{!}{
\begin{minipage}{0.18\textwidth}
\begin{tikzpicture}[node distance = .3cm, font=\Huge]
\tikzstyle{every node}=[scale=0.3]
% NODES
\node[state] (E) at (0.5,1) {};
\node[state] (F) at (1.5,1) {};
\node[state] (G) at (2.5,1) {};
\node[state] (H) at (3.5,1) {};
\node[state] (A) at (0.5,2) {};
\node[state] (B) at (1.5,2) {};
\node[state] (C) at (2.5,2) {};
\node[state] (D) at (3.5,2) {};
% LETTERS
\node[thick] at (0,1) {$R$};
\node[thick] at (0,2) {$P$};
\node[thick] at (1.5,2.75) {$I$};
\node[thick] at (3.5,2.75) {$E$};
\draw[dotted, thick] (2,0.5) -- (2,2.5);
% ARROWS
\draw[every loop]
(A) edge[bend left] node [scale=1.2, above=0.1 of C] {} (B)
(B) edge[bend left] node [scale=1.2, above=0.1 of C] {\mbox{\Huge \boldmath$ ~~~~~u_1\frac{1}{s}$}} (C);

\end{tikzpicture}
\end{minipage}
}

We secondly study $\mathrm{TA}^3_{3}$ with action ``Exclude''.

\begin{minipage}{0.18\textwidth}
Condition: $x_{1}=0$, $x_{2}=1$, $y=1$, $\mathrm{TA}^3_{4}$=E. \\
Therefore, Type I, $x_{2} = 1$, $C_{3}= 0$.
\end{minipage}
\hspace{0.32cm}\resizebox{0.18\textwidth}{!}{
\begin{minipage}{0.18\textwidth}
\begin{tikzpicture}[node distance = .3cm, font=\Huge]
\tikzstyle{every node}=[scale=0.3]
% NODES
\node[state] (E) at (0.5,1) {};
\node[state] (F) at (1.5,1) {};
\node[state] (G) at (2.5,1) {};
\node[state] (H) at (3.5,1) {};
\node[state] (A) at (0.5,2) {};
\node[state] (B) at (1.5,2) {};
\node[state] (C) at (2.5,2) {};
\node[state] (D) at (3.5,2) {};
% LETTERS
\node[thick] at (0,1) {$R$};
\node[thick] at (0,2) {$P$};
\node[thick] at (1.5,2.75) {$I$};
\node[thick] at (3.5,2.75) {$E$};
\draw[dotted, thick] (2,0.5) -- (2,2.5);
% ARROWS
\draw[every loop]
(G) edge[bend left] node [scale=1.2, above=0.1 of C] {} (H)
(H) edge[loop right] node [scale=1.2, below=0.1 of H] {\mbox{\Huge \boldmath$ ~~~~~~u_1\frac{1}{s}$}} (H);

\end{tikzpicture}
\end{minipage}
}

\begin{minipage}{0.18\textwidth}
Condition: $x_{1}=0$, $x_{2}=1$, $y=1$, $\mathrm{TA}^3_{4}$=I. \\
Therefore, Type I, $x_{2} = 1$, $C_{3}= 0$.
\end{minipage}
\hspace{0.32cm}\resizebox{0.18\textwidth}{!}{
\begin{minipage}{0.18\textwidth}
\begin{tikzpicture}[node distance = .3cm, font=\Huge]
\tikzstyle{every node}=[scale=0.3]
% NODES
\node[state] (E) at (0.5,1) {};
\node[state] (F) at (1.5,1) {};
\node[state] (G) at (2.5,1) {};
\node[state] (H) at (3.5,1) {};
\node[state] (A) at (0.5,2) {};
\node[state] (B) at (1.5,2) {};
\node[state] (C) at (2.5,2) {};
\node[state] (D) at (3.5,2) {};
% LETTERS
\node[thick] at (0,1) {$R$};
\node[thick] at (0,2) {$P$};
\node[thick] at (1.5,2.75) {$I$};
\node[thick] at (3.5,2.75) {$E$};
\draw[dotted, thick] (2,0.5) -- (2,2.5);
% ARROWS
\draw[every loop]
(G) edge[bend left] node [scale=1.2, above=0.1 of C] {} (H)
(H) edge[loop right] node [scale=1.2, below=0.1 of H] {\mbox{\Huge \boldmath$ ~~~~~u_1\frac{1}{s}$}} (H);

\end{tikzpicture}
\end{minipage}
}

Now, we study $\mathrm{TA}^3_{4}$ with action ``Include''.

\begin{minipage}{0.18\textwidth}
Condition: $x_{1}=0$, $x_{2}=1$, $y=1$, $\mathrm{TA}^3_{3}$=E. \\
Therefore, Type I, $\neg x_{2} = 0$, $C_{3}= 0$.
\end{minipage}
\hspace{0.32cm}\resizebox{0.18\textwidth}{!}{
\begin{minipage}{0.18\textwidth}
\begin{tikzpicture}[node distance = .3cm, font=\Huge]
\tikzstyle{every node}=[scale=0.3]
% NODES
\node[state] (E) at (0.5,1) {};
\node[state] (F) at (1.5,1) {};
\node[state] (G) at (2.5,1) {};
\node[state] (H) at (3.5,1) {};
\node[state] (A) at (0.5,2) {};
\node[state] (B) at (1.5,2) {};
\node[state] (C) at (2.5,2) {};
\node[state] (D) at (3.5,2) {};
% LETTERS
\node[thick] at (0,1) {$R$};
\node[thick] at (0,2) {$P$};
\node[thick] at (1.5,2.75) {$I$};
\node[thick] at (3.5,2.75) {$E$};
\draw[dotted, thick] (2,0.5) -- (2,2.5);
% ARROWS
\draw[every loop]
(A) edge[bend left] node [scale=1.2, above=0.1 of C] {} (B)
(B) edge[bend left] node [scale=1.2, above=0.1 of C] {\mbox{\Huge \boldmath$ ~~~~~u_1\frac{1}{s}$}} (C);

\end{tikzpicture}
\end{minipage}

}

\begin{minipage}{0.18\textwidth}
Condition: $x_{1}=0$, $x_{2}=1$, $y=1$, $\mathrm{TA}^3_{3}$=I. \\
Therefore, Type I,  $\neg x_{2} = 0$, $C_{3}= 0$.
\end{minipage}
\hspace{0.32cm}\resizebox{0.18\textwidth}{!}{
\begin{minipage}{0.18\textwidth}
\begin{tikzpicture}[node distance = .3cm, font=\Huge]
\tikzstyle{every node}=[scale=0.3]
% NODES
\node[state] (E) at (0.5,1) {};
\node[state] (F) at (1.5,1) {};
\node[state] (G) at (2.5,1) {};
\node[state] (H) at (3.5,1) {};
\node[state] (A) at (0.5,2) {};
\node[state] (B) at (1.5,2) {};
\node[state] (C) at (2.5,2) {};
\node[state] (D) at (3.5,2) {};
% LETTERS
\node[thick] at (0,1) {$R$};
\node[thick] at (0,2) {$P$};
\node[thick] at (1.5,2.75) {$I$};
\node[thick] at (3.5,2.75) {$E$};
\draw[dotted, thick] (2,0.5) -- (2,2.5);
% ARROWS
\draw[every loop]
(A) edge[bend left] node [scale=1.2, above=0.1 of C] {} (B)
(B) edge[bend left] node [scale=1.2, above=0.1 of C] {~~~~~\mbox{\Huge \boldmath$ u_1\frac{1}{s}$}} (C);

\end{tikzpicture}
\end{minipage}
}

We lastly study $\mathrm{TA}^3_{4}$ with action ``Exclude''.

\begin{minipage}{0.18\textwidth}
Condition: $x_1=0$, $x_2=1$, $y=1$, $\mathrm{TA}^3_{3}$=E.\\
Therefore, Type I, $\neg x_{2} = 0$, $C_{3}= 0$.
\end{minipage}
\hspace{0.32cm}\resizebox{0.18\textwidth}{!}{
\begin{minipage}{0.18\textwidth}
\begin{tikzpicture}[node distance = .3cm, font=\Huge]
\tikzstyle{every node}=[scale=0.3]
% NODES
\node[state] (E) at (0.5,1) {};
\node[state] (F) at (1.5,1) {};
\node[state] (G) at (2.5,1) {};
\node[state] (H) at (3.5,1) {};
\node[state] (A) at (0.5,2) {};
\node[state] (B) at (1.5,2) {};
\node[state] (C) at (2.5,2) {};
\node[state] (D) at (3.5,2) {};
% LETTERS
\node[thick] at (0,1) {$R$};
\node[thick] at (0,2) {$P$};
\node[thick] at (1.5,2.75) {$I$};
\node[thick] at (3.5,2.75) {$E$};
\draw[dotted, thick] (2,0.5) -- (2,2.5);
% ARROWS
\draw[every loop]
(G) edge[bend left] node [scale=1.2, above=0.1 of C] {} (H)
(H) edge[loop right] node [scale=1.2, above=0.1 of H] {\mbox{\Huge \boldmath$ u_1\frac{1}{s}$}} (H);

\end{tikzpicture}
\end{minipage}
}

\begin{minipage}{0.18\textwidth}
Condition: $x_1=0$, $x_2=1$, $y=1$, $\mathrm{TA}^3_{3}$=I.\\
Therefore, Type I, $\neg x_{2} = 0$, $C_{3}= 0$.
\end{minipage}
\hspace{0.32cm}\resizebox{0.18\textwidth}{!}{
\begin{minipage}{0.18\textwidth}
\begin{tikzpicture}[node distance = .3cm, font=\Huge]
\tikzstyle{every node}=[scale=0.3]
% NODES
\node[state] (E) at (0.5,1) {};
\node[state] (F) at (1.5,1) {};
\node[state] (G) at (2.5,1) {};
\node[state] (H) at (3.5,1) {};
\node[state] (A) at (0.5,2) {};
\node[state] (B) at (1.5,2) {};
\node[state] (C) at (2.5,2) {};
\node[state] (D) at (3.5,2) {};
% LETTERS
\node[thick] at (0,1) {$R$};
\node[thick] at (0,2) {$P$};
\node[thick] at (1.5,2.75) {$I$};
\node[thick] at (3.5,2.75) {$E$};
\draw[dotted, thick] (2,0.5) -- (2,2.5);
% ARROWS
\draw[every loop]
(G) edge[bend left] node [scale=1.2, above=0.1 of C] {} (H)
(H) edge[loop right] node [scale=1.2, above=0.1 of H] {\mbox{\Huge \boldmath$ u_1\frac{1}{s}$}} (H);

\end{tikzpicture}
\end{minipage}
}

%To summarize Case 4, we realize that both $\mathrm{TA}^3_{3}$ and $\mathrm{TA}^3_{4}$ will converge to ``Exclude''. 

Based on the above analyses, we can now summarize the transitions of $\mathrm{TA}^3_{3}$ and $\mathrm{TA}^3_{4}$, given different configurations of $\mathrm{TA}^3_{1}$ and $\mathrm{TA}^3_{2}$ in Case 1 -- Case 4 (i.e., given four different combinations of $x_{1}$ and $\neg x_{1}$). The arrow shown below means the direction of transitions. %, and ``I'' and ``E'' means ``Include'' and ``Exclude'' respectively. 

\textbf{Scenario 1:} Study $\mathrm{TA}^3_{3}$ = I and $\mathrm{TA}^3_{4}$ = I.

\begin{minipage}{0.225\textwidth}
\textbf{Case 1:} we can see that \\
$\mathrm{TA}^3_{3}$ $\rightarrow$ E \\
$\mathrm{TA}^3_{4}$ $\rightarrow$ E 
\end{minipage}
\begin{minipage}{0.225\textwidth}
\textbf{Case 2:} we can see that \\
$\mathrm{TA}^3_{3}$ $\rightarrow$ E \\
$\mathrm{TA}^3_{4}$ $\rightarrow$ E 
\end{minipage}

\vspace{.5cm}

\begin{minipage}{0.225\textwidth}
\textbf{Case 3:} we can see that \\
$\mathrm{TA}^3_{3}$ $\rightarrow$ E \\
$\mathrm{TA}^3_{4}$ $\rightarrow$ E 
\end{minipage}
\begin{minipage}{0.225\textwidth}
\textbf{Case 4:} we can see that \\
$\mathrm{TA}^3_{3}$ $\rightarrow$ E \\
$\mathrm{TA}^3_{4}$ $\rightarrow$ E 
\end{minipage}

From the facts presented above, it is confirmed that regardless of the state of $\mathrm{TA}^3_{1}$ and $\mathrm{TA}^3_{2}$,
if $\mathrm{TA}^3_{3}$=I and $\mathrm{TA}^3_{4}$=I, they ($\mathrm{TA}^3_{3}$ and $\mathrm{TA}^3_{4}$) will move towards the opposite half of the state space (i.e., towards ``Exclude" ), away from the current state. So, the state with $\mathrm{TA}^3_{3}$=I and $\mathrm{TA}^3_{4}$=I is not absorbing. 

\textbf{Scenario 2:} Study $\mathrm{TA}^3_{3}$ = I and $\mathrm{TA}^3_{4}$= E.

\begin{minipage}{0.225\textwidth}
\textbf{Case 1:} we can see that \\
$\mathrm{TA}^3_{3}$ $\rightarrow$ {\color{red}E} \\
$\mathrm{TA}^3_{4}$ $\rightarrow$ E 
\end{minipage}
\begin{minipage}{0.225\textwidth}
\textbf{Case 2:} we can see that \\
$\mathrm{TA}^3_{3}$ $\rightarrow$ E \\
$\mathrm{TA}^3_{4}$ $\rightarrow$ I, E 
\end{minipage}

\vspace{.5cm}

\begin{minipage}{0.225\textwidth}
\textbf{Case 3:} we can see that \\
$\mathrm{TA}^3_{3}$ $\rightarrow$ I \\
$\mathrm{TA}^3_{4}$ $\rightarrow$ I, E 
\end{minipage}
\begin{minipage}{0.225\textwidth}
\textbf{Case 4:} we can see that \\
$\mathrm{TA}^3_{3}$ $\rightarrow$ E \\
$\mathrm{TA}^3_{4}$ $\rightarrow$ E 
\end{minipage}

In this scenario, the starting point of $\mathrm{TA}^3_{3}$ is ``Include" and that of $\mathrm{TA}^3_{4}$ is ``Exclude". Clearly, actions ``Include" and ``Exclude" for $\mathrm{TA}^3_{3}$ and $\mathrm{TA}^3_{4}$ are not absorbing because none of the cases will make $\mathrm{TA}^3_{3}$ and $\mathrm{TA}^3_{4}$ only move towards ``Include" and ``Exclude". %In Case 1, where $\mathrm{TA}^3_{1}$ = E and $\mathrm{TA}^3_{2}$ = I, both $\mathrm{TA}^3_{3}$ and $\mathrm{TA}^3_{4}$ will move towards ``Exclude". Therefore, given $\mathrm{TA}^3_{1}$ = E and $\mathrm{TA}^3_{2}$ = I hold, $\mathrm{TA}^3_{3}$ in ``Include" and $\mathrm{TA}^3_{4}$ in ``Exclude" are not absorbing anymore. %while in other cases (i.e., in other configurations of $\mathrm{TA}^3_{1}$ and $\mathrm{TA}^3_{2}$), actions ``Include" and ``Exclude" for $\mathrm{TA}^3_{3}$ and $\mathrm{TA}^3_{4}$ are not absorbing. 

\textbf{Scenario 3:} Study $\mathrm{TA}^3_{3}$ = E and $\mathrm{TA}^3_{4}$ = I.

\begin{minipage}{0.225\textwidth}
\textbf{Case 1:} we can see that \\
$\mathrm{TA}^3_{3}$ $\rightarrow$ I, E \\
$\mathrm{TA}^3_{4}$ $\rightarrow$ E 
\end{minipage}
\begin{minipage}{0.225\textwidth}
\textbf{Case 2:} we can see that \\
$\mathrm{TA}^3_{3}$ $\rightarrow$ E \\
$\mathrm{TA}^3_{4}$ $\rightarrow$ E 
\end{minipage}

\vspace{.5cm}

\begin{minipage}{0.225\textwidth}
\textbf{Case 3:} we can see that \\
$\mathrm{TA}^3_{3}$ $\rightarrow$ I, E \\
$\mathrm{TA}^3_{4}$ $\rightarrow$ E 
\end{minipage}
\begin{minipage}{0.225\textwidth}
\textbf{Case 4:} we can see that \\
$\mathrm{TA}^3_{3}$ $\rightarrow$ E \\
$\mathrm{TA}^3_{4}$ $\rightarrow$ E 
\end{minipage}
From the transitions of $\mathrm{TA}^3_{3}$ and $\mathrm{TA}^3_{4}$ in Scenario 3, we can conclude that the state with $\mathrm{TA}^3_{3}$ = E and $\mathrm{TA}^3_{4}$ = I is not absorbing.

\textbf{Scenario 4:} Study $\mathrm{TA}^3_{3}$ = E and $\mathrm{TA}^3_{4}$ = E.

\begin{minipage}{0.225\textwidth}
\textbf{Case 1:} we can see that \\
$\mathrm{TA}^3_{3}$ $\rightarrow$ I \\
$\mathrm{TA}^3_{4}$ $\rightarrow$ E 
\end{minipage}
\begin{minipage}{0.225\textwidth}
\textbf{Case 2:} we can see that \\
$\mathrm{TA}^3_{3}$ $\rightarrow$ E \\
$\mathrm{TA}^3_{4}$ $\rightarrow$ I, E 
\end{minipage}

\vspace{.5cm}

\begin{minipage}{0.225\textwidth}
\textbf{Case 3:} we can see that \\
$\mathrm{TA}^3_{3}$ $\rightarrow$I \\
$\mathrm{TA}^3_{4}$ $\rightarrow$I, E
\end{minipage}
\begin{minipage}{0.225\textwidth}
\textbf{Case 4:} we can see that \\
$\mathrm{TA}^3_{3}$ $\rightarrow$ E \\
$\mathrm{TA}^3_{4}$ $\rightarrow$ E 
\end{minipage}

From the transitions of $\mathrm{TA}^3_{3}$ and $\mathrm{TA}^3_{4}$ in Scenario 4, we can conclude that the state with $\mathrm{TA}^3_{3}$ = E and $\mathrm{TA}^3_{4}$ = E is also absorbing in Case 4, when $\mathrm{TA}^3_{1}$ and $\mathrm{TA}^3_{2}$ have both actions as Include. 

%From the above analysis, we can conclude that when we freeze $\mathrm{TA}^3_{1}$ and $\mathrm{TA}^3_{2}$ with certain actions, there are altogether two absorbing cases. (1) Given that $\mathrm{TA}^3_{1}$ selects ``Exclude" and $\mathrm{TA}^3_{2}$ selects ``Include", $\mathrm{TA}^3_{3}$ selects ``Include" and $\mathrm{TA}^3_{4}$ selects ``Exclude". (2) Given that $\mathrm{TA}^3_{1}$ selects ``Include" and $\mathrm{TA}^3_{2}$ selects ``Include", $\mathrm{TA}^3_{3}$ selects ``Exclude" and $\mathrm{TA}^3_{4}$ selects ``Exclude". 

From the above analysis, we can conclude that when we freeze $\mathrm{TA}^3_{1}$ and $\mathrm{TA}^3_{2}$ with certain actions, there is no absorbing case. Although it seems absorbing for $\mathrm{TA}^3_{3}$ and $\mathrm{TA}^3_{4}$ with both ``Exclude", given the condition that $\mathrm{TA}^3_{1}$ has ``Include" and $\mathrm{TA}^3_{2}$ has ``Include",  due to the condition that is in fact transient,  the absorbing state is not true. % are altogether two absorbing cases. (1) Given that $\mathrm{TA}^3_{1}$ selects ``Exclude" and $\mathrm{TA}^3_{2}$ selects ``Include", $\mathrm{TA}^3_{3}$ selects ``Include" and $\mathrm{TA}^3_{4}$ selects ``Exclude". (2) Given that $\mathrm{TA}^3_{1}$ selects ``Include" and $\mathrm{TA}^3_{2}$ selects ``Include", $\mathrm{TA}^3_{3}$ selects ``Exclude" and $\mathrm{TA}^3_{4}$ selects ``Exclude". 

So far, we have studied the behavior of $\mathrm{TA}^3_3$ and $\mathrm{TA}^3_4$ when the transitions of $\mathrm{TA}^3_1$ and $\mathrm{TA}^3_2$ are frozen. In what follows, we freeze the actions of $\mathrm{TA}^3_3$ and $\mathrm{TA}^3_4$ and study the transitions of $\mathrm{TA}^3_1$ and $\mathrm{TA}^3_2$. \\
\\
\noindent {\bf Case 1} 

Here $\mathrm{TA}^3_3$ is frozen as ``Exclude" and $\mathrm{TA}^3_4$ is ``Include". In this situation, the outputs of $\mathrm{TA}^3_3$ and $\mathrm{TA}^3_4$ give $\neg x_{2}$.

{\bf We firstly study $\mathrm{TA}^3_{1}$ with action ``Include''.}

\begin{minipage}{0.18\textwidth}
Condition: $x_{1}=0$, $x_{2}=1$, $y=1$, $\mathrm{TA}^3_{2}$=E. \\
Therefore, Type I, $x_{1} = 0$, \\$C_3=x_{1} \wedge \neg x_{2}=0$.
\end{minipage}
\hspace{0.32cm}\resizebox{0.18\textwidth}{!}{
\begin{minipage}{0.18\textwidth}
\begin{tikzpicture}[node distance = .3cm, font=\Huge]
\tikzstyle{every node}=[scale=0.3]
% NODES
\node[state] (E) at (0.5,1) {};
\node[state] (F) at (1.5,1) {};
\node[state] (G) at (2.5,1) {};
\node[state] (H) at (3.5,1) {};
\node[state] (A) at (0.5,2) {};
\node[state] (B) at (1.5,2) {};
\node[state] (C) at (2.5,2) {};
\node[state] (D) at (3.5,2) {};
% LETTERS
\node[thick] at (0,1) {$R$};
\node[thick] at (0,2) {$P$};
\node[thick] at (1.5,2.75) {$I$};
\node[thick] at (3.5,2.75) {$E$};
\draw[dotted, thick] (2,0.5) -- (2,2.5);
% ARROWS
\draw[every loop]
(A) edge[bend left] node [scale=1.2, above=0.1 of C] {} (B)
(B) edge[bend left] node [scale=1.2, above=0.1 of C] {\mbox{\Huge \boldmath$ ~~~~~u_1\frac{1}{s}$}} (C);

\end{tikzpicture}
\end{minipage}
}

\begin{minipage}{0.18\textwidth}
Condition: $x_{1}=0$, $x_{2}=1$, $y=1$, $\mathrm{TA}^3_{2}$=I. \\
Therefore, Type I, $x_{1} = 0$,\\ $C_3=0$. 
\end{minipage}
\hspace{0.32cm}\resizebox{0.18\textwidth}{!}{
\begin{minipage}{0.18\textwidth}
\begin{tikzpicture}[node distance = .3cm, font=\Huge]
\tikzstyle{every node}=[scale=0.3]
% NODES
\node[state] (E) at (0.5,1) {};
\node[state] (F) at (1.5,1) {};
\node[state] (G) at (2.5,1) {};
\node[state] (H) at (3.5,1) {};
\node[state] (A) at (0.5,2) {};
\node[state] (B) at (1.5,2) {};
\node[state] (C) at (2.5,2) {};
\node[state] (D) at (3.5,2) {};
% LETTERS
\node[thick] at (0,1) {$R$};
\node[thick] at (0,2) {$P$};
\node[thick] at (1.5,2.75) {$I$};
\node[thick] at (3.5,2.75) {$E$};
\draw[dotted, thick] (2,0.5) -- (2,2.5);
\draw[every loop]
(A) edge[bend left] node [scale=1.2, above=0.1 of C] {} (B)
(B) edge[bend left] node [scale=1.2, above=0.1 of C] {\mbox{\Huge \boldmath$ ~~~~~u_1\frac{1}{s}$}} (C);

\end{tikzpicture}
\end{minipage}
}

{\bf We now study $\mathrm{TA}^3_{1}$ with action ``Exclude''.}

\begin{minipage}{0.18\textwidth}
Condition: $x_{1}=0$, $x_{2}=1$, $y=1$, $\mathrm{TA}^3_{2}$=E. \\
Therefore, Type I, $x_{1} = 0$, \\$C_3=\neg x_{2}=0$. 
\end{minipage}
\hspace{0.32cm}\resizebox{0.18\textwidth}{!}{
\begin{minipage}{0.18\textwidth}
\begin{tikzpicture}[node distance = .3cm, font=\Huge]
\tikzstyle{every node}=[scale=0.3]
% NODES
\node[state] (E) at (0.5,1) {};
\node[state] (F) at (1.5,1) {};
\node[state] (G) at (2.5,1) {};
\node[state] (H) at (3.5,1) {};
\node[state] (A) at (0.5,2) {};
\node[state] (B) at (1.5,2) {};
\node[state] (C) at (2.5,2) {};
\node[state] (D) at (3.5,2) {};
% LETTERS
\node[thick] at (0,1) {$R$};
\node[thick] at (0,2) {$P$};
\node[thick] at (1.5,2.75) {$I$};
\node[thick] at (3.5,2.75) {$E$};
\draw[dotted, thick] (2,0.5) -- (2,2.5);
% ARROWS
\draw[every loop]
(G) edge[bend left] node [scale=1.2, above=0.1 of C] {} (H)
(H) edge[loop right] node [scale=1.2, below=0.1 of H] {\mbox{\Huge \boldmath$ u_1\frac{1}{s}$}} (H);

\end{tikzpicture}
\end{minipage}

}

\begin{minipage}{0.18\textwidth}
Condition: $x_{1}=0$, $x_{2}=0$, $y=0$, $\mathrm{TA}^3_{2}$=E. \\
Therefore, Type II, $x_{1} =0$,\\ $C_3=\neg x_{2}=1$. 
\end{minipage}
\hspace{0.32cm}\resizebox{0.18\textwidth}{!}{
\begin{minipage}{0.18\textwidth}
\begin{tikzpicture}[node distance = .3cm, font=\Huge]
\tikzstyle{every node}=[scale=0.3]
% NODES
\node[state] (E) at (0.5,1) {};
\node[state] (F) at (1.5,1) {};
\node[state] (G) at (2.5,1) {};
\node[state] (H) at (3.5,1) {};
\node[state] (A) at (0.5,2) {};
\node[state] (B) at (1.5,2) {};
\node[state] (C) at (2.5,2) {};
\node[state] (D) at (3.5,2) {};
% LETTERS
\node[thick] at (0,1) {$R$};
\node[thick] at (0,2) {$P$};
\node[thick] at (1.5,2.75) {$I$};
\node[thick] at (3.5,2.75) {$E$};
\draw[dotted, thick] (2,0.5) -- (2,2.5);
% ARROWS
\draw[every loop]
(D) edge[bend right] node [scale=1.2, above=0.1 of C] {\mbox{\Huge \boldmath$ u_2\times 1$}} (C)
(C) edge[bend right] node [scale=1.2, above=0.1 of B] {} (B);

\end{tikzpicture}
\end{minipage}
}

\begin{minipage}{0.18\textwidth}
Condition: $x_{1}=0$, $x_{2}=1$, $y=1$, $\mathrm{TA}^3_{2}$=I. \\
Therefore, Type I, $x_{1}=0$, \\$C_3=\neg x_{1} \wedge \neg x_{2}=0$. 
\end{minipage}
\hspace{0.32cm}\resizebox{0.18\textwidth}{!}{
\begin{minipage}{0.18\textwidth}
\begin{tikzpicture}[node distance = .3cm, font=\Huge]
\tikzstyle{every node}=[scale=0.3]
% NODES
\node[state] (E) at (0.5,1) {};
\node[state] (F) at (1.5,1) {};
\node[state] (G) at (2.5,1) {};
\node[state] (H) at (3.5,1) {};
\node[state] (A) at (0.5,2) {};
\node[state] (B) at (1.5,2) {};
\node[state] (C) at (2.5,2) {};
\node[state] (D) at (3.5,2) {};
% LETTERS
\node[thick] at (0,1) {$R$};
\node[thick] at (0,2) {$P$};
\node[thick] at (1.5,2.75) {$I$};
\node[thick] at (3.5,2.75) {$E$};
\draw[dotted, thick] (2,0.5) -- (2,2.5);
% ARROWS
\draw[every loop]
(G) edge[bend left] node [scale=1.2, above=0.1 of C] {} (H)
(H) edge[loop right] node [scale=1.2, below=0.1 of H] {\mbox{\Huge \boldmath$ u_1\frac{1}{s}$}} (H);

\end{tikzpicture}
\end{minipage}
}

\begin{minipage}{0.18\textwidth}
Condition: $x_{1}=0$, $x_{2}=0$, $y=0$, $\mathrm{TA}^3_{2}$=I. \\
Therefore, Type II, $x_{1} = 0$, \\$C_3=\neg x_{1} \wedge \neg x_{2}=1$.
\end{minipage}
\hspace{0.32cm}\resizebox{0.18\textwidth}{!}{
\begin{minipage}{0.18\textwidth}
\begin{tikzpicture}[node distance = .3cm, font=\Huge]
\tikzstyle{every node}=[scale=0.3]
% NODES
\node[state] (E) at (0.5,1) {};
\node[state] (F) at (1.5,1) {};
\node[state] (G) at (2.5,1) {};
\node[state] (H) at (3.5,1) {};
\node[state] (A) at (0.5,2) {};
\node[state] (B) at (1.5,2) {};
\node[state] (C) at (2.5,2) {};
\node[state] (D) at (3.5,2) {};
% LETTERS
\node[thick] at (0,1) {$R$};
\node[thick] at (0,2) {$P$};
\node[thick] at (1.5,2.75) {$I$};
\node[thick] at (3.5,2.75) {$E$};
\draw[dotted, thick] (2,0.5) -- (2,2.5);
% ARROWS
\draw[every loop]
(D) edge[bend right] node [scale=1.2, above=0.1 of C] {} (C)
(C) edge[bend right] node [scale=1.2, above=0.1 of B] {\mbox{\Huge \boldmath$ u_2\times1$}} (B);
\end{tikzpicture}
\end{minipage}
}

%\pagebreak

{\bf We thirdly study $\mathrm{TA}^3_{2}$ with action ``Include''.} 

\begin{minipage}{0.18\textwidth}
Condition: $x_{1}=0$, $x_{2}=1$, $y=1$, $\mathrm{TA}^3_{1}$=E. \\
Therefore, Type I, $\neg x_{1}=1$, \\$C_3=\neg x_{1} \wedge \neg x_{2}=0$
\end{minipage}
\hspace{0.32cm}\resizebox{0.18\textwidth}{!}{
\begin{minipage}{0.18\textwidth}
\begin{tikzpicture}[node distance = .3cm, font=\Huge]
\tikzstyle{every node}=[scale=0.3]
% NODES
\node[state] (E) at (0.5,1) {};
\node[state] (F) at (1.5,1) {};
\node[state] (G) at (2.5,1) {};
\node[state] (H) at (3.5,1) {};
\node[state] (A) at (0.5,2) {};
\node[state] (B) at (1.5,2) {};
\node[state] (C) at (2.5,2) {};
\node[state] (D) at (3.5,2) {};
% LETTERS
\node[thick] at (0,1) {$R$};
\node[thick] at (0,2) {$P$};
\node[thick] at (1.5,2.75) {$I$};
\node[thick] at (3.5,2.75) {$E$};
\draw[dotted, thick] (2,0.5) -- (2,2.5);
% ARROWS
\draw[every loop]
(A) edge[bend left] node [scale=1.2, above=0.1 of C] {} (B)
(B) edge[bend left] node [scale=1.2, above=0.1 of C] {\mbox{\Huge \boldmath$ ~~~~~u_1\frac{1}{s}$}} (C);

\end{tikzpicture}
\end{minipage}
}

\begin{minipage}{0.18\textwidth}
Condition: $x_{1}=0$, $x_{2}=1$, $y=1$, $\mathrm{TA}^3_1$=I. \\
Therefore, Type I, $\neg x_{1}=1$,\\ $C_3=0$
\end{minipage}
\hspace{0.32cm}\resizebox{0.18\textwidth}{!}{
\begin{minipage}{0.18\textwidth}
\begin{tikzpicture}[node distance = .3cm, font=\Huge]
\tikzstyle{every node}=[scale=0.3]
% NODES
\node[state] (E) at (0.5,1) {};
\node[state] (F) at (1.5,1) {};
\node[state] (G) at (2.5,1) {};
\node[state] (H) at (3.5,1) {};
\node[state] (A) at (0.5,2) {};
\node[state] (B) at (1.5,2) {};
\node[state] (C) at (2.5,2) {};
\node[state] (D) at (3.5,2) {};
% LETTERS
\node[thick] at (0,1) {$R$};
\node[thick] at (0,2) {$P$};
\node[thick] at (1.5,2.75) {$I$};
\node[thick] at (3.5,2.75) {$E$};
\draw[dotted, thick] (2,0.5) -- (2,2.5);
% ARROWS
\draw[every loop]
(A) edge[bend left] node [scale=1.2, above=0.1 of C] {} (B)
(B) edge[bend left] node [scale=1.2, above=0.1 of C] {\mbox{\Huge \boldmath$ ~~~~~u_1\frac{1}{s}$}} (C);
\end{tikzpicture}
\end{minipage}
}

%\pagebreak

{\bf We finally study $\mathrm{TA}^3_{2}$ with action ``Exclude''.}

\begin{minipage}{0.18\textwidth}
Condition: $x_{1}=0$, $x_{2}=1$, $y=1$, $\mathrm{TA}^3_{1}$=E. \\
Therefore, Type I, $\neg x_{1}=1$,\\ $C_3=\neg x_{2}=0$
\end{minipage}
\hspace{0.32cm}\resizebox{0.18\textwidth}{!}{
\begin{minipage}{0.18\textwidth}
\begin{tikzpicture}[node distance = .3cm, font=\Huge]
\tikzstyle{every node}=[scale=0.3]
% NODES
\node[state] (E) at (0.5,1) {};
\node[state] (F) at (1.5,1) {};
\node[state] (G) at (2.5,1) {};
\node[state] (H) at (3.5,1) {};
\node[state] (A) at (0.5,2) {};
\node[state] (B) at (1.5,2) {};
\node[state] (C) at (2.5,2) {};
\node[state] (D) at (3.5,2) {};
% LETTERS
\node[thick] at (0,1) {$R$};
\node[thick] at (0,2) {$P$};
\node[thick] at (1.5,2.75) {$I$};
\node[thick] at (3.5,2.75) {$E$};
\draw[dotted, thick] (2,0.5) -- (2,2.5);
% ARROWS
\draw[every loop]
(G) edge[bend left] node [scale=1.2, above=0.1 of C] {} (H)
(H) edge[loop right] node [scale=1.2, below=0.1 of H] {\mbox{\Huge \boldmath$ ~~u_1\frac{1}{s}$}} (H);

\end{tikzpicture}
\end{minipage}
}

\begin{minipage}{0.18\textwidth}
Condition: $x_{1}=0$, $x_{2}=1$, $y=1$, $\mathrm{TA}^3_{1}$=I. \\
Therefore, Type I, $\neg x_{1}=1$, \\$C_3= x_{1} \wedge \neg x_{2}=0$. 
\end{minipage}
\hspace{0.32cm}\resizebox{0.18\textwidth}{!}{
\begin{minipage}{0.18\textwidth}
\begin{tikzpicture}[node distance = .3cm, font=\Huge]
\tikzstyle{every node}=[scale=0.3]
% NODES
\node[state] (E) at (0.5,1) {};
\node[state] (F) at (1.5,1) {};
\node[state] (G) at (2.5,1) {};
\node[state] (H) at (3.5,1) {};
\node[state] (A) at (0.5,2) {};
\node[state] (B) at (1.5,2) {};
\node[state] (C) at (2.5,2) {};
\node[state] (D) at (3.5,2) {};
% LETTERS
\node[thick] at (0,1) {$R$};
\node[thick] at (0,2) {$P$};
\node[thick] at (1.5,2.75) {$I$};
\node[thick] at (3.5,2.75) {$E$};
\draw[dotted, thick] (2,0.5) -- (2,2.5);
% ARROWS
\draw[every loop]
(G) edge[bend left] node [scale=1.2, above=0.1 of C] {} (H)
(H) edge[loop right] node [scale=1.2, below=0.1 of H] {\mbox{\Huge \boldmath$ u_1\frac{1}{s}$}} (H);

\end{tikzpicture}
\end{minipage}
}

%Clearly $\mathrm{TA}_{3,2}$ will only become excluded, give $\mathrm{TA}_{3,2}$ is excluded, $\mathrm{TA}_{3,1}$ can be either included or excluded.
%\pagebreak

%%%%%%%%%%%%%
%%%%%%%%%%%%%%%%
%%%%%%%%%%%%%%%%
%%%%%%%%%%%%%%%%case 2
%\pagebreak

\noindent {\bf Case 2} \\
Here $\mathrm{TA}^3_3$ is frozen as ``Include" and $\mathrm{TA}^3_4$ is as ``Exclude". In this situation, the outputs of $\mathrm{TA}^3_3$ and $\mathrm{TA}^3_4$ give $x_{2}$.

{\bf We now study $\mathrm{TA}^3_{1}$ with action ``Include''.} \\

\begin{minipage}{0.18\textwidth}
Condition: $x_{1}=0$, $x_{2}=1$, $y=1$, $\mathrm{TA}^3_{2}$=E. \\
Therefore, Type I, $x_{1} = 0$,\\ $C_{3}=x_{1} \wedge x_{2}=0$. 
\end{minipage}
\hspace{0.32cm}\resizebox{0.18\textwidth}{!}{
\begin{minipage}{0.18\textwidth}
\begin{tikzpicture}[node distance = .3cm, font=\Huge]
\tikzstyle{every node}=[scale=0.3]
% NODES
\node[state] (E) at (0.5,1) {};
\node[state] (F) at (1.5,1) {};
\node[state] (G) at (2.5,1) {};
\node[state] (H) at (3.5,1) {};
\node[state] (A) at (0.5,2) {};
\node[state] (B) at (1.5,2) {};
\node[state] (C) at (2.5,2) {};
\node[state] (D) at (3.5,2) {};
% LETTERS
\node[thick] at (0,1) {$R$};
\node[thick] at (0,2) {$P$};
\node[thick] at (1.5,2.75) {$I$};
\node[thick] at (3.5,2.75) {$E$};
\draw[dotted, thick] (2,0.5) -- (2,2.5);
% ARROWS
\draw[every loop]
(A) edge[bend left] node [scale=1.2, above=0.1 of C] {} (B)
(B) edge[bend left] node [scale=1.2, above=0.1 of C] {\mbox{\Huge \boldmath$ ~~~~~u_1\frac{1}{s}$}} (C);

\end{tikzpicture}
\end{minipage}
}

\begin{minipage}{0.18\textwidth}
Condition: $x_{1}=0$, $x_{2}=1$, $y=1$, $\mathrm{TA}^3_{2}$=I. \\
Therefore, Type I, $x_{1} = 0$, \\$C_{3}=\neg x_{1} \wedge x_{1}\wedge x_{2}=0$. 
\end{minipage}
\hspace{0.32cm}\resizebox{0.18\textwidth}{!}{
\begin{minipage}{0.18\textwidth}
\begin{tikzpicture}[node distance = .3cm, font=\Huge]
\tikzstyle{every node}=[scale=0.3]
% NODES
\node[state] (E) at (0.5,1) {};
\node[state] (F) at (1.5,1) {};
\node[state] (G) at (2.5,1) {};
\node[state] (H) at (3.5,1) {};
\node[state] (A) at (0.5,2) {};
\node[state] (B) at (1.5,2) {};
\node[state] (C) at (2.5,2) {};
\node[state] (D) at (3.5,2) {};
% LETTERS
\node[thick] at (0,1) {$R$};
\node[thick] at (0,2) {$P$};
\node[thick] at (1.5,2.75) {$I$};
\node[thick] at (3.5,2.75) {$E$};
\draw[dotted, thick] (2,0.5) -- (2,2.5);
% ARROWS
\draw[every loop]
(A) edge[bend left] node [scale=1.2, above=0.1 of C] {} (B)
(B) edge[bend left] node [scale=1.2, above=0.1 of C] {\mbox{\Huge \boldmath$ ~~~~~u_1\frac{1}{s}$}} (C);

\end{tikzpicture}
\end{minipage}
}

%\pagebreak

%\subsubsection*{Study $\mathrm{TA}_{3,1}$ with action \textsl{exclude}}

{\bf We now study $\mathrm{TA}^3_{1}$ with action ``Exclude''.}

\begin{minipage}{0.18\textwidth}
Condition: 
$x_{1}=0$, $x_{2}=1$, $y=1$, $\mathrm{TA}^3_{2}$=E. \\
Therefore, Type I, $x_{1} = 0$,\\ $C_{3}=x_{1}=1$. 
\end{minipage}
\hspace{0.32cm}\resizebox{0.18\textwidth}{!}{
\begin{minipage}{0.18\textwidth}
\begin{tikzpicture}[node distance = .3cm, font=\Huge]
\tikzstyle{every node}=[scale=0.3]
% NODES
\node[state] (E) at (0.5,1) {};
\node[state] (F) at (1.5,1) {};
\node[state] (G) at (2.5,1) {};
\node[state] (H) at (3.5,1) {};
\node[state] (A) at (0.5,2) {};
\node[state] (B) at (1.5,2) {};
\node[state] (C) at (2.5,2) {};
\node[state] (D) at (3.5,2) {};
% LETTERS
\node[thick] at (0,1) {$R$};
\node[thick] at (0,2) {$P$};
\node[thick] at (1.5,2.75) {$I$};
\node[thick] at (3.5,2.75) {$E$};
\draw[dotted, thick] (2,0.5) -- (2,2.5);
% ARROWS
\draw[every loop]
(G) edge[bend left] node [scale=1.2, above=0.1 of C] {} (H)
(H) edge[loop right] node [scale=1.2, below=0.1 of H] {\mbox{\Huge \boldmath$ u_1\frac{1}{s}$}} (H);

\end{tikzpicture}
\end{minipage}
}

\begin{minipage}{0.18\textwidth}
Condition: $x_{1}=0$, $x_{2}=1$, $y=1$, $\mathrm{TA}^3_{2}$=I. \\
Therefore, Type I, $x_{1} = 0$, \\{\color{red}$C_{3}=\neg x_{1}\wedge x_{2} \wedge 0=0$.} 
\end{minipage}
\hspace{0.32cm}\resizebox{0.18\textwidth}{!}{
\begin{minipage}{0.18\textwidth}
\begin{tikzpicture}[node distance = .3cm, font=\Huge]
\tikzstyle{every node}=[scale=0.3]
% NODES
\node[state] (E) at (0.5,1) {};
\node[state] (F) at (1.5,1) {};
\node[state] (G) at (2.5,1) {};
\node[state] (H) at (3.5,1) {};
\node[state] (A) at (0.5,2) {};
\node[state] (B) at (1.5,2) {};
\node[state] (C) at (2.5,2) {};
\node[state] (D) at (3.5,2) {};
% LETTERS
\node[thick] at (0,1) {$R$};
\node[thick] at (0,2) {$P$};
\node[thick] at (1.5,2.75) {$I$};
\node[thick] at (3.5,2.75) {$E$};
\draw[dotted, thick] (2,0.5) -- (2,2.5);
% ARROWS
\draw[every loop]
(G) edge[bend left] node [scale=1.2, above=0.1 of C] {} (H)
(H) edge[loop right] node [scale=1.2, below=0.1 of H] {\mbox{\Huge \boldmath$ u_1\frac{1}{s}$}} (H);

\end{tikzpicture}
\end{minipage}
}

%\pagebreak

%Study $\mathrm{TA}_{3,2}$ with action \textsl{include}

{\bf We now study $\mathrm{TA}^3_{2}$ with action ``Include''.} 

{\color{red}
\begin{minipage}{0.18\textwidth}
Condition: 
$x_{1}=0$, $x_{2}=1$, $y=1$, $\mathrm{TA}^3_{1}$=E. \\
Therefore, Type I, $\neg x_{1} = 1$, \\$C_{3}=\neg x_{1}\wedge x_{2}\wedge 0=0$.
\end{minipage}
\hspace{0.32cm}\resizebox{0.18\textwidth}{!}{
\begin{minipage}{0.18\textwidth}
\begin{tikzpicture}[node distance = .3cm, font=\Huge]
\tikzstyle{every node}=[scale=0.3]
% NODES
\node[state] (E) at (0.5,1) {};
\node[state] (F) at (1.5,1) {};
\node[state] (G) at (2.5,1) {};
\node[state] (H) at (3.5,1) {};
\node[state] (A) at (0.5,2) {};
\node[state] (B) at (1.5,2) {};
\node[state] (C) at (2.5,2) {};
\node[state] (D) at (3.5,2) {};
% LETTERS
\node[thick] at (0,1) {$R$};
\node[thick] at (0,2) {$P$};
\node[thick] at (1.5,2.75) {$I$};
\node[thick] at (3.5,2.75) {$E$};
\draw[dotted, thick] (2,0.5) -- (2,2.5);
% ARROWS
\draw [every loop]
(A) edge[bend left] node [scale=1.2, above=0.1 of C] {} (B)
(B) edge[bend left] node [scale=1.2, above=0.1 of C] {\mbox{\Huge \boldmath$ ~~~~~u_1\frac{1}{s}$}} (C);

\end{tikzpicture}
\end{minipage}
}
}

\begin{minipage}{0.18\textwidth}
Condition: $x_{1}=0$, $x_{2}=1$, $y=1$, $\mathrm{TA}^3_{1}$=I. \\
Therefore, Type I, $\neg x_{1} = 1$, \\$C_{3}=0$. 
\end{minipage}
\hspace{0.32cm}\resizebox{0.18\textwidth}{!}{
\begin{minipage}{0.18\textwidth}
\begin{tikzpicture}[node distance = .3cm, font=\Huge]
\tikzstyle{every node}=[scale=0.3]
% NODES
\node[state] (E) at (0.5,1) {};
\node[state] (F) at (1.5,1) {};
\node[state] (G) at (2.5,1) {};
\node[state] (H) at (3.5,1) {};
\node[state] (A) at (0.5,2) {};
\node[state] (B) at (1.5,2) {};
\node[state] (C) at (2.5,2) {};
\node[state] (D) at (3.5,2) {};
% LETTERS
\node[thick] at (0,1) {$R$};
\node[thick] at (0,2) {$P$};
\node[thick] at (1.5,2.75) {$I$};
\node[thick] at (3.5,2.75) {$E$};
\draw[dotted, thick] (2,0.5) -- (2,2.5);
% ARROWS
\draw[every loop]
(A) edge[bend left] node [scale=1.2, above=0.1 of C] {} (B)
(B) edge[bend left] node [scale=1.2, above=0.1 of C] {\mbox{\Huge \boldmath$ ~~~~~u_1\frac{1}{s}$}} (C);

\end{tikzpicture}
\end{minipage}
}

%\pagebreak

{\bf We now study $\mathrm{TA}^3_{2}$ with action ``Exclude''.} 

\begin{minipage}{0.18\textwidth}
Condition: $x_{1}=1$, $x_{2}=1$, $y=0$, $\mathrm{TA}^3_{1}$=E. \\
Therefore, Type II, $\neg x_{1} = 0$, \\$C_3= x_{2}=1$. 
\end{minipage}
\hspace{0.32cm}\resizebox{0.18\textwidth}{!}{
\begin{minipage}{0.18\textwidth}
\begin{tikzpicture}[node distance = .3cm, font=\Huge]
\tikzstyle{every node}=[scale=0.3]
% NODES
\node[state] (E) at (0.5,1) {};
\node[state] (F) at (1.5,1) {};
\node[state] (G) at (2.5,1) {};
\node[state] (H) at (3.5,1) {};
\node[state] (A) at (0.5,2) {};
\node[state] (B) at (1.5,2) {};
\node[state] (C) at (2.5,2) {};
\node[state] (D) at (3.5,2) {};
% LETTERS
\node[thick] at (0,1) {$R$};
\node[thick] at (0,2) {$P$};
\node[thick] at (1.5,2.75) {$I$};
\node[thick] at (3.5,2.75) {$E$};
\draw[dotted, thick] (2,0.5) -- (2,2.5);
% ARROWS
\draw[every loop]
(D) edge[bend right] node [scale=1.2, above=0.1 of C] {} (C)
(C) edge[bend right] node [scale=1.2, above=0.1 of B] {\mbox{\Huge \boldmath$ u_2\times1$}} (B);

\end{tikzpicture}
\end{minipage}
}

\begin{minipage}{0.18\textwidth}
Condition: 
$x_{1}=0$, $x_{2}=1$, $y=1$, $\mathrm{TA}^3_{1}$=E \\
Therefore, Type I, $\neg x_{1} = 1$, \\$C_3= x_{2}=1$.
\end{minipage}
\hspace{0.32cm}\resizebox{0.18\textwidth}{!}{
\begin{minipage}{0.18\textwidth}
\begin{tikzpicture}[node distance = .3cm, font=\Huge]
\tikzstyle{every node}=[scale=0.3]
% NODES
\node[state] (E) at (0.5,1) {};
\node[state] (F) at (1.5,1) {};
\node[state] (G) at (2.5,1) {};
\node[state] (H) at (3.5,1) {};
\node[state] (A) at (0.5,2) {};
\node[state] (B) at (1.5,2) {};
\node[state] (C) at (2.5,2) {};
\node[state] (D) at (3.5,2) {};
% LETTERS
\node[thick] at (0,1) {$R$};
\node[thick] at (0,2) {$P$};
\node[thick] at (1.5,2.75) {$I$};
\node[thick] at (3.5,2.75) {$E$};
\draw[dotted, thick] (2,0.5) -- (2,2.5);
% ARROWS
\draw[every loop]
(D) edge[bend right] node [scale=1.2, above=0.1 of C] {} (C)
(C) edge[bend right] node [scale=1.2, above=0.1 of C] {\mbox{\Huge \boldmath$ ~~~~~~~~u_1\frac{s-1}{s}$}} (B);

\end{tikzpicture}
\end{minipage}
}

\begin{minipage}{0.18\textwidth}
Condition: 
$x_{1}=1$, $x_{2}=1$, $y=0$, $\mathrm{TA}^3_{1}$=I.\\
Therefore, Type II, $\neg x_{1} = 0$, \\$C_3= x_{1} \wedge x_{2}=1$. 
\end{minipage}
\hspace{0.32cm}\resizebox{0.18\textwidth}{!}{
\begin{minipage}{0.18\textwidth}
\begin{tikzpicture}[node distance = .3cm, font=\Huge]
\tikzstyle{every node}=[scale=0.3]
% NODES
\node[state] (E) at (0.5,1) {};
\node[state] (F) at (1.5,1) {};
\node[state] (G) at (2.5,1) {};
\node[state] (H) at (3.5,1) {};
\node[state] (A) at (0.5,2) {};
\node[state] (B) at (1.5,2) {};
\node[state] (C) at (2.5,2) {};
\node[state] (D) at (3.5,2) {};
% LETTERS
\node[thick] at (0,1) {$R$};
\node[thick] at (0,2) {$P$};
\node[thick] at (1.5,2.75) {$I$};
\node[thick] at (3.5,2.75) {$E$};
\draw[dotted, thick] (2,0.5) -- (2,2.5);
% ARROWS
\draw[every loop]
(D) edge[bend right] node [scale=1.2, above=0.1 of C] {} (C)
(C) edge[bend right] node [scale=1.2, above=0.1 of C] {\mbox{\Huge \boldmath$ u_2\times1$}} (B);

\end{tikzpicture}
\end{minipage}
}

\begin{minipage}{0.18\textwidth}
Condition: 
$x_{1}=0$, $x_{2}=1$, $y=1$, $\mathrm{TA}^3_{1}$=I.\\
Therefore, Type I, $\neg x_{1} = 1$, \\$C_3= x_{1} \wedge x_{2}=0$. 
\end{minipage}
\hspace{0.32cm}\resizebox{0.18\textwidth}{!}{
\begin{minipage}{0.18\textwidth}
\begin{tikzpicture}[node distance = .3cm, font=\Huge]
\tikzstyle{every node}=[scale=0.3]
% NODES
\node[state] (E) at (0.5,1) {};
\node[state] (F) at (1.5,1) {};
\node[state] (G) at (2.5,1) {};
\node[state] (H) at (3.5,1) {};
\node[state] (A) at (0.5,2) {};
\node[state] (B) at (1.5,2) {};
\node[state] (C) at (2.5,2) {};
\node[state] (D) at (3.5,2) {};
% LETTERS
\node[thick] at (0,1) {$R$};
\node[thick] at (0,2) {$P$};
\node[thick] at (1.5,2.75) {$I$};
\node[thick] at (3.5,2.75) {$E$};
\draw[dotted, thick] (2,0.5) -- (2,2.5);
% ARROWS
\draw[every loop]
(G) edge[bend left] node [scale=1.2, above=0.1 of C] {} (H)
(H) edge[loop right] node [scale=1.2, below=0.1 of H] {\mbox{\Huge \boldmath$ u_1\frac{1}{s}$}} (H);

\end{tikzpicture}
\end{minipage}
}

%In this situation, $\mathrm{TA}^3_{1}$ will only move to ``Exclude", and  $\mathrm{TA}^3_{2}$ will become ``Include".

%\pagebreak

%%%%%%%%%%%%%%%
%%%%%%%%%%%%%%%%%
%%%%%%%%%%%%%%%%%%
%%%%%%%%%%%case 3

\noindent {\bf Case 3} \\
Here $\mathrm{TA}^3_3$ is frozen as ``Exclude" and $\mathrm{TA}^3_4$ is as ``Exclude". In this case, the second bit $x_2$ does not
play any role for the output. %In this situation, the output of $\mathrm{TA}^3_3$ and $\mathrm{TA}^3_4$ is $x_{2}$.

{\bf We now study $\mathrm{TA}^3_{1}$ with action ``Include''.}

\begin{minipage}{0.18\textwidth}
Condition: $x_{1}=0$, $x_{2}=1$, $y=1$, $\mathrm{TA}^3_{2}$=E. \\
Therefore, Type I, $x_1=0$, \\$C_{3}= x_{1}=0$. 
\end{minipage}
\hspace{0.32cm}\resizebox{0.18\textwidth}{!}{
\begin{minipage}{0.18\textwidth}
\begin{tikzpicture}[node distance = .3cm, font=\Huge]
\tikzstyle{every node}=[scale=0.3]
% NODES
\node[state] (E) at (0.5,1) {};
\node[state] (F) at (1.5,1) {};
\node[state] (G) at (2.5,1) {};
\node[state] (H) at (3.5,1) {};
\node[state] (A) at (0.5,2) {};
\node[state] (B) at (1.5,2) {};
\node[state] (C) at (2.5,2) {};
\node[state] (D) at (3.5,2) {};
% LETTERS
\node[thick] at (0,1) {$R$};
\node[thick] at (0,2) {$P$};
\node[thick] at (1.5,2.75) {$I$};
\node[thick] at (3.5,2.75) {$E$};
\draw[dotted, thick] (2,0.5) -- (2,2.5);
% ARROWS
\draw[every loop]
(A) edge[bend left] node [scale=1.2, above=0.1 of C] {} (B)
(B) edge[bend left] node [scale=1.2, above=0.1 of C] {\mbox{\Huge \boldmath$ ~~~~~u_1\frac{1}{s}$}} (C);
\end{tikzpicture}
\end{minipage}
}

\begin{minipage}{0.18\textwidth}
Condition: $x_{1}=0$, $x_{2}=1$, $y=1$, $\mathrm{TA}^3_{2}$=I. \\
Therefore, Type I, $x_{1} = 0$, \\$C_{3}= x_{1} \wedge \neg x_{1}=0$. 
\end{minipage}
\hspace{0.32cm}\resizebox{0.18\textwidth}{!}{
\begin{minipage}{0.18\textwidth}
\begin{tikzpicture}[node distance = .3cm, font=\Huge]
\tikzstyle{every node}=[scale=0.3]
% NODES
\node[state] (E) at (0.5,1) {};
\node[state] (F) at (1.5,1) {};
\node[state] (G) at (2.5,1) {};
\node[state] (H) at (3.5,1) {};
\node[state] (A) at (0.5,2) {};
\node[state] (B) at (1.5,2) {};
\node[state] (C) at (2.5,2) {};
\node[state] (D) at (3.5,2) {};
% LETTERS
\node[thick] at (0,1) {$R$};
\node[thick] at (0,2) {$P$};
\node[thick] at (1.5,2.75) {$I$};
\node[thick] at (3.5,2.75) {$E$};
\draw[dotted, thick] (2,0.5) -- (2,2.5);
% ARROWS
\draw[every loop]
(A) edge[bend left] node [scale=1.2, above=0.1 of C] {} (B)
(B) edge[bend left] node [scale=1.2, above=0.1 of C] {\mbox{\Huge \boldmath$ ~~~~~u_1\frac{1}{s}$}} (C);

\end{tikzpicture}
\end{minipage}
}

%\pagebreak
{\bf We now study $\mathrm{TA}^3_{1}$ with action ``Exclude''.}

%Study $\mathrm{TA}^3_{1}$ with action \textsl{exclude}

%In this situation, if $\mathrm{TA}_{3,2}$ is also excluded, there is No transition. Therefore, we only study where $\mathrm{TA}_{3,2}$ is included.

\begin{minipage}{0.18\textwidth}
Condition: $x_{1}=0$, $x_{2}=0$, $y=0$,  $\mathrm{TA}^3_{2}$=E. \\
Therefore, Type II, $x_{1} = 0$,\\ $C_{3}=1$.
\end{minipage}
\hspace{0.32cm}\resizebox{0.18\textwidth}{!}{
\begin{minipage}{0.18\textwidth}
\begin{tikzpicture}[node distance = .3cm, font=\Huge]
\tikzstyle{every node}=[scale=0.3]
% NODES
\node[state] (E) at (0.5,1) {};
\node[state] (F) at (1.5,1) {};
\node[state] (G) at (2.5,1) {};
\node[state] (H) at (3.5,1) {};
\node[state] (A) at (0.5,2) {};
\node[state] (B) at (1.5,2) {};
\node[state] (C) at (2.5,2) {};
\node[state] (D) at (3.5,2) {};
% LETTERS
\node[thick] at (0,1) {$R$};
\node[thick] at (0,2) {$P$};
\node[thick] at (1.5,2.75) {$I$};
\node[thick] at (3.5,2.75) {$E$};
\draw[dotted, thick] (2,0.5) -- (2,2.5);
% ARROWS
\draw[every loop]
(D) edge[bend right] node [scale=1.2, above=0.1 of C] {} (C)
(C) edge[bend right] node [scale=1.2, above=0.1 of B] {\mbox{\Huge \boldmath$ u_2\times1$}} (B);

\end{tikzpicture}
\end{minipage}
}

\begin{minipage}{0.18\textwidth}
Condition: $x_{1}=0$, $x_{2}=1$, $y=1$, $\mathrm{TA}^3_{2}$=E. \\
Therefore, Type I, $x_{1} = 0$, \\$C_{3}=1$. 
\end{minipage}
\hspace{0.32cm}\resizebox{0.18\textwidth}{!}{
\begin{minipage}{0.18\textwidth}
\begin{tikzpicture}[node distance = .3cm, font=\Huge]
\tikzstyle{every node}=[scale=0.3]
% NODES
\node[state] (E) at (0.5,1) {};
\node[state] (F) at (1.5,1) {};
\node[state] (G) at (2.5,1) {};
\node[state] (H) at (3.5,1) {};
\node[state] (A) at (0.5,2) {};
\node[state] (B) at (1.5,2) {};
\node[state] (C) at (2.5,2) {};
\node[state] (D) at (3.5,2) {};
% LETTERS
\node[thick] at (0,1) {$R$};
\node[thick] at (0,2) {$P$};
\node[thick] at (1.5,2.75) {$I$};
\node[thick] at (3.5,2.75) {$E$};
\draw[dotted, thick] (2,0.5) -- (2,2.5);
% ARROWS
\draw[every loop]
(G) edge[bend left] node [scale=1.2, above=0.1 of C] {} (H)
(H) edge[loop right] node [scale=1.2, below=0.1 of H] {\mbox{\Huge \boldmath$ ~~~~u_1\frac{1}{s}$}} (H);

\end{tikzpicture}
\end{minipage}
}

\begin{minipage}{0.18\textwidth}
Condition: $x_{1}=0$, $x_{2}=0$, $y=0$,  $\mathrm{TA}^3_{2}$=I. \\
Therefore, Type II, $x_{1} = 0$,\\ $C_{3}=\neg x_{1}=1$.
\end{minipage}
\hspace{0.32cm}\resizebox{0.18\textwidth}{!}{
\begin{minipage}{0.18\textwidth}
\begin{tikzpicture}[node distance = .3cm, font=\Huge]
\tikzstyle{every node}=[scale=0.3]
% NODES
\node[state] (E) at (0.5,1) {};
\node[state] (F) at (1.5,1) {};
\node[state] (G) at (2.5,1) {};
\node[state] (H) at (3.5,1) {};
\node[state] (A) at (0.5,2) {};
\node[state] (B) at (1.5,2) {};
\node[state] (C) at (2.5,2) {};
\node[state] (D) at (3.5,2) {};
% LETTERS
\node[thick] at (0,1) {$R$};
\node[thick] at (0,2) {$P$};
\node[thick] at (1.5,2.75) {$I$};
\node[thick] at (3.5,2.75) {$E$};
\draw[dotted, thick] (2,0.5) -- (2,2.5);
% ARROWS
\draw[every loop]
(D) edge[bend right] node [scale=1.2, above=0.1 of C] {} (C)
(C) edge[bend right] node [scale=1.2, above=0.1 of B] {\mbox{\Huge \boldmath$ u_2\times1$}} (B);

\end{tikzpicture}
\end{minipage}
}

%In this situation, if $\mathrm{TA}_{3,2}$ is also excluded, there is No transition. Therefore, we only study where $\mathrm{TA}_{3,2}$ is included.

\begin{minipage}{0.18\textwidth}
Condition; $x_{1}=0$, $x_{2}=1$, $y=1$, $\mathrm{TA}^3_{2}$=I. \\
Therefore, Type I, $x_{1} = 0$, \\$C_{3}=\neg x_{1}=1$. 
\end{minipage}
\hspace{0.32cm}\resizebox{0.18\textwidth}{!}{
\begin{minipage}{0.18\textwidth}
\begin{tikzpicture}[node distance = .3cm, font=\Huge]
\tikzstyle{every node}=[scale=0.3]
% NODES
\node[state] (E) at (0.5,1) {};
\node[state] (F) at (1.5,1) {};
\node[state] (G) at (2.5,1) {};
\node[state] (H) at (3.5,1) {};
\node[state] (A) at (0.5,2) {};
\node[state] (B) at (1.5,2) {};
\node[state] (C) at (2.5,2) {};
\node[state] (D) at (3.5,2) {};
% LETTERS
\node[thick] at (0,1) {$R$};
\node[thick] at (0,2) {$P$};
\node[thick] at (1.5,2.75) {$I$};
\node[thick] at (3.5,2.75) {$E$};
\draw[dotted, thick] (2,0.5) -- (2,2.5);
% ARROWS
\draw[every loop]
(G) edge[bend left] node [scale=1.2, above=0.1 of C] {} (H)
(H) edge[loop right] node [scale=1.2, below=0.1 of H] {\mbox{\Huge \boldmath$ u_1\frac{1}{s}$}} (H);

\end{tikzpicture}
\end{minipage}
}

%\pagebreak

{\bf We now study $\mathrm{TA}^3_{2}$ with action ``Include''.}

%Study $\mathrm{TA}_{3,2}$ with action \textsl{include}

\begin{minipage}{0.18\textwidth}
Condition: $x_{1}=0$, $x_{2}=1$, $y=1$, $\mathrm{TA}^3_{1}$=E. \\
Therefore, Type I, $\neg x_{1} = 1$, \\$C_3=\neg x_{1}=1$.
\end{minipage}
\hspace{0.32cm}\resizebox{0.18\textwidth}{!}{
\begin{minipage}{0.18\textwidth}
\begin{tikzpicture}[node distance = .3cm, font=\Huge]
\tikzstyle{every node}=[scale=0.3]
% NODES
\node[state] (E) at (0.85,1) {};
\node[state] (F) at (1.5,1) {};
\node[state] (G) at (2.5,1) {};
\node[state] (H) at (3.5,1) {};
\node[state] (A) at (0.85,2) {};
\node[state] (B) at (1.5,2) {};
\node[state] (C) at (2.5,2) {};
\node[state] (D) at (3.5,2) {};
% LETTERS
\node[thick] at (0,1) {$R$};
\node[thick] at (0,2) {$P$};
\node[thick] at (1.5,2.75) {$I$};
\node[thick] at (3.5,2.75) {$E$};
\draw[dotted, thick] (2,0.5) -- (2,2.5);
% ARROWS
\draw[every loop]
(F) edge[bend left] node [scale=1.2, above=0.1 of E] {} (E)
(E) edge[loop left] node [scale=1.2, below=0.1 of E] {\mbox{\Huge \boldmath$ u_1\frac{s-1}{s}$}} (E);

\end{tikzpicture}
\end{minipage}
}

\begin{minipage}{0.18\textwidth}
Condition: $x_{1}=0$, $x_{2}=1$, $y=1$, $\mathrm{TA}^3_{1}$=I. \\
Therefore, Type I, $\neg x_{1} = 1$, \\$C_3=0$. \\
\end{minipage}
\hspace{0.32cm}\resizebox{0.18\textwidth}{!}{
\begin{minipage}{0.18\textwidth}
\begin{tikzpicture}[node distance = .3cm, font=\Huge]
\tikzstyle{every node}=[scale=0.3]
% NODES
\node[state] (E) at (0.5,1) {};
\node[state] (F) at (1.5,1) {};
\node[state] (G) at (2.5,1) {};
\node[state] (H) at (3.5,1) {};
\node[state] (A) at (0.5,2) {};
\node[state] (B) at (1.5,2) {};
\node[state] (C) at (2.5,2) {};
\node[state] (D) at (3.5,2) {};
% LETTERS
\node[thick] at (0,1) {$R$};
\node[thick] at (0,2) {$P$};
\node[thick] at (1.5,2.75) {$I$};
\node[thick] at (3.5,2.75) {$E$};
\draw[dotted, thick] (2,0.5) -- (2,2.5);
% ARROWS
\draw[every loop]
(A) edge[bend left] node [scale=1.2, above=0.1 of C] {} (B)
(B) edge[bend left] node [scale=1.2, above=0.1 of C] {\mbox{\Huge \boldmath$ ~~~~~u_1\frac{1}{s}$}} (C);

\end{tikzpicture}
\end{minipage}
}

{\bf We now study $\mathrm{TA}^3_{2}$ with action ``Exclude''.}

%Study $\mathrm{TA}_{3,2}$ with action \textsl{exclude}

%In this situation, if $\mathrm{TA}_{3,1}$ is also exclude, there is No transition. Therefore, we only study for $\mathrm{TA}_{3,1}$ is included. 

\begin{minipage}{0.18\textwidth}
Condition: $x_{1}=1$, $x_{2}=1$, $y=0$, $\mathrm{TA}^3_{1}$=E.\\
Therefore, Type II, $\neg x_{1} = 0$,\\ $C_3=x_{1}=1$. 
\end{minipage}
\hspace{0.32cm}\resizebox{0.18\textwidth}{!}{
\begin{minipage}{0.18\textwidth}
\begin{tikzpicture}[node distance = .3cm, font=\Huge]
\tikzstyle{every node}=[scale=0.3]
% NODES
\node[state] (E) at (0.5,1) {};
\node[state] (F) at (1.5,1) {};
\node[state] (G) at (2.5,1) {};
\node[state] (H) at (3.5,1) {};
\node[state] (A) at (0.5,2) {};
\node[state] (B) at (1.5,2) {};
\node[state] (C) at (2.5,2) {};
\node[state] (D) at (3.5,2) {};
% LETTERS
\node[thick] at (0,1) {$R$};
\node[thick] at (0,2) {$P$};
\node[thick] at (1.5,2.75) {$I$};
\node[thick] at (3.5,2.75) {$E$};
\draw[dotted, thick] (2,0.5) -- (2,2.5);
% ARROWS
\draw[every loop]
(D) edge[bend right] node [scale=1.2, above=0.1 of C] {\mbox{\Huge \boldmath$ u_2\times1$}} (C)
(C) edge[bend right] node [scale=1.2, above=0.1 of C] {} (B);

\end{tikzpicture}
\end{minipage}
}

\begin{minipage}{0.18\textwidth}
Condition: $x_{1}=0$, $x_{2}=1$, $y=1$, $\mathrm{TA}^3_{1}$=E. \\
Therefore, Type I, $\neg x_{1} = 0$,\\ $C_3=x_{1}=1$.
\end{minipage}
\hspace{0.32cm}\resizebox{0.18\textwidth}{!}{
\begin{minipage}{0.18\textwidth}
\begin{tikzpicture}[node distance = .3cm, font=\Huge]
\tikzstyle{every node}=[scale=0.3]
% NODES
\node[state] (E) at (0.5,1) {};
\node[state] (F) at (1.5,1) {};
\node[state] (G) at (2.5,1) {};
\node[state] (H) at (3.5,1) {};
\node[state] (A) at (0.5,2) {};
\node[state] (B) at (1.5,2) {};
\node[state] (C) at (2.5,2) {};
\node[state] (D) at (3.5,2) {};
% LETTERS
\node[thick] at (0,1) {$R$};
\node[thick] at (0,2) {$P$};
\node[thick] at (1.5,2.75) {$I$};
\node[thick] at (3.5,2.75) {$E$};
\draw[dotted, thick] (2,0.5) -- (2,2.5);
% ARROWS
\draw[every loop]
(G) edge[bend left] node [scale=1.2, above=0.1 of C] {} (H)
(H) edge[loop right] node [scale=1.2, below=0.1 of H] {\mbox{\Huge \boldmath$ u_1\frac{1}{s}$}} (H);

\end{tikzpicture}
\end{minipage}
}

\begin{minipage}{0.18\textwidth}
Condition: $x_{1}=1$, $x_{2}=1$, $y=0$, $\mathrm{TA}^3_{1}$=I. \\
Therefore, Type II, $\neg x_{1} = 0$,\\ $C_3=x_{1}=1$. 
\end{minipage}
\hspace{0.32cm}\resizebox{0.18\textwidth}{!}{
\begin{minipage}{0.18\textwidth}
\begin{tikzpicture}[node distance = .3cm, font=\Huge]
\tikzstyle{every node}=[scale=0.3]
% NODES
\node[state] (E) at (0.5,1) {};
\node[state] (F) at (1.5,1) {};
\node[state] (G) at (2.5,1) {};
\node[state] (H) at (3.5,1) {};
\node[state] (A) at (0.5,2) {};
\node[state] (B) at (1.5,2) {};
\node[state] (C) at (2.5,2) {};
\node[state] (D) at (3.5,2) {};
% LETTERS
\node[thick] at (0,1) {$R$};
\node[thick] at (0,2) {$P$};
\node[thick] at (1.5,2.75) {$I$};
\node[thick] at (3.5,2.75) {$E$};
\draw[dotted, thick] (2,0.5) -- (2,2.5);
% ARROWS
\draw[every loop]
(D) edge[bend right] node [scale=1.2, above=0.1 of C] {\mbox{\Huge \boldmath$ u_2\times1$}} (C)
(C) edge[bend right] node [scale=1.2, above=0.1 of C] {} (B);

\end{tikzpicture}
\end{minipage}
}

\begin{minipage}{0.18\textwidth}
Condition: $x_{1}=0$, $x_{2}=1$, $y=1$, $\mathrm{TA}^3_{1}$=I. \\
Therefore, Type I, $\neg x_{1} = 0$,\\ $C_3=x_{1}=1$.
\end{minipage}
\hspace{0.32cm}\resizebox{0.18\textwidth}{!}{
\begin{minipage}{0.18\textwidth}
\begin{tikzpicture}[node distance = .3cm, font=\Huge]
\tikzstyle{every node}=[scale=0.3]
% NODES
\node[state] (E) at (0.5,1) {};
\node[state] (F) at (1.5,1) {};
\node[state] (G) at (2.5,1) {};
\node[state] (H) at (3.5,1) {};
\node[state] (A) at (0.5,2) {};
\node[state] (B) at (1.5,2) {};
\node[state] (C) at (2.5,2) {};
\node[state] (D) at (3.5,2) {};
% LETTERS
\node[thick] at (0,1) {$R$};
\node[thick] at (0,2) {$P$};
\node[thick] at (1.5,2.75) {$I$};
\node[thick] at (3.5,2.75) {$E$};
\draw[dotted, thick] (2,0.5) -- (2,2.5);
% ARROWS
\draw[every loop]
(G) edge[bend left] node [scale=1.2, above=0.1 of C] {} (H)
(H) edge[loop right] node [scale=1.2, below=0.1 of H] {\mbox{\Huge \boldmath$ u_1\frac{1}{s}$}} (H);

\end{tikzpicture}
\end{minipage}
}

%\pagebreak

%%%%%%%%%%%%%%%
%%%%%%%%%%%%%%%%%
%%%%%%%%%%%%%%%%%%
%%%%%%%%%%%case 3
\noindent {\bf Case 4} \\
Here $\mathrm{TA}^3_3$ is frozen as ``Include" and $\mathrm{TA}^3_4$ is as ``Include". In this situation, the output of the clause is always 0.

{\bf We now study $\mathrm{TA}^3_{1}$ with action ``Include''.}

\begin{minipage}{0.18\textwidth}
Condition: $x_{1}=0$, $x_{2}=1$, $y=1$, $\mathrm{TA}^3_{2}$=E. \\
Therefore, Type I, $x_{1} = 0$, \\$C_{3}=0$. 
\end{minipage}
\hspace{0.32cm}\resizebox{0.18\textwidth}{!}{
\begin{minipage}{0.18\textwidth}
\begin{tikzpicture}[node distance = .3cm, font=\Huge]
\tikzstyle{every node}=[scale=0.3]
% NODES
\node[state] (E) at (0.5,1) {};
\node[state] (F) at (1.5,1) {};
\node[state] (G) at (2.5,1) {};
\node[state] (H) at (3.5,1) {};
\node[state] (A) at (0.5,2) {};
\node[state] (B) at (1.5,2) {};
\node[state] (C) at (2.5,2) {};
\node[state] (D) at (3.5,2) {};
% LETTERS
\node[thick] at (0,1) {$R$};
\node[thick] at (0,2) {$P$};
\node[thick] at (1.5,2.75) {$I$};
\node[thick] at (3.5,2.75) {$E$};
\draw[dotted, thick] (2,0.5) -- (2,2.5);
% ARROWS
\draw[every loop]
(A) edge[bend left] node [scale=1.2, above=0.1 of C] {} (B)
(B) edge[bend left] node [scale=1.2, above=0.1 of C] {\mbox{\Huge \boldmath$ ~~~~~u_1\frac{1}{s}$}} (C);

\end{tikzpicture}
\end{minipage}
}

\begin{minipage}{0.18\textwidth}
Condition: $x_{1}=0$, $x_{2}=1$, $y=1$, $\mathrm{TA}^3_{2}$=I. \\
Therefore, Type I, $x_{1} = 0$, \\$C_{3}=0$
\end{minipage}
\hspace{0.32cm}\resizebox{0.18\textwidth}{!}{
\begin{minipage}{0.18\textwidth}
\begin{tikzpicture}[node distance = .3cm, font=\Huge]
\tikzstyle{every node}=[scale=0.3]
% NODES
\node[state] (E) at (0.5,1) {};
\node[state] (F) at (1.5,1) {};
\node[state] (G) at (2.5,1) {};
\node[state] (H) at (3.5,1) {};
\node[state] (A) at (0.5,2) {};
\node[state] (B) at (1.5,2) {};
\node[state] (C) at (2.5,2) {};
\node[state] (D) at (3.5,2) {};
% LETTERS
\node[thick] at (0,1) {$R$};
\node[thick] at (0,2) {$P$};
\node[thick] at (1.5,2.75) {$I$};
\node[thick] at (3.5,2.75) {$E$};
\draw[dotted, thick] (2,0.5) -- (2,2.5);
% ARROWS
\draw[every loop]
(A) edge[bend left] node [scale=1.2, above=0.1 of C] {} (B)
(B) edge[bend left] node [scale=1.2, above=0.1 of C] {\mbox{\Huge \boldmath$ ~~~~~u_1\frac{1}{s}$}} (C);

\end{tikzpicture}
\end{minipage}
}

%\pagebreak

{\bf We now study $\mathrm{TA}^3_{1}$ with action ``Exclude''.}

\begin{minipage}{0.18\textwidth}
Condition: $x_{1}=0$, $x_{2}=1$, $y=1$, $\mathrm{TA}^3_{2}$=E. \\
Therefore, Type I, $x_{1}=0$, \\$C_{3}=0$. 
\end{minipage}
\hspace{0.32cm}\resizebox{0.18\textwidth}{!}{
\begin{minipage}{0.18\textwidth}
\begin{tikzpicture}[node distance = .3cm, font=\Huge]
\tikzstyle{every node}=[scale=0.3]
% NODES
\node[state] (E) at (0.5,1) {};
\node[state] (F) at (1.5,1) {};
\node[state] (G) at (2.5,1) {};
\node[state] (H) at (3.5,1) {};
\node[state] (A) at (0.5,2) {};
\node[state] (B) at (1.5,2) {};
\node[state] (C) at (2.5,2) {};
\node[state] (D) at (3.5,2) {};
% LETTERS
\node[thick] at (0,1) {$R$};
\node[thick] at (0,2) {$P$};
\node[thick] at (1.5,2.75) {$I$};
\node[thick] at (3.5,2.75) {$E$};
\draw[dotted, thick] (2,0.5) -- (2,2.5);
% ARROWS
\draw[every loop]
(G) edge[bend left] node [scale=1.2, above=0.1 of C] {} (H)
(H) edge[loop right] node [scale=1.2, below=0.1 of H] {\mbox{\Huge \boldmath$ u_1\frac{1}{s}$}} (H);

\end{tikzpicture}
\end{minipage}
}

\begin{minipage}{0.18\textwidth}
Condition: $x_{1}=0$, $x_{2}=1$, $y=0$, $\mathrm{TA}^3_{2}$=I. \\
Therefore, Type II, $x_{1}=0$, \\$C_{3}=0$. 
\end{minipage}
\hspace{0.32cm}\resizebox{0.18\textwidth}{!}{
\begin{minipage}{0.18\textwidth}
\begin{tikzpicture}[node distance = .3cm, font=\Huge]
\tikzstyle{every node}=[scale=0.3]
% NODES
\node[state] (E) at (0.5,1) {};
\node[state] (F) at (1.5,1) {};
\node[state] (G) at (2.5,1) {};
\node[state] (H) at (3.5,1) {};
\node[state] (A) at (0.5,2) {};
\node[state] (B) at (1.5,2) {};
\node[state] (C) at (2.5,2) {};
\node[state] (D) at (3.5,2) {};
% LETTERS
\node[thick] at (0,1) {$R$};
\node[thick] at (0,2) {$P$};
\node[thick] at (1.5,2.75) {$I$};
\node[thick] at (3.5,2.75) {$E$};
\draw[dotted, thick] (2,0.5) -- (2,2.5);
% ARROWS
\draw[every loop]
(G) edge[bend left] node [scale=1.2, above=0.1 of C] {} (H)
(H) edge[loop right] node [scale=1.2, below=0.1 of H] {\mbox{\Huge \boldmath$ u_1\frac{1}{s}$}} (H);

\end{tikzpicture}
\end{minipage}
}

{\bf We now study $\mathrm{TA}^3_{2}$ with action ``Include''.}

\begin{minipage}{0.18\textwidth}
Condition: $x_{1}=0$, $x_{2}=1$, $y=1$, $\mathrm{TA}^3_{1}$=E. \\
Therefore, Type I, $\neg x_{1}=1$, \\$C_3=0$. 
\end{minipage}
\hspace{0.32cm}\resizebox{0.18\textwidth}{!}{
\begin{minipage}{0.18\textwidth}
\begin{tikzpicture}[node distance = .3cm, font=\Huge]
\tikzstyle{every node}=[scale=0.3]
% NODES
\node[state] (E) at (0.5,1) {};
\node[state] (F) at (1.5,1) {};
\node[state] (G) at (2.5,1) {};
\node[state] (H) at (3.5,1) {};
\node[state] (A) at (0.5,2) {};
\node[state] (B) at (1.5,2) {};
\node[state] (C) at (2.5,2) {};
\node[state] (D) at (3.5,2) {};
% LETTERS
\node[thick] at (0,1) {$R$};
\node[thick] at (0,2) {$P$};
\node[thick] at (1.5,2.75) {$I$};
\node[thick] at (3.5,2.75) {$E$};
\draw[dotted, thick] (2,0.5) -- (2,2.5);
% ARROWS
\draw[every loop]
(A) edge[bend left] node [scale=1.2, above=0.1 of C] {} (B)
(B) edge[bend left] node [scale=1.2, above=0.1 of C] {\mbox{\Huge \boldmath$ ~~~~~u_1\frac{1}{s}$}} (C);

\end{tikzpicture}
\end{minipage}
}

\begin{minipage}{0.18\textwidth}
Condition: $x_{1}=0$, $x_{2}=1$, $y=1$, $\mathrm{TA}^3_{1}$=I. \\
Therefore, Type I, $\neg x_{1}=1$, \\$C_3=0$.
\end{minipage}
\hspace{0.32cm}\resizebox{0.18\textwidth}{!}{
\begin{minipage}{0.18\textwidth}
\begin{tikzpicture}[node distance = .3cm, font=\Huge]
\tikzstyle{every node}=[scale=0.3]
% NODES
\node[state] (E) at (0.5,1) {};
\node[state] (F) at (1.5,1) {};
\node[state] (G) at (2.5,1) {};
\node[state] (H) at (3.5,1) {};
\node[state] (A) at (0.5,2) {};
\node[state] (B) at (1.5,2) {};
\node[state] (C) at (2.5,2) {};
\node[state] (D) at (3.5,2) {};
% LETTERS
\node[thick] at (0,1) {$R$};
\node[thick] at (0,2) {$P$};
\node[thick] at (1.5,2.75) {$I$};
\node[thick] at (3.5,2.75) {$E$};
\draw[dotted, thick] (2,0.5) -- (2,2.5);
% ARROWS
\draw[every loop]
(A) edge[bend left] node [scale=1.2, above=0.1 of C] {} (B)
(B) edge[bend left] node [scale=1.2, above=0.1 of C] {\mbox{\Huge \boldmath$ ~~~~~u_1\frac{1}{s}$}} (C);

\end{tikzpicture}
\end{minipage}
}

%\pagebreak

{\bf We now study $\mathrm{TA}^3_{2}$ with action ``Exclude''.}

\begin{minipage}{0.18\textwidth}
Condition: $x_{1}=0$, $x_{2}=1$, $y=1$, $\mathrm{TA}^3_{1}$=E. \\
Therefore, Type I, $\neg x_{1}=1$, \\$C_{3}=0$.
\end{minipage}
\hspace{0.32cm}\resizebox{0.18\textwidth}{!}{
\begin{minipage}{0.18\textwidth}
\begin{tikzpicture}[node distance = .3cm, font=\Huge]
\tikzstyle{every node}=[scale=0.3]
% NODES
\node[state] (E) at (0.5,1) {};
\node[state] (F) at (1.5,1) {};
\node[state] (G) at (2.5,1) {};
\node[state] (H) at (3.5,1) {};
\node[state] (A) at (0.5,2) {};
\node[state] (B) at (1.5,2) {};
\node[state] (C) at (2.5,2) {};
\node[state] (D) at (3.5,2) {};
% LETTERS
\node[thick] at (0,1) {$R$};
\node[thick] at (0,2) {$P$};
\node[thick] at (1.5,2.75) {$I$};
\node[thick] at (3.5,2.75) {$E$};
\draw[dotted, thick] (2,0.5) -- (2,2.5);
% ARROWS
\draw[every loop]
(G) edge[bend left] node [scale=1.2, above=0.1 of C] {} (H)
(H) edge[loop right] node [scale=1.2, below=0.1 of H] {\mbox{\Huge \boldmath$ u_1\frac{1}{s}$}} (H);

\end{tikzpicture}
\end{minipage}
}

\begin{minipage}{0.18\textwidth}
Condition: $x_{1}=0$, $x_{2}=1$, $y=1$, $\mathrm{TA}^3_{1}$=I. \\
Therefore, Type I, $\neg x_{1}=1$, \\$C_{3}=0$. 
\end{minipage}
\hspace{0.32cm}\resizebox{0.18\textwidth}{!}{
\begin{minipage}{0.18\textwidth}
\begin{tikzpicture}[node distance = .3cm, font=\Huge]
\tikzstyle{every node}=[scale=0.3]
% NODES
\node[state] (E) at (0.5,1) {};
\node[state] (F) at (1.5,1) {};
\node[state] (G) at (2.5,1) {};
\node[state] (H) at (3.5,1) {};
\node[state] (A) at (0.5,2) {};
\node[state] (B) at (1.5,2) {};
\node[state] (C) at (2.5,2) {};
\node[state] (D) at (3.5,2) {};
% LETTERS
\node[thick] at (0,1) {$R$};
\node[thick] at (0,2) {$P$};
\node[thick] at (1.5,2.75) {$I$};
\node[thick] at (3.5,2.75) {$E$};
\draw[dotted, thick] (2,0.5) -- (2,2.5);
% ARROWS
\draw[every loop]
(G) edge[bend left] node [scale=1.2, above=0.1 of C] {} (H)
(H) edge[loop right] node [scale=1.2, below=0.1 of H] {\mbox{\Huge \boldmath$ u_1\frac{1}{s}$}} (H);

\end{tikzpicture}
\end{minipage}
}

%\pagebreak

Based on the analysis performed above, we can show the directions of transitions for $\mathrm{TA}^3_1$ and $\mathrm{TA}^3_2$ given different configurations of $\mathrm{TA}^3_3$ and $\mathrm{TA}^3_4$.

\textbf{Scenario 1:} Study $\mathrm{TA}^3_1$ = I and $\mathrm{TA}^3_2$ = E.

\begin{minipage}{0.225\textwidth}
\textbf{Case 1:} we can see that \\
$\mathrm{TA}^3_1$ $\rightarrow$ E \\
$\mathrm{TA}^3_2$ $\rightarrow$ E 
\end{minipage}
\begin{minipage}{0.225\textwidth}
\textbf{Case 2:} we can see that \\
$\mathrm{TA}^3_1$ $\rightarrow$ E \\
$\mathrm{TA}^3_2$ $\rightarrow$ I, E 
\end{minipage}

\begin{minipage}{0.225\textwidth}
\textbf{Case 3:} we can see that \\
$\mathrm{TA}^3_1$ $\rightarrow$ E \\
$\mathrm{TA}^3_2$ $\rightarrow$ I,E
\end{minipage}
\begin{minipage}{0.225\textwidth}
\textbf{Case 4:} we can see that \\
$\mathrm{TA}^3_1$ $\rightarrow$ E \\
$\mathrm{TA}^3_2$ $\rightarrow$ E 
\end{minipage}

\vspace{.5cm}

\textbf{Scenario 2:} Study $\mathrm{TA}^3_1$ = I and $\mathrm{TA}^3_2$ = I.

\begin{minipage}{0.225\textwidth}
\textbf{Case 1:} we can see that \\
$\mathrm{TA}^3_1$ $\rightarrow$ E \\
$\mathrm{TA}^3_2$ $\rightarrow$ E 
\end{minipage}
\begin{minipage}{0.225\textwidth}
\textbf{Case 2:} we can see that \\
$\mathrm{TA}^3_1$ $\rightarrow$ E \\
$\mathrm{TA}^3_2$ $\rightarrow$ E 
\end{minipage}

\begin{minipage}{0.225\textwidth}
\textbf{Case 3:} we can see that \\
$\mathrm{TA}^3_1$ $\rightarrow$ E \\
$\mathrm{TA}^3_2$ $\rightarrow$ E 
\end{minipage}
\begin{minipage}{0.225\textwidth}
\textbf{Case 4:} we can see that \\
$\mathrm{TA}^3_1$ $\rightarrow$ E \\
$\mathrm{TA}^3_2$ $\rightarrow$ E 
\end{minipage}

\vspace{.5cm}

\textbf{Scenario 3:} Study $\mathrm{TA}^3_1$ = E and $\mathrm{TA}^3_2$ = I.

\begin{minipage}{0.225\textwidth}
\textbf{Case 1:} we can see that \\
$\mathrm{TA}^3_1$ $\rightarrow$ I, E \\
$\mathrm{TA}^3_2$ $\rightarrow$ E
\end{minipage}
\begin{minipage}{0.225\textwidth}
\textbf{Case 2:} we can see that \\
$\mathrm{TA}^3_1$ $\rightarrow$ E \\
$\mathrm{TA}^3_2$ $\rightarrow$ {\color{red} E}
\end{minipage}

\begin{minipage}{0.225\textwidth}
\textbf{Case 3:} we can see that \\
$\mathrm{TA}^3_1$ $\rightarrow$ I \\
$\mathrm{TA}^3_2$ $\rightarrow$ I
\end{minipage}
\begin{minipage}{0.225\textwidth}
\textbf{Case 4:} we can see that \\
$\mathrm{TA}^3_1$ $\rightarrow$ E \\
$\mathrm{TA}^3_2$ $\rightarrow$ E 
\end{minipage}

\vspace{.5cm}

\textbf{Scenario 4:} Study $\mathrm{TA}^3_1$ = E and $\mathrm{TA}^3_2$ = E.

\begin{minipage}{0.225\textwidth}
\textbf{Case 1:} we can see that \\
$\mathrm{TA}^3_1$ $\rightarrow$ I, E \\
$\mathrm{TA}^3_2$ $\rightarrow$ E
\end{minipage}
\begin{minipage}{0.225\textwidth}
\textbf{Case 2:} we can see that \\
$\mathrm{TA}^3_1$ $\rightarrow$ E\\
$\mathrm{TA}^3_2$ $\rightarrow$ I 
\end{minipage}

\begin{minipage}{0.225\textwidth}
\textbf{Case 3:} we can see that \\
$\mathrm{TA}^3_1$ $\rightarrow$ E\\
$\mathrm{TA}^3_2$ $\rightarrow$ I, E 
\end{minipage}
\begin{minipage}{0.225\textwidth}
\textbf{Case 4:} we can see that \\
$\mathrm{TA}^3_1$ $\rightarrow$ E \\
$\mathrm{TA}^3_2$ $\rightarrow$ E 
\end{minipage}

From the above transitions, we can conclude that state $\mathrm{TA}^3_1$=E and $\mathrm{TA}^3_2$=I is absorbing when the state $\mathrm{TA}^3_3$=I and $\mathrm{TA}^3_4$=E are frozen. Similarly, state $\mathrm{TA}^3_1$=E and $\mathrm{TA}^3_2$=E is also absorbing when $\mathrm{TA}^3_3$ and $\mathrm{TA}^3_4$ are both frozen as Include. The other states are not absorbing. 

%\pagebreak

%\subsection{Additional Empirical Results}

\end{document}